\documentclass{article}
\ifdefined\pdfminorversion\pdfminorversion=7\fi

\PassOptionsToPackage{numbers}{natbib}
\usepackage[main,final]{neurips_2026}
\setcitestyle{numbers,square,comma}

\usepackage[T1]{fontenc}
\usepackage{microtype}
\usepackage{graphicx}
\usepackage{subcaption}
\usepackage{booktabs}
\usepackage{amssymb}
\usepackage{mathtools}
\usepackage{amsthm}
\usepackage{xcolor}
\usepackage{listings}
\usepackage{tabularx}
\usepackage[skins,breakable]{tcolorbox}
\usepackage{pifont}
\usepackage{float}
\usepackage{wrapfig}
\usepackage{needspace}
\usepackage{algorithm}
\usepackage{algpseudocode}
\usepackage{colortbl}
\usepackage{tikz}
\usepackage[hidelinks]{hyperref}
\hypersetup{
  pdftitle={CauSim: Scaling Causal Reasoning with Increasingly Complex Causal Simulators},
  pdfauthor={Nicolás Astorga, Anita Kriz, Mihaela van der Schaar}
}

\definecolor{generalGreen}{RGB}{235,240,179}
\definecolor{darkgreen}{RGB}{34,120,34}
\definecolor{darkred}{RGB}{180,50,50}
\definecolor{myPurple}{HTML}{8E76C9}
\definecolor{myBlue}{HTML}{3DAAD6}
\definecolor{teal}{HTML}{008080}

\newcommand{\cmark}{\textcolor{darkgreen}{\ding{51}}}
\newcommand{\xmark}{\textcolor{darkred}{\ding{55}}}
\newcommand{\basedelta}{\hphantom{\textcolor{green!60!black}{(+0.00 )}}}

\theoremstyle{definition}
\newtheorem{definition}{Definition}[section]
\algrenewcommand\alglinenumber[1]{\tiny #1:}

\usetikzlibrary{calc}
\tikzset{stairwayStyle/.style={line join=round,line cap=round,line width=0.04em}}

\newcommand{\stairwayupoutline}{\mathbin{%
\tikz[baseline=(stairwayanchor.base)]{
    \node (stairwayanchor) {\quad};

    \draw[stairwayStyle, draw=black, fill=black]
        ($(stairwayanchor.south west) + (0.0em,0.4em)$)
        -- ++(0,0.20em)
        -- ++(0.20em,0) -- ++(0,0.20em)
        -- ++(0.20em,0) -- ++(0,0.20em)
        -- ++(0.20em,0)
        -- ++(0,-0.60em)
        -- cycle;

    \draw[stairwayStyle, ->, draw=black]
        ($(stairwayanchor.south west) + (-.1em,0.7em)$)
        -- ++(0.4em,0.4em);
}}%
}

\newcommand{\stairwayupfilled}{\mathbin{%
\tikz[baseline=(stairwayanchor.base)]{
    \node (stairwayanchor) {\quad};

    \draw[stairwayStyle]
        ($(stairwayanchor.south west) + (0.0em,0.4em)$)
        -- ++(0,0.20em)
        -- ++(0.20em,0) -- ++(0,0.20em)
        -- ++(0.20em,0) -- ++(0,0.20em)
        -- ++(0.20em,0)
        -- ++(0,-0.60em)
        -- cycle;

    \draw[stairwayStyle, ->]
        ($(stairwayanchor.south west) + (-.1em,0.7em)$)
        -- ++(0.4em,0.4em);
}}%
}

\title{\textbf{$\stairwayupoutline$CauSim}: Scaling Causal Reasoning with\\ Increasingly Complex Causal Simulators}

\author{%
  \textbf{Nicolás Astorga}\textsuperscript{1}\thanks{Equal contribution.}\quad
  \textbf{Anita Kriz}\textsuperscript{1}\footnotemark[1]\quad
  \textbf{Mihaela van der Schaar}\textsuperscript{1} \\
  \textsuperscript{1}DAMTP, University of Cambridge, Cambridge, UK \\
  \texttt{nja46@cam.ac.uk, ak2680@cam.ac.uk}
}

\newcommand{\appindexrow}[2]{%
  \hyperref[#1]{\textbf{Appendix~\ref*{#1}}} &
  #2 &
  \hyperref[#1]{\pageref*{#1}}\\
}

\newcommand{\appindexsubrow}[2]{%
  \hspace{1.1em}\hyperref[#1]{Appendix~\ref*{#1}} &
  #2 &
  \hyperref[#1]{\pageref*{#1}}\\
}

\newcommand{\appendixindex}{%
  \clearpage
  \phantomsection
  \section*{Appendix Index}
  \addcontentsline{toc}{section}{Appendix Index}
  \vspace{-0.5em}
  \begingroup
  \small
  \setlength{\tabcolsep}{3.5pt}
  \renewcommand{\arraystretch}{1.12}
  \noindent
  \begin{tabular}{@{}p{0.19\linewidth}p{\dimexpr0.74\linewidth-4\tabcolsep\relax}>{\raggedleft\arraybackslash}p{0.07\linewidth}@{}}
  \toprule
  \rowcolor{teal!10}
  \textbf{Link} & \textbf{What to look for} & \textbf{Page} \\
  \midrule

  \appindexrow{app:experimental_details}{
    \textbf{Experimental details, prompts, and examples.}
    Construction, verification, translation, dataset configurations, and representative SCMs.
  }
  \addlinespace[0.3em]
  \appindexrow{app:performance_anchoring}{
    \textbf{Absolute-performance anchoring.}
    Comparisons against the base, larger/newer Qwen models, and GPT-5.5.
  }
  \appindexsubrow{app:performance_anchoring_details}{
    Reported evaluation protocol and query-type results.
  }
  \addlinespace[0.3em]
  \appindexrow{app:model_family}{
    \textbf{Generalization across model families.}
    Llama~3.2~3B results across domains and representations.
  }
  \addlinespace[0.3em]
  \appindexrow{app:7b-comparisons}{
    \textbf{7B generator comparisons.}
    Self-generated, GPT-5-mini, and GPT-5 training data, including delta-highlighted results.
  }
  \addlinespace[0.3em]
  \appindexrow{app:informal-experiments}{
    \textbf{Informal representation experiments.}
    Medical and nonsensical natural-language SCMs, with topology-specific results.
  }
  \addlinespace[0.3em]
  \appindexrow{app:additional-benchmarks}{
    \textbf{Additional benchmarks and diagnostics.}
    Formalized GSM8K, bnlearn networks, and one-shot generation failures.
  }
  \appindexsubrow{app:partial_gsm8k}{
    Partially specified, verbalized GSM8K evaluation.
  }
  \appindexsubrow{app:passk_majority}{
    External-benchmark Pass@$k$ and majority-vote results.
  }
  \addlinespace[0.3em]
  \appindexrow{appendix:causal_sim}{
    \textbf{Why causal simulators?}
    Domain examples and meaningful causal queries.
  }
  \addlinespace[0.3em]
  \appindexrow{app:llm_vs_procedural}{
    \textbf{LLM-based SCM generation.}
    Design rationale and procedural-generator baselines.
  }
  \addlinespace[0.3em]
  \appindexrow{app:discrete_scms}{
    \textbf{Discrete SCMs.}
    Finite-support verification and the distinction from continuous inference.
  }
  \addlinespace[0.3em]
  \appindexrow{app:appendix-related-work}{
    \textbf{Related-work comparison.}
    Capabilities of prior SCM-based approaches.
  }
  \addlinespace[0.3em]
  \appindexrow{app:pattern-vs-mechanism}{
    \textbf{Pattern recognition versus mechanism learning.}
    Scope of causal-query accuracy and tests of mechanistic reasoning.
  }
  \addlinespace[0.3em]
  \appindexrow{app:self-improvement-limitations}{
    \textbf{Limits of self-improvement.}
    One-step bootstrapping, generator support, and recursive-training limitations.
  }
  \addlinespace[0.3em]
  \appindexrow{appendix:algorithm}{
    \textbf{Incremental SCM construction.}
    Full procedure and pseudocode.
  }
  \addlinespace[0.3em]
  \appindexrow{appendix:prompts}{
    \textbf{Generation prompts.}
    Instantiation, topology, planning, and localized execution templates.
  }
  \addlinespace[0.3em]
  \appindexrow{sec:prompt}{
    \textbf{Evaluation prompts.}
    Counterfactual and intervention examples.
  }
  \addlinespace[0.3em]
  \appindexrow{appendix:exp_details}{
    \textbf{Training details.}
    Pipeline, SFT settings, hyperparameter selection, and LoRA configurations.
  }
  \addlinespace[0.3em]
  \appindexrow{app:data_generation}{
    \textbf{Data generation.}
    Support-valued semantics, trace sampling, query construction, and prompt assembly.
  }
  \addlinespace[0.3em]
  \appindexrow{app:additional_results}{
    \textbf{Additional results.}
    Query types, NIHSS augmentation, curriculum, data volume, and SCM/query allocation.
  }
  \bottomrule
  \end{tabular}
  \endgroup
}

\newtcolorbox{customblockquote}{
  colframe=myPurple,
  colback=myPurple!10,
  boxrule=0pt,
  left=5pt,
  right=4pt,
  top=5pt,
  bottom=3pt,
  arc=0pt,
  breakable,
  before skip=1.2\baselineskip,
  after skip=0.7\baselineskip,
  left skip=0pt,
  right skip=0pt,
  enhanced jigsaw,
  frame hidden,
  overlay={
    \draw[myPurple, line width=2pt]
      ([yshift=1pt]frame.north west) -- (frame.south west);
    \node[inner sep=0pt] at ([xshift=0pt, yshift=-1.3pt]frame.north west) {};
  },
  fontupper=\fontfamily{ptm}\selectfont,
  boxsep=1pt,
}

\newtcolorbox{customblocktakeaway}{
  colframe=myBlue,
  colback=myBlue!10,
  boxrule=0pt,
  left=4pt,
  right=5pt,
  top=5pt,
  bottom=3pt,
  arc=0pt,
  breakable,
  before skip=1.2\baselineskip,
  after skip=0.7\baselineskip,
  left skip=0pt,
  right skip=0pt,
  enhanced jigsaw,
  frame hidden,
   overlay={
    \draw[myBlue, line width=2pt]
      ([yshift=1pt]frame.north west) -- (frame.south west);
  },
  fontupper=\fontfamily{ptm}\selectfont,
  boxsep=1pt,
}

\begin{document}

\maketitle
\setcounter{footnote}{0}

\begin{abstract}
Despite surpassing human performance across mathematics, coding, and other knowledge intensive tasks, large language models (LLMs) continue to struggle to causally reason. One obstacle is the target data itself: causal systems are complex and often expressed in non-executable forms, and ground-truth answers to causal queries are inherently scarce. We introduce \textbf{$\protect\stairwayupfilled$CauSim}, a framework that turns causal reasoning from a scarce-label problem into a scalable, supervised one. It constructs \textit{increasingly complex causal simulators}: executable structural causal models (SCMs), incrementally built by LLMs, that scale to globally complex systems while maintaining verifiable answers to any causal query. \textbf{$\protect\stairwayupfilled$CauSim} operates across representations by formalizing non-executable causal knowledge into code, allowing for data augmentation, and informalizing executable SCMs into natural language, enabling supervision in previously unsupervisable representations. We structure our research into two parts: (1) how to construct increasingly complex causal simulators, and (2) a systematic study of what \textbf{$\protect\stairwayupfilled$CauSim} enables, demonstrating generalization across representations, consistent gains from curriculum scaling and data volume, LLM self-improvement through self-generated simulators, and data augmentation via formalization of existing domain knowledge. 
Code: \href{https://github.com/jumpynitro/causim}{%
\textcolor[RGB]{0,0,238}{GitHub Repository}}.
\vspace{-.17cm}

\end{abstract}

\vspace{-.185cm}
\section{Introduction}
\vspace{-.75em}

Understanding how a system behaves and reasoning causally about that system are two distinct capabilities. Knowing the laws governing a physical system, for example, does not by itself tell us where a ball would have landed had it been pushed with a different force. Even when the causal mechanisms are explicitly provided, answering such what-if questions requires using the observed world to infer the relevant conditions and reasoning through how a hypothetical change would propagate through the system. Despite surpassing human performance across mathematics~\cite{math_olympics}, coding~\cite{chen2021evaluatinglargelanguagemodels,Li_2022}, and other knowledge-intensive tasks~\cite{10.1145/3728963,Luo_2024}, Large Language Models (LLMs) struggle to causally reason~\cite{jin2023cladder,ashwani2024causeandeffect,chi2024unveiling,zevcevic2023causal}.

\begin{figure}[t]
    \centering
\includegraphics[width=1.0\textwidth]{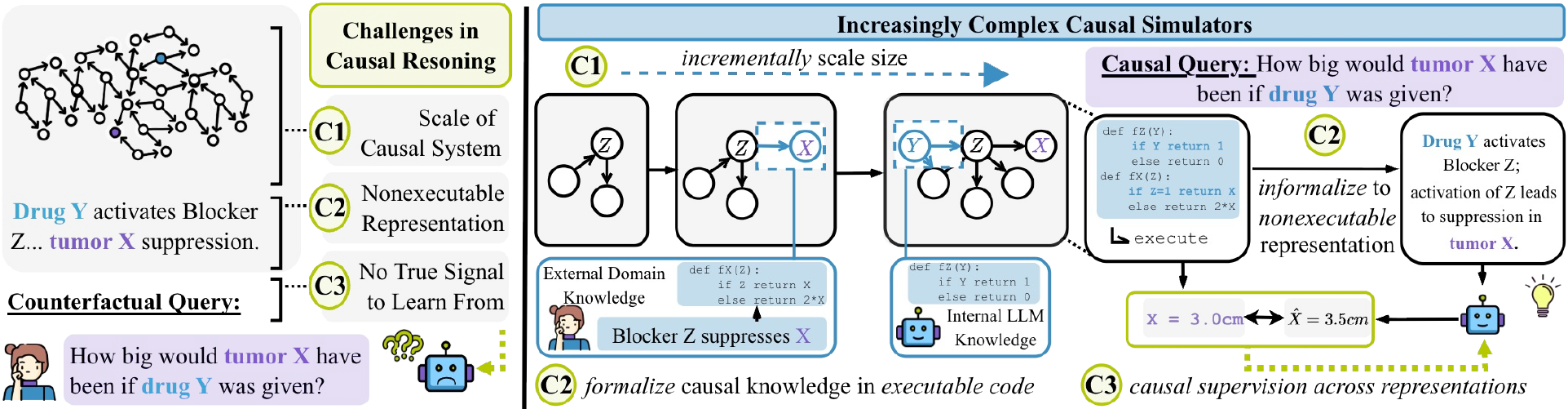}
    \captionsetup{font=footnotesize}
    \caption{\textbf{Overview of $\protect\stairwayupfilled$CauSim.} \textbf{(Left)} Causal reasoning faces three challenges: \textbf{[C1]} scale, \textbf{[C2]} non-executable representations, and \textbf{[C3]} lack of ground-truth supervision. \textbf{(Right)} \textbf{$\protect\stairwayupfilled$CauSim} addresses these through increasingly complex causal simulators: SCMs are built \textit{incrementally} to scale complexity \textbf{[C1]}; non-executable causal knowledge is \textit{formalized} into executable code and executable SCMs are \textit{informalized} into natural language \textbf{[C2]}; enabling causal supervision \textbf{[C3]}}
    \label{fig:framework}
    \vspace{-0.295cm}
\end{figure}

\textbf{Why is causal reasoning hard, even when the rules are known?} Consider the counterfactual query in Fig.~\ref{fig:framework}: \textit{How big would Tumor X have been if Drug Y had been given?} Suppose we are already given the relevant causal mechanisms: how Drug Y affects Blocker Z, and how Blocker Z influences Tumor X. Reasoning about the query is still hard for two reasons. First, $\blacktriangleright$ \textbf{[C1] scale}: as the number of variables and interactions grows, correctly composing mechanisms and tracing their effects becomes increasingly challenging. Second, $\blacktriangleright$ \textbf{[C2] representation}: even when the causal knowledge is available, it may be expressed in non-executable forms (e.g., natural language), requiring the underlying mechanisms to be correctly extracted and composed before the query can be answered. In principle, LLMs could be trained to overcome both challenges. However, $\blacktriangleright$ \textbf{[C3] supervision}: ground-truth interventional outcomes are limited and costly, while individual counterfactual outcomes are  unobservable~\cite{holland1986statistics,hernan2020whatif,pearl2009causality}. This makes scalable direct supervision of causal reasoning difficult, distinguishing it from other modes of reasoning.

\vspace{-0.4cm}
\begin{customblockquote}
    \textbf{Key questions.} How can we get supervision for causal reasoning across scales and representations?
\end{customblockquote}
\vspace{-0.05cm}

\textbf{Increasingly Complex Causal Simulators.} Inspired by recent work leveraging the executability of code for scalable supervision \cite{zhao2025absolutezero}, we argue that causal reasoning needs its own scalable, verifiable training environment. We introduce \textit{increasingly complex causal simulators}: structural causal models (SCMs)\footnote{Structural causal models, introduced by Judea Pearl, are a widely used formalism for representing causal systems in the causal inference literature. We provide a brief overview in Sec.~\ref{subsec:preliminaries}. See \cite{bareinboim_2016, pearl1995causal, pearl2009causality} for more details.}, constructed by LLMs, that are expressible across representations. Our insight is two-fold. First, by formalizing SCMs as code, any causal query can be answered via code execution. Second, by leveraging LLMs' code generation and contextual capabilities, we can build causal simulators \textit{incrementally}, with each modification compilable to ensure causal consistency. This enables complex simulators that would be difficult to construct in a single step. To operate across causal representations, we utilize LLMs to (1) informalize executable representations into non-executable forms, facilitating supervised training in previously unlabeled distributions, and (2) formalize non-executable representations into executable form, enabling data augmentation with both interventional and counterfactual samples. In all, \textbf{$\stairwayupfilled$CauSim} turns causal reasoning from a complex scarce-label problem into a scalable, supervised one.

\textbf{Key Findings.} We evaluate \textbf{$\stairwayupfilled$CauSim} by building and training on causal simulators over meaningless variables to reduce confounding from prior domain knowledge. We first demonstrate that step-by-step construction scales more reliably than one-shot generation, enabling the construction and translation of larger SCMs (Sec.~4). We then study what these causal simulators enable for training (Sec.~\ref{sec:experiments}). \textbf{(1) Generalization:} Training on executable SCMs improves causal reasoning on unseen medical content and non-executable representations, with further gains from additional training on non-executable SCMs over meaningless variables; improvements also transfer to independently constructed external benchmarks. \textbf{(2) Scaling:} Progressive complexity curricula and larger amounts of simulator-generated supervision consistently improve performance, particularly on more complex SCMs. \textbf{(3) Self-improvement:} An LLM can improve its causal reasoning using SCMs that it generates itself, demonstrating one-step bootstrapping from self-generated environments. \textbf{(4) Data augmentation:} Formalizing existing causal knowledge as executable SCMs enables additional supervision that improves performance in the original domain. \textbf{(5) Anchoring performance:} Models trained under \textbf{$\stairwayupfilled$CauSim} outperform larger untrained models from the same family.

\textbf{Contributions.} Our contributions are two-fold: (1) We propose \textbf{$\stairwayupfilled$CauSim}, a novel framework for constructing increasingly complex causal simulators across scales and representations (Sec.~\ref{sec:method}). (2) We systematically study what \textbf{$\stairwayupfilled$CauSim} enables, demonstrating $\blacktriangleright$ generalization to out-of-distribution and non-executable representations, $\blacktriangleright$ consistent gains by curriculum scaling and data volume, $\blacktriangleright$ LLM self-improvement, and $\blacktriangleright$ data augmentation (Sec.~\ref{sec:experiments}).

\vspace{-.75em}
\section{Related work}
\vspace{-.5em}
\label{sec:related_work}

\vspace{-.125cm}

\textbf{Evaluating Causal Reasoning.}
There is an abundant literature on the \textit{evaluation} of LLM causal reasoning, including on textual problems derived from SCMs~\cite{jin2023cladder, jin2023corrfromcaus, chi2024unveiling}, code and mathematical domains~\cite{wang2024causalbench, chen2025counterbench}, and abstract out-of-distribution regimes~\cite{maasch2025causalarc}. Complementary studies consider specific causal reasoning properties: compositional causal inference~\cite{maasch2025compositional}, probabilistic necessity and sufficiency~\cite{gonzalez2024does}, and counterfactual reasoning in hypothetical~\cite{li2023counterfactual} and multimodal settings~\cite{eyescandeceive}. \textit{All of these methods rely on static, human-curated datasets.} Re-Imagine~\cite{xu2025re} is a notable exception that applies mutations to create causal queries from benchmark-derived causal graphs, but they do \textit{not} study scaling graph complexity. In contrast to all of the aforementioned methods, our work is not concerned with evaluation. We focus on improving LLM causal reasoning.

\textbf{Improving Causal Reasoning.} Although there is substantial research on improving \textit{general} reasoning~\cite{cobbe2021training, schulman2015trpo, schulman2017ppo, shao2024deepseekmath, guo2025deepseekr1, wen2025rlvrincentivizes, chen2025sec, astorga2026stepbystep}, improving \textit{causal} reasoning remains limited. Liu et al. \cite{liu-etal-2025-eliciting} demonstrate that prompting LLMs with code significantly improves causal reasoning performance. Hüyük et al. \cite{huyuk2024reasoning} finetune causal reasoning via SFT and DPO on datasets of factual and counterfactual examples over binary variables. More recently, executable code-based SCMs have been explored as a setting for causal supervision \cite{vashishtha2025executable}, though restricted to \textit{fixed}, template-generated graphs. In contrast, our framework exploits executability to incrementally scale causal simulators, enabling globally complex systems with verifiable supervision.

\begin{table}[t]
\centering
\footnotesize
\captionsetup{font=footnotesize}
\caption{Comparison of \textbf{$\protect\stairwayupfilled$CauSim} with related work across key properties. \cmark~indicates supported; \xmark~indicates not supported.
\colorbox{generalGreen}{\strut~Improve Reasoning~}~
\colorbox{myPurple!30}{\strut~Evaluate Causal Reasoning~}~
\colorbox{myBlue!30}{\strut~Improve Causal Reasoning~}}
\label{tab:related_work}

\scalebox{0.925}{%
\begin{tabular}{!{\color{white}\vrule width 4pt}lcccccc}
\toprule
\textbf{Method} & \textbf{Causal} & \textbf{Scalable} &
\textbf{Executable} & \textbf{Training} & \textbf{Formalize} &
\textbf{Informalize}\\
\midrule
\cellcolor{generalGreen}
Absolute Zero~\cite{zhao2025absolutezero}
& \xmark & \cmark & \cmark & \cmark & \xmark & \xmark \\
\midrule
\cellcolor{myPurple!30}
CLadder~\cite{jin2023cladder}
& \cmark & \xmark & \xmark & \xmark & \xmark & \xmark\\
\cellcolor{myPurple!30}
Re-Imagine~\cite{xu2025re}
& \cmark & \xmark & \cmark & \xmark & \cmark & \cmark\\
\midrule
\cellcolor{myBlue!30}
Hüyük et al.~\cite{huyuk2024reasoning}
& \cmark & \xmark & \xmark & \cmark & \xmark & \xmark\\
\cellcolor{myBlue!30}
Exec. Counterfactuals~\cite{vashishtha2025executable}
& \cmark & \xmark & \cmark & \cmark & \xmark & \xmark\\
\cellcolor{myBlue!30}
\textbf{$\stairwayupfilled$CauSim. (Ours)}
& \cmark & \cmark & \cmark & \cmark & \cmark & \cmark\\
\bottomrule
\end{tabular}%
}

\vspace{-.225cm}
\end{table}
\vspace{-.125cm}

\textbf{Improving Reasoning through LLM-Generated Data.} A body of work explores using LLMs to generate training data and environments to improve reasoning performance~\cite{yuan2024selfrewarding,xin2024deepseekproverv15harnessingproofassistant,zhao2025absolutezero}. In particular, Absolute Zero~\cite{zhao2025absolutezero} proposes and solves tasks to maximize learning progress. We build on this scalable, environment-centric body of work, but specifically target causal reasoning by having LLMs generate increasingly complex causal simulators as verifiable environments. To our knowledge, this is the first work to apply LLM-generated environments to improve causal reasoning.

\vspace{-.75em}
\section{Formalism and Background}
\vspace{-.75em}
Our framework is designed to build executable supervision across \textit{scales} and \textit{representations}. For clarity, we first formalize the problem of answering causal queries (\S\ref{subsec:preliminaries}) and outline the key challenges this formalization reveals (\S\ref{subsec:challenges}).
\vspace{-.75em}
\subsection{Problem Setting}
\vspace{-.75em}
\label{subsec:preliminaries}
We consider the task of answering a causal query $q$ about an underlying data generating process (DGP). We adopt Pearl's SCM framework as the underlying DGP.
\begin{definition}[Structural Causal Model (SCM)~\cite{bareinboim_2016}]
A (SCM) is a tuple: $\mathcal{M} \coloneqq \langle \mathbf{U}, p(\mathbf{u}), \mathbf{V}, \mathbf{F} \rangle$, where $\mathbf{U}=\{U_i\}_{i=1}^{m}$ is a set of exogenous variables distributed according to $p(\mathbf{u})$, $\mathbf{V}=\{V_i\}_{i=1}^{n}$ is a set of endogenous variables, and $\mathbf{F}=\{f_i\}_{i=1}^{n}$ is a collection of structural assignments such that each $V_i$ is generated as: $v_i = f_i(\mathrm{pa}_i, u_i)$, where $\mathrm{pa}_i \subseteq \mathbf{V}\setminus\{V_i\}$ denotes the endogenous parents of $V_i$, and $u_i$ denotes the corresponding exogenous context.
\end{definition}

As in~\cite{vashishtha2025executable}, the SCM is provided as input to the LLM. In our work, we make the insight that the same SCM $\mathcal{M}$ may be expressed in substantially different representations $R \in \mathcal{R}$, for instance as natural language descriptions or domain-specific diagrams. We define a representation $R(\mathcal{M})$ as a complete encoding of the SCM $\mathcal{M}$. A representation is \textit{executable} if $R(\mathcal{M})$ can be directly compiled to yield answers to any causal query; otherwise it is \textit{non-executable}. Our framework operates over both, with translation between them defined in \S\ref{sec:method}. Our primary interest is in how reasoning difficulty scales with increasing $n$ and varying representations $\mathcal{R}$.

\textbf{Types of Causal Queries.} We consider four types of causal queries over SCMs. At the core are two primitive processes: \textit{Deduction}: predicting endogenous variables $\mathbf{V}$ from exogenous variables $\mathbf{U}$ via forward propagation through $\mathbf{F}$, and \textit{Abduction}: inferring exogenous values $\mathbf{U}$ from observed endogenous values $\mathbf{V}$. Building on these, \textit{Intervention} is a modified form of deduction in which one or more endogenous variables are externally set via the $\mathrm{do}(\cdot)$ operator. \textit{Counterfactual} queries combine both: abduction first infers exogenous variables consistent with observed outcomes, which are then propagated forward through the intervened SCM. Formal definitions in Appendix~\ref{app:data_generation}.
\vspace{-0.35cm}
\begin{customblockquote}
    \textbf{The Goal.} Let $\mathbf{v}$ denote an observation of endogenous variables $\mathbf{V}$, and $q$ a causal query over $\mathcal{M}$. Our goal is to train an LLM $\pi_\theta$ to answer $q$ over SCMs of varying scale $n$ and representations $R \in \mathcal{R}$ given $R(\mathcal{M})$, by minimizing a loss between its prediction $\hat{a} = \pi_\theta(R(\mathcal{M}), q, \mathbf{v})$ and the ground-truth answer $a$: $\min_{\theta} \; \mathbb{E}\!\left[ \mathcal{L}(\hat{a}, a) \right]$
\end{customblockquote}
\vspace{-0.125cm}
To update the LLM parameters $\theta$, we use reward-based bootstrapping, with the reward determined by the correctness of the predicted answer $\hat{a}$. Other forms of supervision could also be used.

\vspace{-1em}
\subsection{Challenges.}
\vspace{-1em}

\label{subsec:challenges}
The above goal formulation identifies three aspects that make causal reasoning difficult:

\textbf{[C1] System Scale, $n$.}
As the size of the SCM increases, answering causal queries becomes increasingly difficult. Consider the causal system in Fig.~\ref{fig:framework}, where Drug $Y$ activates Blocker $Z$ across a long network pathway to suppress Tumor $X$. To understand the effect of an intervention on $Y$, one must trace how $Y$ influences $Z$, then determine how activation of $Z$ affects $X$. Even though all interactions are fully specified, reasoning through these sequential and interacting steps is nontrivial.

\textbf{[C2] Representation, $R$}.
Some representations may be much easier to interpret than others. For example, a causal mechanism written as a formal logical program may be easier to interpret than the same mechanism described in a lengthy paragraph. Effective causal reasoning within any representation requires identifying the actual mechanisms from superfluous content, composing them correctly, and reasoning over the resulting complete structure.

\textbf{[C3] Limited Supervision, $a$.} In principle, challenges \textbf{[C1-2]} could be addressed with sufficient ground-truth data using training techniques. The problem is that this ground-truth data does not exist for causal reasoning. Interventional data is limited and costly to obtain, and counterfactual outcomes are inherently unobservable. As a result, the ground-truth answers required to directly supervise the causal queries described in Section~\ref{subsec:preliminaries} may not (and likely do not) exist in real world settings.
\vspace{-0.35cm}

\begin{customblockquote}
    \textbf{The Core Challenge.} Together, these challenges point to available data as a limitation in improving causal reasoning. LLMs may struggle to reason over large systems and varying representations, and the ground-truth supervision that could be used to train them is scarce.
\end{customblockquote}

\vspace{-.75em}
\section{The Method: Building Increasingly Complex Causal Simulators}
 \vspace{-.75em}

\label{sec:method}
The limitations of available data motivate our research question: \textit{if we cannot use existing data to train LLMs across the full space of causal queries, can we simulate the data we need?}

\textbf{Overview.} Our method, \textbf{$\stairwayupfilled$CauSim}, uses LLMs to construct increasingly large SCMs as executable Python programs (Sec.~\ref{subsec:scaling_scm}). Since real-world causal knowledge is usually not expressed in executable form, our framework: (1) translates executable SCMs into non-executable representations, which provides verifiable supervision in previously unsupervisable settings, and (2) translates non-executable representations into executable SCMs, enabling existing causal knowledge to be incorporated into the data-generation process (Sec.~\ref{subsec:translate-scm}). This produces supervised data across scales \textbf{[C1]} and representations \textbf{[C2]}, with ground-truth supervision across all causal query types \textbf{[C3]} (Sec.~\ref{subsec:causal-query-generation}).

Motivated by recent work on LLM-generated training environments (Sec.~\ref{sec:related_work}), our framework uses LLM-based SCM construction to jointly support scalable simulator growth and translation across causal representations. Further discussion of this design choice is provided in \ref{app:llm_vs_procedural}.

\vspace{-.75em}
\subsection{Scaling Structural Causal Models}
\begin{figure}[H]
    \centering
    \vspace{-0.25cm}
\includegraphics[width=.8\textwidth]{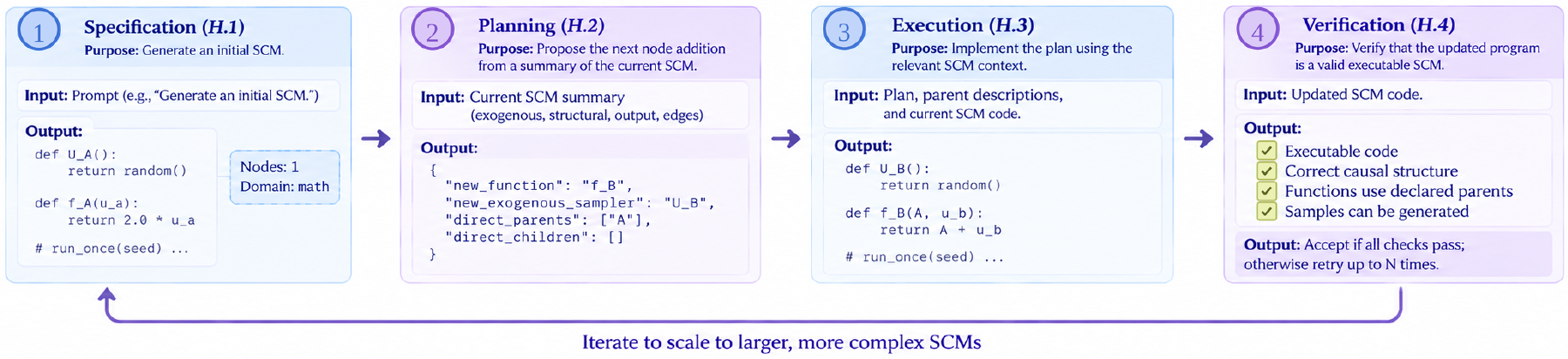}
    \vspace{-0.25cm}
    \captionsetup{font=footnotesize}
    \caption{\textbf{$\protect\stairwayupfilled$CauSim (executable component)} uses LLMs to generate SCMs incrementally, node-by-node, starting from a small initial SCM. At each stage, context is restricted: planning uses only a high-level semantic summary of the current SCM, while execution receives only the code for affected child nodes and descriptions of relevant parents. Planning, execution, and verification repeat until the target size is reached.}
    \label{fig:framework2}
    \vspace{-0.25cm}
\end{figure}

\vspace{-.75em}
\label{subsec:scaling_scm}
As $n$ increases, it becomes increasingly difficult for LLMs to constuct SCMs that maintain fidelity to causal assumptions and preserve code executability (see Experiment 1 below). We address this by constructing executable SCMs \textit{incrementally} through a four-stage procedure: $\blacktriangleright$ \textbf{specification}, $\blacktriangleright$ \textbf{planning}, $\blacktriangleright$ \textbf{execution}, and $\blacktriangleright$ \textbf{verification}.  In the interest of space, we provide a visual overview in Fig.~\ref{fig:framework2}
and defer the full details and walk-through example to Appendix~\ref{app:scm_construction}.

To validate the choice of incremental construction of SCMs, we compare incremental SCM generation versus one-shot generation at increasing scales in \hyperref[exp:inc_vs_one]{\textsc{Experiment 1}}.

\vspace{-.75em}
\subsubsection*{\textsc{Experiment 1. Does Incremental SCM Generation Enable Scalability?}}
\label{exp:inc_vs_one}

\begin{wrapfigure}[17]{r}{0.35\columnwidth}
     \vspace{-1.95em}
     \centering
    \includegraphics[width=1.\linewidth]{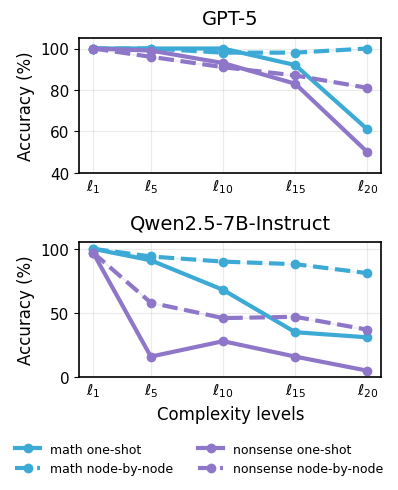}
    \vspace{-1.25em}
    \captionsetup{font=footnotesize}
    \caption{\textbf{Node-by-node generation improves SCM generation \textit{at scale}.}}
    \label{fig:scm_scaling_success}
\end{wrapfigure}

\textbf{Method.} We compare two methods for constructing an SCM with $n$ structural mechanisms and $n$ corresponding exogenous samplers as executable \textit{code}: (i) \textit{One-shot generation}: the LLM is prompted to output the full SCM in a single specification step and (ii) \textit{Incremental generation}: the LLM first specifies a 1-node SCM, and then incrementally grows by adding a single node and its corresponding exogenous node at a time using our planning - execution loop.

\textbf{Experimental Setup.} We evaluate these two approaches on two domains represented in executable format: a symbolic \textsc{math} domain and a \textsc{nonsense} domain, consisting of variables of 4-letter nonsensical strings (a design choice we discuss in Sec.~\ref{sec:experiments}). The full prompts used to generate SCMs in each domain are provided in Appendix~\ref{appendix:prompts} and examples of \textsc{math} and \textsc{nonsense} SCMs are shown in Appendices~\ref{app:construction_example} and~\ref{app:nonsense_code_example}. To study if generating increasingly large SCMs is challenging across models of different capacities, we test both \textsc{GPT-5} and \textsc{Qwen-2.5-7B-Instruct}. Because \textsc{Qwen-2.5-7B-Instruct} is less consistent in following instructions, we allow up to three attempts per model call to meet verification criteria, whereas \textsc{GPT-5} is allowed a single attempt. Accuracy is measured as the fraction of 100 trials that pass all verification checks.

\textbf{Results.}
Figure~\ref{fig:scm_scaling_success} compares one-shot and incremental SCM generation performance as a function of SCM size. Across domains and LLMs, one-shot generation degrades more rapidly as $n$ increases than our node-by-node approach.

\vspace{-0.5cm}
\begin{customblocktakeaway}
    \checkmark\;\textbf{Takeaway.} Incremental SCM construction scales more reliably than one-shot generation, enabling the generation of increasingly large SCMs while preserving structural validity.
\end{customblocktakeaway}

\vspace{-.5em}
\subsection{Translating Structural Causal Models}
\vspace{-.5em}
\label{subsec:translate-scm}

In many settings, causal queries are expressed in \textit{non-executable} representations---for example, natural-language descriptions that cannot be directly compiled or executed \textbf{[C2]}. To operate in these settings, \textbf{$\stairwayupfilled$CauSim} uses two complementary translation steps. First, \textit{informalization} translates an executable SCM into a semantically equivalent non-executable representation, approximating those encountered in practice while retaining an executable counterpart that provides ground-truth answers to causal queries. Second, \textit{formalization}, analogous to optimization model autoformulation~\cite{astorga2025autoformulation}, translates causal knowledge from a non-executable representation into an executable SCM, allowing this knowledge to initialize a simulator that can then be augmented using our incremental construction procedure. Together, these steps enable a \textit{formalization} $\rightarrow$ \textit{augmentation} $\rightarrow$ \textit{informalization} pipeline for generating verifiable supervision from existing causal knowledge.\footnote{Formalization is distinct from causal reasoning: it constructs an initial executable SCM from a source representation, which can subsequently be modified or expanded using the incremental construction procedure in Sec.~4.1.} Further implementation details and prompts are provided in Appendix~\ref{sec:prompt}.

\textbf{Summary.}
Building on our four-stage incremental approach, we formalize non-executable causal knowledge \textit{step by step}, restricting each planning step to parent and child relationships explicitly supported by the source representation. Conversely, we informalize executable SCMs into non-executable representations. As a proxy for verifying semantic preservation, we formalize the resulting representation back into an executable SCM and confirm that its semantic context and execution remain consistent. We evaluate step-by-step formalization against a one-shot approach below.

\vspace{-0.25cm}
\subsubsection*{\textsc{Experiment 2: Can Non-Executable Knowledge Become Executable SCMs?}}
\label{exp3:external}

\textbf{Method.} We evaluate our formalization procedure across different source representations. The NIH Stroke Scale (NIHSS) is a natural-language, point-based clinical scoring system for stroke severity, where individual items measure neurological functions.
The Bayesian networks ASIA and CANCER from \texttt{bnlearn}~\cite{scutari2010learningbayesiannetworksbnlearn} use conditional probability tables to specify the distribution of each variable given its parents. Examples of the source representations and corresponding executable SCMs are provided in Appendices~\ref{app:nihss_formalization} and~\ref{app:bn_formalization}.

\begin{wraptable}[6]{r}{0.32\textwidth}
\vspace{-.96\baselineskip}
\centering

\scalebox{0.75}{%
\begin{tabular}{lcc}
\toprule
\textbf{Dataset} & \textbf{Step-by-step} & \textbf{One-shot} \\
\midrule
NIHSS  & \textbf{52/100} & 20/100 \\
ASIA   & \textbf{84/100} & 69/100 \\
CANCER & \textbf{74/100} & 51/100 \\
\bottomrule
\end{tabular}%
}
\vspace{-.5\baselineskip}
\captionsetup{font=footnotesize}
\caption{\textbf{Node-by-node outperforms one-shot formalization.}}
\label{tab:nihss_accuracy}
\vspace{-1\baselineskip}
\end{wraptable}
\textbf{Experimental Setup.} We use \textsc{GPT-5} to generate a reference SCM from these dataset and manually verify its correctness against the original representation. This reference SCM is used to measure the accuracy of the generated SCMs. We then prompt \textsc{Qwen-2.5-7B-Instruct} to formalize the SCM either in (i) one-shot or (ii) incrementally using our planning-execution loop. Accuracy is measured as the fraction of 100 trials that pass all verification checks and match the causal structure of the GPT-5 generated SCM. All prompts are shown in Appendix~\ref{app:translation_prompts}.

\textbf{Results.} Table~\ref{tab:nihss_accuracy} reports the accuracy of one-shot versus incremental SCM construction from the NIHSS. Node-by-node generation significantly outperforms one-shot generation.

\vspace{-0.25cm}
\begin{customblocktakeaway}
    \checkmark\;\textbf{Takeaway.} Node-by-node formalization of external knowledge into executable SCMs outperforms the one-shot method.
\end{customblocktakeaway}

\vspace{-0.25cm}
\subsection{Causal Query Generation and Answer Verification}
\vspace{-0.125cm}
\label{subsec:causal-query-generation}

\textbf{Query Generation.} After generating the SCMs with our framework, we sample causal queries and execute the SCMs to obtain ground-truth answers. Because our goal is to measure and improve causal reasoning itself, we use discrete $\mathbf{V}$ and $\mathbf{U}$ to avoid conflating causal reasoning with the additional mathematical skills required in continuous probabilistic inference. Instead of predicting only the queried target—which risks exploiting the spurious effect of correctly guessing a single target node—the training LLMs predict all endogenous variables for deduction, intervention, and counterfactual queries; all exogenous variables for abduction queries; and the unobserved subset of $\mathbf{U}$ for counterfactual queries.

\textbf{Answer Verification.} Since structural mechanisms may not be invertible, the same observation may be compatible with multiple assignments of $\mathbf{U}$. We therefore enumerate candidate assignments up to a fixed budget and retain those that reproduce the observation. We apply the intervention and execute the modified SCM under each compatible assignment, producing the set of valid counterfactual outcomes. A prediction is considered correct if it matches any outcome in this set. Additional details and discussion of these design choices are provided in Appendix~\ref{app:discrete_scms}.

\vspace{-0.25cm}
\section{Experiments}
\label{sec:experiments}
\vspace{-0.25cm}

The main claim of our work is that verifiable supervision from scalable causal simulators can improve LLM causal reasoning over scales and representations. We evaluate this claim through five complementary studies:

\begingroup
\setlength{\arrayrulewidth}{1.5pt}
\setlength{\tabcolsep}{5pt}
\renewcommand{\arraystretch}{1.225}
\arrayrulecolor{myPurple}

\par\noindent
\makebox[\linewidth][c]{%
\resizebox{0.75\linewidth}{!}{%
\begin{tabular}{|l|l|}
\hline
\rowcolor{myPurple}
\textcolor{white}{\textbf{Section}} &
\textcolor{white}{\textbf{Goal}} \\
\hline

\S\ref{generalization} Generalization &
Transfer causal reasoning across unseen domains and representations \\
\hline

\S\ref{scaling} Scaling &
Leverage scalable simulators to study curriculum and data scaling \\
\hline

\S\ref{self-improvement} Self-Improvement &
Improve reasoning through self-generated causal simulators \\
\hline

\S\ref{case-study} Data Augmentation &
Formalize real-world causal knowledge for data augmentation \\
\hline

\S\ref{sec:absolute-performance} Performance Anchoring &
Contextualize accuracy against larger and frontier reference models \\
\hline

\end{tabular}%
}%
}
\par

\arrayrulecolor{black}
\endgroup
\textbf{General Setup.} Unless otherwise noted, the following setup is held constant across experiments.
\textbf{Data.} We construct executable SCMs with four-letter variable names (\textsc{nonsense}) and medical concepts (\textsc{medical}) using \texttt{GPT-5}. We expand each SCM node-by-node up to 10 endogenous variables and retain every incremental version. These executable \textit{code} SCMs are then converted into natural-language (\textit{nl}) representations. Unless otherwise stated, training uses \textsc{nonsense}-\textit{code}. \textsc{Medical}-\textit{code} and \textsc{medical}-\textit{nl} are never trained on and solely used for evaluation.

\textbf{Training.} We fine-tune \texttt{Qwen2.5-3B-Instruct}, which we refer to as the \textsc{base model}. We bootstrap causal reasoning using a STaR/RFT/RAFT-style rejection-sampling procedure~\citep{zelikman2022star,yuan2023scalingrelationship,dong2023raft}: for each causal query, the model generates a solution, which we verify against the ground-truth answer obtained by executing the SCM. Only correct solutions are retained for supervised fine-tuning. Training curricula use SCM sizes $l_2$, $l_4$, $l_6$, and $l_8$, where $l_k$ denotes the incremental SCM version containing $k$ endogenous variables. The curriculum and amount of training data vary by experiment which we detail below.

\textbf{Evaluation.} For our synthetic datasets, we use held-out \textsc{nonsense}-\textit{code}, \textsc{medical}-\textit{code}, and \textsc{medical}-\textit{nl} SCMs. Each contains 6,000 intervention and counterfactual queries spanning 20 SCM IDs per topology, 10 incremental scales, and five queries per type. External benchmarks and the NIHSS augmentation study use separate evaluation sets, described in their respective sections.  Examples and full dataset details are provided in Appendices~\ref{app:dataset_examples} and~\ref{app:experiment_configurations}. We use Pass@1 as the primary metric and report additional Pass@$k$ and majority-vote results in Appendix~\ref{app:passk_majority}.

\vspace{-.75em}
\subsection{Improving Causal Reasoning Across Distributions}
\label{generalization}
\vspace{-.75em}

\begin{figure*}[t]
    \centering

    \begin{subfigure}{\textwidth}
        \centering
        \includegraphics[width=\textwidth]{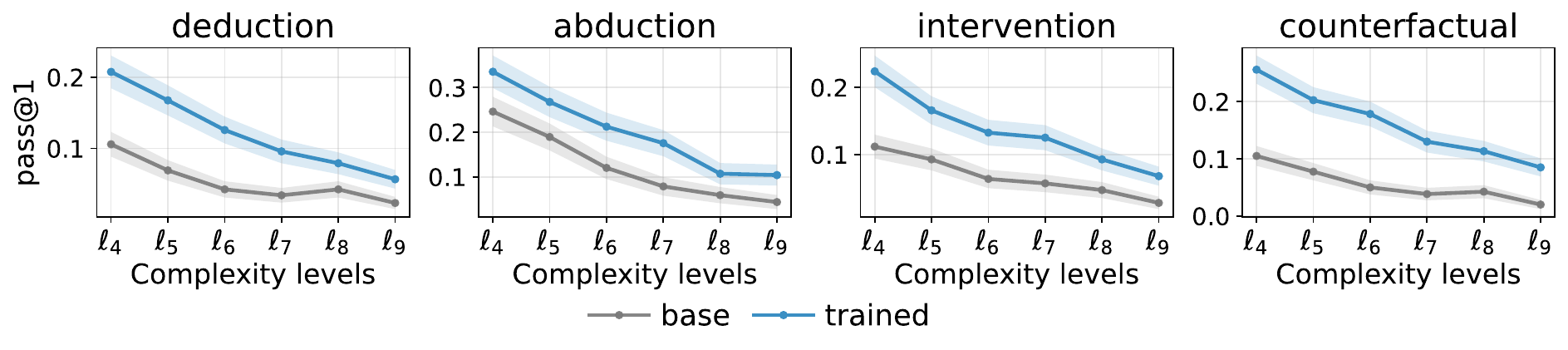}
        \vspace{-2.0em}
        \captionsetup{font=footnotesize}
        \caption{In-distribution causal reasoning on unseen SCMs}
        \label{fig:in_distribution}
    \end{subfigure}
    \begin{subfigure}{0.23\textwidth}
        \centering
        \includegraphics[width=\textwidth]{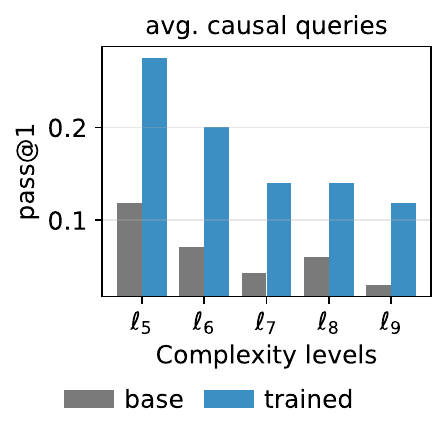}
        \vspace{-2.0em}
        \captionsetup{font=footnotesize}
        \caption{Domain Shift}
        \label{fig:gen_medical)formal}
    \end{subfigure}
    \hfill
    \begin{subfigure}{0.42\textwidth}
        \centering
        \includegraphics[width=\textwidth]{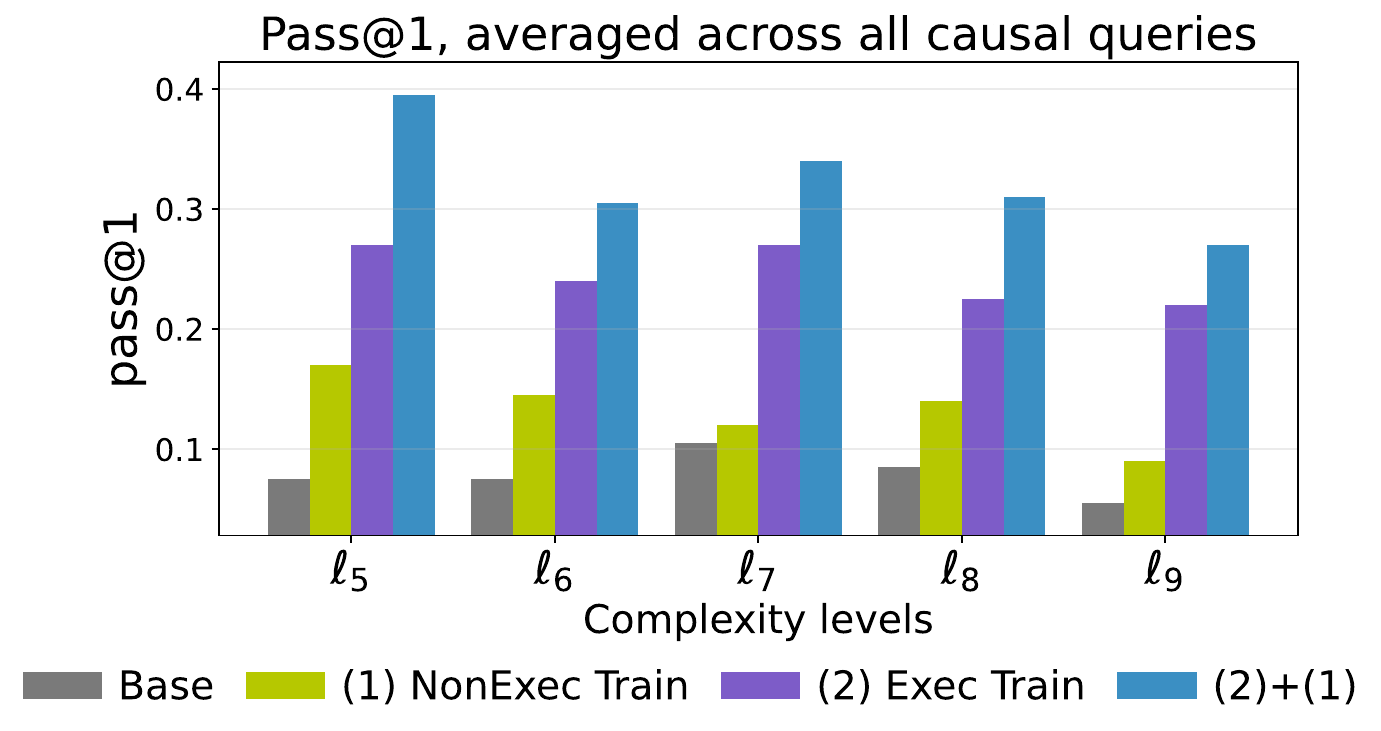}
        \vspace{-2.0em}
        \captionsetup{font=footnotesize}
        \caption{Domain \& Representation Shift }
        \label{fig:gen_medical}
    \end{subfigure}
    \hfill
    \begin{subfigure}{0.24\textwidth}
        \centering
    \includegraphics[width=\textwidth]{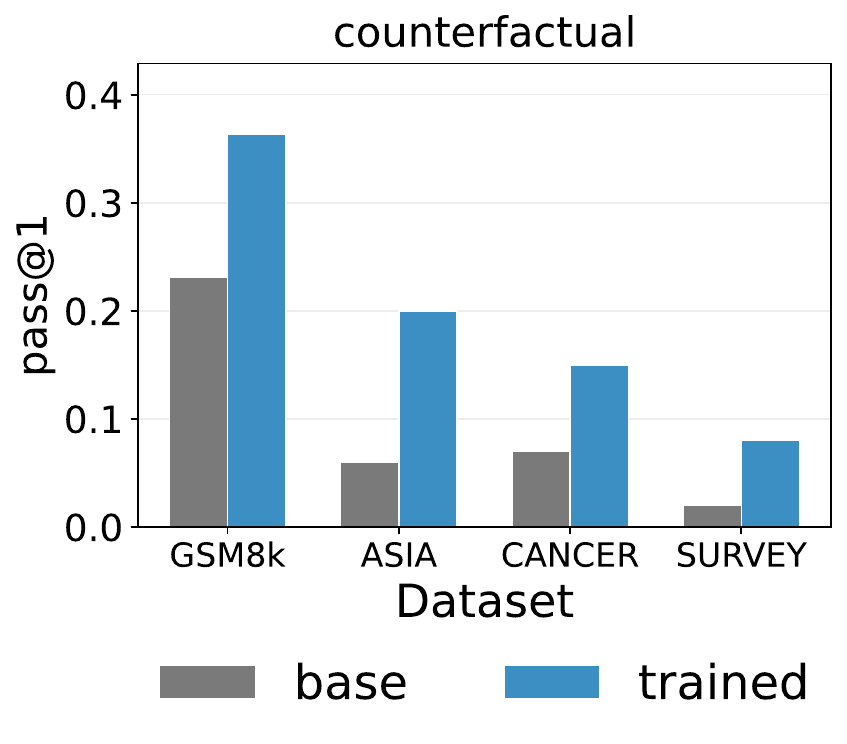}
        \vspace{-1.7em}
        \captionsetup{font=footnotesize}
        \caption{External benchmarks}
        \label{fig:gen_external}
    \end{subfigure}
    \vspace{-.5em}
    \captionsetup{font=footnotesize}
    \caption{\textbf{Improving Causal Reasoning Across Distributions}
    }
    \label{fig:mega_generalization}
\vspace{-.50cm}
\end{figure*}

\textbf{Motivation.} We first ask whether training on executable causal simulators improves causal reasoning beyond the training distribution. We evaluate progressively larger shifts in semantics and representation, followed by evaluation on independently constructed external benchmarks.

\textbf{Method.} Unless otherwise stated, we evaluate the best-performing \textsc{nonsense}-\textit{code} model from our scaling experiments (Sec.~\ref{scaling}), trained with the \textsc{Block} curriculum and 135 queries per type. We compare it against the \textsc{base model} across four evaluation settings:
$\blacktriangleright$ \textbf{In-Distribution:} held-out \textsc{nonsense}-\textit{code} SCMs;
$\blacktriangleright$ \textbf{Domain Shift:} \textsc{medical}-\textit{code} SCMs, which change the semantic content while preserving the code representation;
$\blacktriangleright$ \textbf{Domain \& Representation Shift:} \textsc{medical}-\textit{nl} SCMs, which change both semantics and representation; and
$\blacktriangleright$ \textbf{External Benchmarks:} independently constructed causal problems derived from GSM8K~\cite{xu2025re} and the ASIA, CANCER, and SURVEY Bayesian networks from \texttt{bnlearn}~\cite{scutari2010learningbayesiannetworksbnlearn}.

For the domain and representation shift experiment, we test how the training representation affects transfer. We compare three models: (i) one trained on \textsc{nonsense}-\textit{code}, (ii) one trained directly on \textsc{nonsense}-\textit{nl}, and (iii) the \textsc{nonsense}-\textit{code} model further fine-tuned on \textsc{nonsense}-\textit{nl}. All three are evaluated on the same held-out \textsc{medical}-\textit{nl} set. To further evaluate generalization, we evaluate the \textsc{nonsense}-\textit{code} model on partially specified, verbalized GSM8K SCMs in Appendix~\ref{app:partial_gsm8k}.

\textbf{Results.}
$\blacktriangleright$ \textbf{In-Distribution:} The fine-tuned model consistently outperforms the \textsc{base model} on held-out \textsc{nonsense}-\textit{code} SCMs (Fig.~\ref{fig:in_distribution}). $\blacktriangleright$ \textbf{Domain Shift:} These gains persist on \textsc{medical}-\textit{code} SCMs (Fig.~\ref{fig:gen_medical)formal}). $\blacktriangleright$ \textbf{Domain \& Representation Shift:} Improvements also transfer to \textsc{medical}-\textit{nl} SCMs (Fig.~\ref{fig:gen_medical}). The model trained on \textsc{nonsense}-\textit{code} outperforms the model trained directly on \textsc{nonsense}-\textit{nl}, and further fine-tuning the code-trained model on \textsc{nonsense}-\textit{nl} yields the strongest performance. This suggests that training on executable representations provides a foundation that transfers across representations, while representation-specific training provides additional gains. $\blacktriangleright$ \textbf{External Benchmarks:} The \textsc{nonsense}-\textit{code} model also improves performance on the independently constructed external benchmarks (Fig.~\ref{fig:gen_external}). We observe a similar improvement when evaluating on partially specified verbalized GSM8K SCMs (Table~\ref{tab:extremely_informal}).

\vspace{-.25cm}
\begin{customblocktakeaway}
    \checkmark\;\textbf{Takeaway.} Training on executable causal queries improves LLM causal reasoning on unseen SCMs and generalizes across semantic and representation shifts.
\end{customblocktakeaway}

\vspace{-.75em}
\subsection{Scaling Causal Reasoning via Complexity and Dataset Size}
\vspace{-.75em}
\label{scaling}

\textbf{Motivation.} A benefit of our framework is that we can scale both the complexity of the SCMs and the amount of data they generate. We use the this to study two questions: if progressively increasing SCM complexity improves learning, and if causal reasoning improves with more training data.

\begin{wrapfigure}[16]{R}{0.44\columnwidth}
  \centering
  \vspace{-1.5em} %

  \includegraphics[width=\linewidth]{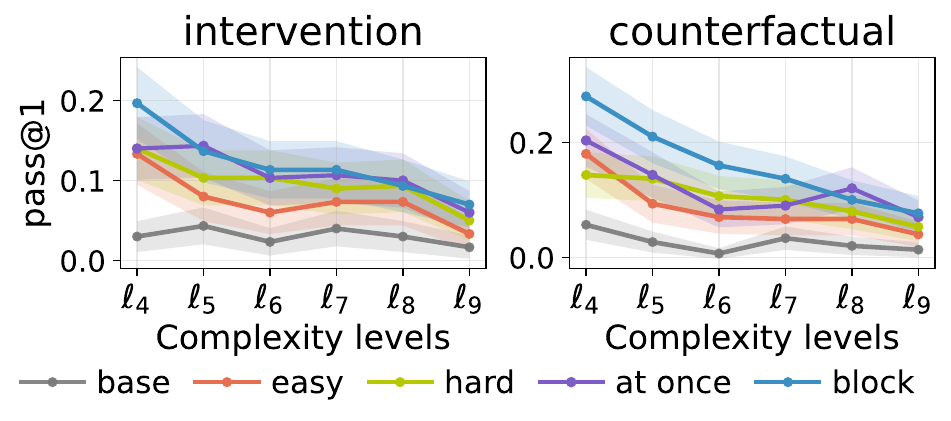}
  \vspace{-2em}
  \captionsetup{font=footnotesize}
  \caption{\textbf{Curriculum Matters.}}
  \label{fig:acc1-rate-marg-struct-by-query}

  \includegraphics[width=\linewidth]{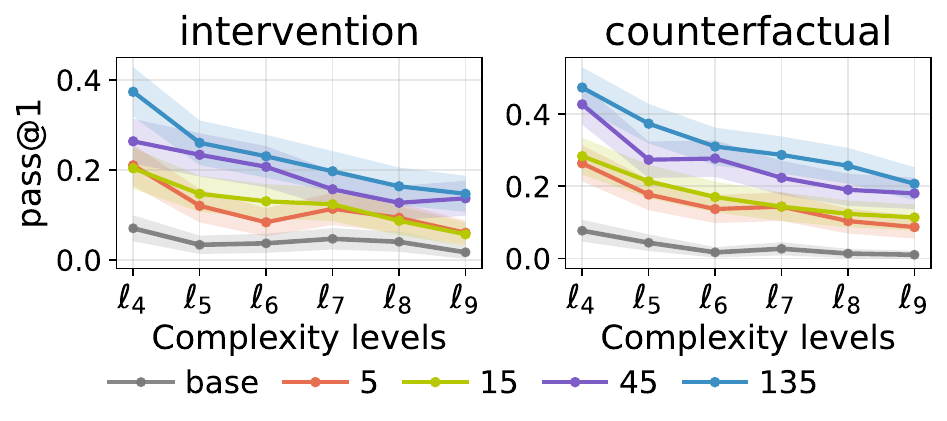}
  \vspace{-2em}
  \captionsetup{font=footnotesize}
  \caption{\textbf{More Data = Improved Performance}}
  \label{fig:acc1-rate-dataamount-marg-struct-by-query}
\end{wrapfigure}
\textbf{Method.} $\blacktriangleright$ \textbf{Curriculum:} In \textsc{Block}, training proceeds iteratively: at each SCM size $l_k$, the current model generates solutions to causal queries, is fine-tuned on the correct solutions, and the resulting model is then used to generate solutions at the next level. In contrast, \textsc{At-Once} uses the initial model to generate solutions for all complexity levels upfront, after which fine-tuning proceeds on the correct solutions through the levels in increasing order. \textsc{EasyOnly} trains only on $l_2$, and \textsc{HardOnly} trains only on $l_8$. To isolate the effect of curriculum, all four settings use the same total budget of 64,800 intervention and counterfactual training queries. $\blacktriangleright$\textbf{Dataset Size.} We then fix the \textsc{Block} curriculum and vary the amount of generated supervision, training separate models with 5, 15, 45, and 135 queries per type at each SCM version. All models are evaluated on the same held-out \textsc{nonsense}-\textit{code} set described in the General Setup.

\textbf{Results.} $\blacktriangleright$ \textbf{Curriculum Matters.} (Fig.~\ref{fig:acc1-rate-marg-struct-by-query}) The \textsc{Block} curriculum outperforms all other schemes. This confirms that progressively increasing SCM size improves learning.
$\blacktriangleright$ \textbf{More Data = Improved Performance.} (Fig.~\ref{fig:acc1-rate-dataamount-marg-struct-by-query}) Scaling up the number of causal queries during training consistently yields higher accuracy across query types. See App~\ref{app:curriculum_ablation} \&~\ref{app:data_amount_ablation} for additional results. We perform an ablation varying the number of distinct SCMs versus the number of queries, see App.~\ref{app:scm_query_ablation}.

\vspace{-0.75em}
\begin{customblocktakeaway}
\checkmark\textbf{Takeaway.} Causal reasoning improves with a progressive curriculum over increasingly complex SCMs and greater amounts of generated supervision.
\end{customblocktakeaway}

\vspace{-.75em}
\subsection{Self-Improvement}
\vspace{-.85em}
\label{self-improvement}
\begin{wrapfigure}[12]{r}{0.35\textwidth}
    \centering
    \vspace{-1.4\baselineskip} %
    \includegraphics[width=\linewidth]{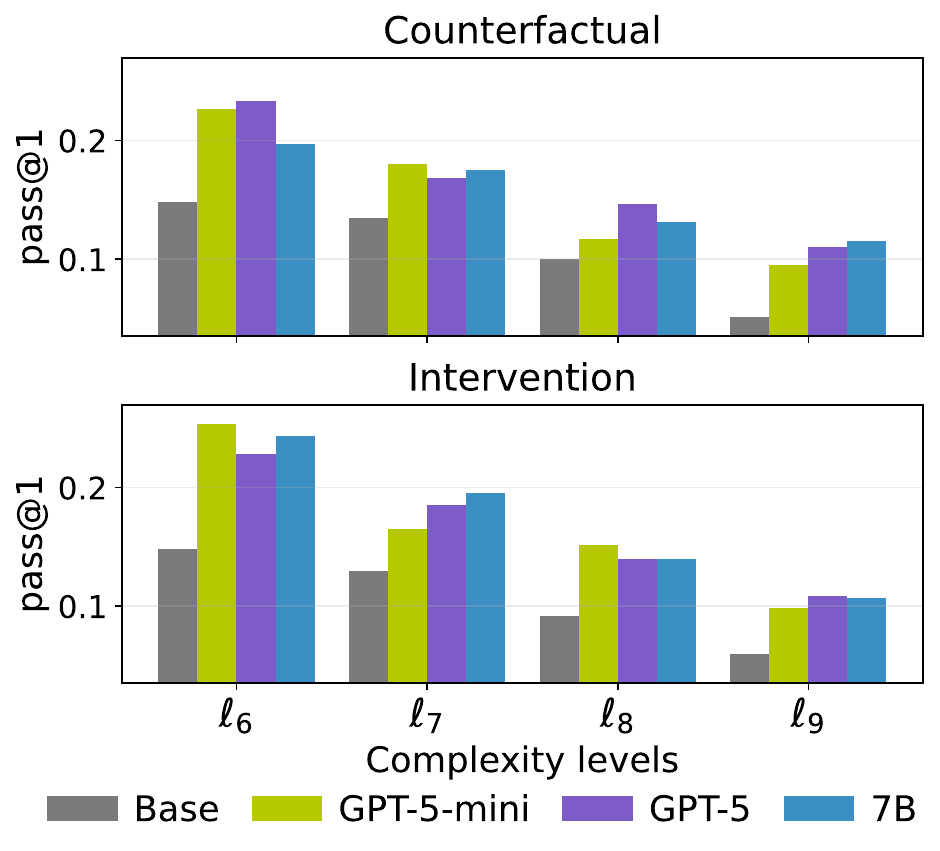}
    \captionsetup{font=footnotesize}
    \caption{\textbf{Self-Improvement}}
    \label{fig:generator-ablation}
    \vspace{-0.5em} %
\end{wrapfigure}
\textbf{Motivation.}
We next ask whether an LLM can bootstrap its causal reasoning ability using SCMs that it generates itself, and how the effectiveness of this self-generated data compares with data generated by stronger models.

\textbf{Method.}
We construct three \textsc{nonsense}-\textit{code} datasets using \texttt{GPT-5}, \texttt{GPT-5-mini}, and \texttt{Qwen2.5-7B-Instruct}. For each generator, we produce 180 SCM IDs and train \texttt{Qwen2.5-7B-Instruct} using the \textsc{Block} curriculum with 15 intervention and 15 counterfactual queries per level. In the final setting, the same \texttt{Qwen2.5-7B-Instruct} model therefore serves as both the SCM generator and the model being improved. All three models are evaluated on the common held-out \textsc{nonsense}-\textit{code} test set described above.

\textbf{Results.}
All three generator sources yield clear improvements over the corresponding base 7B model(Fig.~\ref{fig:generator-ablation}). Performance is also similar across SCMs generated by \texttt{GPT-5}, \texttt{GPT-5-mini}, and \texttt{Qwen2.5-7B-Instruct} itself. Thus, within the scale, representation, and domain studied here, self-generated SCMs provide sufficient training signal for the model to improve its own causal reasoning. Additional results are provided in Appendix~\ref{app:7b-comparisons}.

\vspace{-0.75em}
\begin{customblocktakeaway}
    \checkmark\;\textbf{Takeaway.} An LLM can improve its causal reasoning using SCM that it generates itself.
\end{customblocktakeaway}

\vspace{-.75em}
\subsection{Case-Study: Augmenting Existing Data}
\vspace{-.75em}
\label{case-study}

\textbf{Motivation.}
A benefit of our framework is that if there exits external knowledge thatcan be formalized as an executable SCM, the resulting simulator can generate additional interventional and counterfactual training data beyond what is originally available.

\textbf{Method.}
We use the NIH Stroke Scale (NIHSS) as a case study and formalize its clinical scoring rules into an executable SCM using our procedure from Sec.~\ref{subsec:translate-scm}. We then compare two training-data regimes over SCM sizes: a \textit{Small} dataset with 50 intervention and 50 counterfactual queries per size (400 queries total), and an \textit{Augmented} dataset with 1,000 queries per type per size (8,000 queries total). Both models use the same \textsc{Block} training procedure and are evaluated on a separate held-out NIHSS set spanning SCM sizes $l_1$--$l_{10}$ with 400 queries per type per size.

\begin{wrapfigure}[6]{r}{0.35\textwidth}
    \centering
    \vspace{-0.45cm} %
    \includegraphics[width=\linewidth]{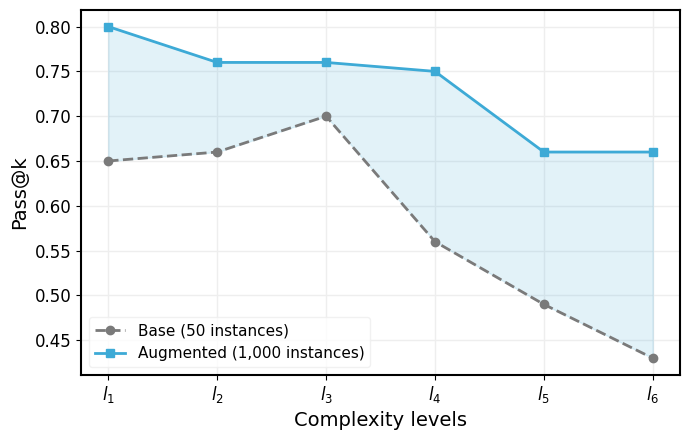}
    \vspace{-0.6cm}
    \captionsetup{font=footnotesize}
    \caption{\textbf{Data Augmentation}}
    \label{fig:nihss-augmentation}
    \vspace{-0.4cm} %
\end{wrapfigure}
\textbf{Results.}
The model trained on the augmented dataset consistently outperforms the small-data model across SCM sizes (Fig.~\ref{fig:nihss-augmentation}), with the improvement becoming larger as SCM complexity increases.

\vspace{-0.95em}
\begin{customblocktakeaway}
    \checkmark\;\textbf{Takeaway.} Formalizing non-executable causal knowledge into an executable SCM enables data augmentation that improves causal reasoning in the original domain.
\end{customblocktakeaway}
\vspace{-0.125em}

\Needspace{20\baselineskip}
\vspace{-.75em}
\subsection{Anchoring Absolute Performance}
\label{sec:absolute-performance}
\vspace{-.75em}

\textbf{Motivation.} A natural question is how does this performance compare to larger and frontier models.

\begin{wraptable}[9]{r}{0.32\textwidth}
  \vspace{-1.2\baselineskip}
  \centering

  \captionsetup{
    font=footnotesize,
    justification=raggedright,
    singlelinecheck=false
  }

  \caption{Pass@1 (\%). N: \mbox{\textsc{nonsense}-\textit{code}};
  M: \mbox{\textsc{medical}-\textit{code}}.}
  \label{tab:absolute-performance}

  \vspace{-0.2cm}

  \footnotesize
  \setlength{\tabcolsep}{4pt}
  \renewcommand{\arraystretch}{1.08}

  \begin{tabular}{@{}lrr@{}}
    \toprule
    Model                   & N     & M     \\
    \midrule
    Qwen2.5-3B, base        & 14.73 & 18.26 \\
    \rowcolor{myPurple!15}
    Qwen2.5-3B, trained     & \textbf{39.18} & \textbf{31.07} \\
    Qwen2.5-7B, base        & 22.48 & 28.90 \\
    Qwen3-4B, base          & 25.83 & 39.59 \\
    GPT-5.5                 & 44.73 & 60.88 \\
    \bottomrule
  \end{tabular}

  \vspace{-1.0\baselineskip}
\end{wraptable}

\textbf{Method.} We compare the trained 3B model with its untrained checkpoint, larger/newer Qwen models, and GPT-5.5 on the same held-out code-domain sets (Table~\ref{tab:absolute-performance}). Appendix~\ref{app:performance_anchoring_details} gives query-type results and the reported protocol.

\textbf{Results.} The trained model exceeds the base 7B model in both domains and reaches 39.18\% versus GPT-5.5's 44.73\% on \textsc{nonsense}-\textit{code}. The larger medical-domain gap (31.07\% versus 60.88\%) highlights remaining transfer limitations. These are performance anchors, not compute-matched comparisons or evidence that causal reasoning is solved. We verify that the training procedure transfers across model families, obtaining substantial improvements by training Llama~3.2~3B with the same framework (Appendix~\ref{app:model_family}).

\section{Discussions}
\vspace{-.5em}

We introduce \textbf{$\stairwayupfilled$CauSim}, a novel framework that builds \textit{increasingly complex causal simulators}, unlocking executable supervision across scales and representations. Our key innovation is the \textit{incremental} growth of these simulators, enabling globally complex causal reasoning environments that would be difficult to construct otherwise.  \textbf{$\stairwayupfilled$CauSim} operates across representations, \textit{informalizing} executable representations in non-executable ones, which we instantiate as verbalized graphs, and \textit{formalizing} non-executable representations into executable code. Our study of \textbf{$\stairwayupfilled$CauSim} reveals its utility for improving causal reasoning: generalization to unseen domains and representations, consistent performance gains from curriculum and data scaling, LLM self-improvement through self-generated simulators, and data augmentation.

$\blacktriangleright$ \textbf{Limitations.} While \textbf{$\stairwayupfilled$CauSim} demonstrates the promise of increasingly complex causal simulators as verifiable training environments, two limitations should be noted. First, although we demonstrate generalization across several settings, this does not guarantee models acquire fully invariant causal mechanisms (see App.~\ref{app:pattern-vs-mechanism}). Second, the limits of scaling simulator-generated supervision remain empirical. Within the range studied, we observe no plateau as additional training data is introduced, and our generator ablation suggests that generator quality is not yet the dominant bottleneck. However, further gains will ultimately depend on factors such as model capacity, the diversity and quality of generated SCMs, and alignment between the training and evaluation distributions. We discuss these limits of recursive self-improvement and generator support further in Appendix~\ref{app:self-improvement-limitations}.

$\blacktriangleright$ \textbf{Future Work.} We see three promising directions. First, our current informalization instantiates non-executable representations as natural language, but the broader vision extends to richer causal systems: for instance, 2D/3D environments grounded by SCMs could enable spatial causal reasoning grounded with SCM supervison (see Table~\ref{tab:simulator_eg}). Second, we assume an explicit SCM as input; future work could draw on LLM parametric knowledge or elicit missing causal specifications through active task disambiguation~\cite{kobalczyk2025active} or amortized sequential experimental design~\cite{kriz2026amortizing}. Third, we focus primarily on deterministic settings. Future work should extend to stochastic causal reasoning, better capturing human-like reasoning under uncertainty where counterfactuals are ambiguous.

\section*{Acknowledgments}
We thank the anonymous NeurIPS reviewers and members of the van der Schaar lab for their insightful comments and suggestions. NA thanks W.D. Armstrong Trust for their support. NA's PhD studentship is funded by the Cystic Fibrosis Trust. AK's PhD studentship is funded by Sanofi. This work was supported by Azure sponsorship credits granted by Microsoft’s AI for Good Research Lab.

\setlength{\bibsep}{3pt plus 1pt minus 1pt}
\bibliographystyle{unsrtnat}
\bibliography{references}

\appendix

\appendixindex

\clearpage

\section{Experimental Details, Prompts, and Examples}
\label{app:experimental_details}

This section provides additional details for SCM construction and translation,
the programmatic verification used to retain generated SCMs, the experimental
configurations used in Sec.~\ref{sec:experiments}, and representative examples
of the datasets used for training and evaluation.

\subsection{Incremental SCM Construction}
\label{app:scm_construction}

CauSim constructs SCMs incrementally using four stages: specification,
planning, execution, and verification. We first illustrate the procedure
with a running example and then provide the corresponding prompt templates.

\subsubsection{Running Example}
\label{app:construction_example}

Some inputs and outputs below are abbreviated for readability.

\paragraph{Step 1: Specification.}
The specification stage creates an initial executable SCM. For example, given

\begin{lstlisting}[basicstyle=\ttfamily\footnotesize,breaklines=true]
Nodes: 1
Domain: math
\end{lstlisting}

the model may produce

\begin{lstlisting}[basicstyle=\ttfamily\footnotesize,breaklines=true]
def U_A():
    return random()

def f_A(u_a):
    return 2.0 * u_a

def run_once(seed):
    u_a = U_A()
    A = f_A(u_a)
    return {"A": A}
\end{lstlisting}

\paragraph{Step 2: Planning.}
To expand the SCM, the model receives a semantic view of the current SCM
rather than the complete program. For example,

\begin{lstlisting}[basicstyle=\ttfamily\footnotesize,breaklines=true]
{
  "summary":
    "Exogenous: U_A | Structural: f_A |
     Output: A | Edge: U_A -> f_A",

  "U_A": {
    "description": "Samples the exogenous input for A.",
    "children": ["f_A"]
  },

  "f_A": {
    "description": "Computes A from its exogenous input.",
    "parents": ["U_A"]
  }
}
\end{lstlisting}

The planner proposes one local graph update:

\begin{lstlisting}[basicstyle=\ttfamily\footnotesize,breaklines=true]
{
  "new_function": "f_B",
  "new_exo_noise": "U_B",
  "direct_parents": ["f_A"],
  "direct_children": []
}
\end{lstlisting}

\paragraph{Step 3: Execution.}
The execution stage receives the plan together with only the relevant local
SCM context and implements the update:

\begin{lstlisting}[basicstyle=\ttfamily\footnotesize,breaklines=true]
def U_B():
    return random()

def f_B(A, u_b):
    return A + u_b

def run_once(seed):
    u_a = U_A()
    A = f_A(u_a)
    u_b = U_B()
    B = f_B(A, u_b)
    return {"A": A, "B": B}
\end{lstlisting}

\paragraph{Step 4: Verification.}
The resulting program is subjected to deterministic structural and execution
checks described in Sec.~\ref{app:verification}. Invalid candidates are
rejected and regenerated up to the specified retry budget. Steps 2--4 repeat
until the requested SCM size is reached.

\subsubsection{Specification Prompt}
\label{app:specification_prompt}

For the \textsc{nonsense} domain, the initial SCM is generated using:

\begin{lstlisting}[basicstyle=\ttfamily\tiny,breaklines=true]
You are generating an EXECUTABLE, STOCHASTIC Structural
Causal Model (SCM) in Python.

HARD REQUIREMENTS:
- All variables used in the SCM must have no semantic
  meaning. Variable names should be non-existent words
  that are short, i.e., within 5 characters.
- Functions may use if/else, for-loops, boolean logic,
  and string comparisons. DO NOT make the SCM mathematical.
- Allowed imports: random, typing. No others.
- Use docstrings ONLY (no # comments).
- Define exogenous samplers named U_* (pure, no I/O).
  - Each U_* must be independent of all other
    exogenous variables.
  - There must be no exogenous confounding.
- Define structural assignments named f_* (pure, no I/O).
  - Each f_* must have ONE exogenous parent U_*.
  - Each f_* must take all parent variables explicitly
    as arguments.
  - Structural functions are deterministic given inputs.
  - The SCM MUST be acyclic.
  - The SCM MUST be connected.
- Include a driver:
    def run_once(seed: int | None = None)
  - If seed is not None, call random.seed(seed).
  - Assign each function call on a separate line.
  - Return ONLY outputs of structural assignments f_*.

OUTPUT FORMAT:
1) All U_* sampler definitions
2) All f_* structural function definitions
3) The run_once(seed) driver

Return ONLY a single Python code block. No prose.

Task:
- Create an SCM with [NUM_NODES] exogenous samplers and
  [NUM_NODES] structural functions.
- The internal nodes must follow a [TOPOLOGY] topology.

[TOPOLOGY DEFINITION]
\end{lstlisting}

For \textsc{medical} SCMs, we use the corresponding logic-based variant:

\begin{lstlisting}[basicstyle=\ttfamily\tiny,breaklines=true]
You are generating a SMALL, EXECUTABLE, STOCHASTIC
Structural Causal Model (SCM) in Python.

This SCM must be logic-based, symbolic, or mechanistic,
based on a clinical, biological, or medical domain.
DO NOT make the SCM mathematical.

Functions may return booleans, categorical strings,
or symbolic states.

HARD REQUIREMENTS:
- Allowed imports: random, typing. No others.
- Use docstrings ONLY.
- Define independent exogenous samplers named U_*.
- Define structural assignments named f_*.
- Each f_* must take all parent variables explicitly.
- Each f_* must have ONE exogenous parent.
- The SCM must be acyclic.
- Structural functions are deterministic given inputs.
- Include a run_once(seed) driver that returns outputs
  from f_* only.

OUTPUT FORMAT:
1) All U_* exogenous sampler definitions
2) All f_* structural mechanism definitions
3) The run_once(seed) driver

Return ONLY a single Python code block. No prose.

Task:
- Create an SCM with [NUM_NODES] exogenous samplers and
  [NUM_NODES] structural functions.
- The internal nodes must follow a [TOPOLOGY] topology.

[TOPOLOGY DEFINITION]
\end{lstlisting}

\subsubsection{Topology Constraints}
\label{app:topology_prompts}

Our main experiments use chain, inverted-star, and layered endogenous
graph structures.

\paragraph{Chain.}

\begin{lstlisting}[basicstyle=\ttfamily\tiny,breaklines=true]
A chain topology is a single linear sequence of
internal nodes.

Rules:
- Exactly one parent per node except the first.
- Exactly one child per node except the last.
- Shape: A -> B -> C -> ...
- If inserting a new node between X and Y:
  - Y removes X as a parent.
  - Y adds the new node as its only parent.
  - The new node takes X as its parent.
\end{lstlisting}

\paragraph{Inverted Star.}

\begin{lstlisting}[basicstyle=\ttfamily\tiny,breaklines=true]
An inverted star topology is the reverse of a star:
many parents converge into a single child.

Rules:
- A single sink node has multiple parents.
- All parents have exactly one child: the sink.
- The sink has no children.
- Parents do not depend on each other.
- Shape: {P1, P2, P3, ...} -> Child.
- New nodes can only be inserted as direct parents
  of the sink.
\end{lstlisting}

\paragraph{Layered.}

\begin{lstlisting}[basicstyle=\ttfamily\tiny,breaklines=true]
Rules:
- Every node belongs to exactly one layer.
- Multiple nodes may occupy one layer, but there
  are no edges within a layer.
- Edges are permitted only from layer k to layer k+1.
- Edges may not skip layers or point backwards.
\end{lstlisting}

\subsubsection{Planning Prompt}
\label{app:planning_prompt}

At each incremental step, the planner receives a semantic view of the
existing SCM:

\begin{lstlisting}[basicstyle=\ttfamily\tiny,breaklines=true]
You are planning the NEXT INCREMENTAL GROWTH of an existing
EXECUTABLE, STOCHASTIC Structural Causal Model (SCM)
in Python.

You are provided a SEMANTIC VIEW of the existing SCM:
- Exogenous samplers: U_* (external nodes)
- Structural assignments: f_* (internal nodes)

TASK
- Propose ONE new structural assignment and its
  corresponding exogenous sampler.
- Define the direct parents and direct children
  of this new structural assignment.

HARD REQUIREMENTS:
- The corresponding exogenous sampler is REQUIRED.
- direct_parents and direct_children may contain ONLY
  pre-existing structural assignments f_*.
- At least one of direct_parents or direct_children
  must be non-empty.
- The update must not introduce cycles.
- New function names must be unique.

OUTPUT FORMAT
Return STRICT JSON:
{
  "rationale": "one-sentence design intent",
  "new_function": "name of the new f_* function",
  "new_exo_noise": "name of the new U_* function",
  "direct_parents": ["f_*", ...],
  "direct_children": ["f_*", ...]
}

Additional Topology HARD Requirement:
- The updated SCM must obey the [TOPOLOGY] topology.

[TOPOLOGY DEFINITION]

SEMANTIC VIEW:
[CURRENT SCM SEMANTIC VIEW]
\end{lstlisting}

\subsubsection{Execution Prompt}
\label{app:execution_prompt}

The execution stage receives the decision, local code context, and current
driver:

\begin{lstlisting}[basicstyle=\ttfamily\tiny,breaklines=true]
You will IMPLEMENT a small update to an existing
EXECUTABLE, STOCHASTIC structural causal model (SCM)
in Python.

You are given:

1. DECISION JSON:
   - rationale
   - new_function
   - new_exo_noise
   - direct_parents
   - direct_children

2. CONTEXT CODE:
   - For each direct child: full function
   - For each direct parent: semantic information

3. DRIVER FUNCTION:
   def run_once(seed: int | None)

TASK:
- Define the new exogenous noise sampler.
- Define the new structural function.
- Its arguments must contain its corresponding
  exogenous input and all direct structural parents.
- Structural logic must be deterministic given inputs.
- If direct children are specified, modify their
  signatures and mechanisms.
- Update the driver in topological order.

HARD REQUIREMENTS:
- Preserve purity.
- Do not emit unrelated functions.
- Do not emit direct-parent implementations.

DECISION JSON:
[DECISION JSON]

CONTEXT CODE:
[LOCAL CODE CONTEXT]

DRIVER FUNCTION:
[CURRENT DRIVER]
\end{lstlisting}

\subsection{Programmatic SCM Verification}
\label{app:verification}

Verification is deterministic and does not use an LLM. Generated or updated
SCMs are retained only after passing the required structural and execution
checks. Failed candidates are rejected and regenerated up to the retry budget.

\paragraph{Program restrictions.}
We check that imports belong to an allowed set and reject unsafe or impure
operations such as file access, external requests, subprocess calls, or other
prohibited side effects.

\paragraph{Acyclicity and execution order.}
We extract the generated causal graph and perform a topological sort.
Candidates for which a complete topological ordering cannot be obtained are
rejected as cyclic.

\paragraph{Exogenous structure.}
The validator checks that an exogenous node points only to endogenous
structural nodes, that each exogenous variable affects at most one structural
node, and that an endogenous structural node has at most one exogenous parent.
We additionally require that exogenous randomness reaches at least one
structural variable used by the driver.

\paragraph{Topology.}
For experiments with a specified topology, the endogenous subgraph is checked
against the requested structural family. In our main experiments, this includes
chain, inverted-star, and layered structures.

\paragraph{Node count.}
When a target SCM size is specified, the number of endogenous and exogenous
nodes is checked against the requested value.

\paragraph{Executability.}
Generated programs are dynamically loaded and required to define the driver
function. We execute the resulting SCM under controlled exogenous values and
reject programs that raise an exception.

\paragraph{Reference-graph agreement.}
For formalization settings in which an explicit source graph is available,
we additionally compare the endogenous nodes and directed edges of the
generated SCM against the source graph. Missing or additional nodes and
missing or spurious edges cause validation to fail.

Together, these checks ensure that retained SCMs are executable, acyclic,
structurally consistent with the requested graph constraints, and suitable
for mechanically computing causal-query answers.

\subsection{SCM Translation and Formalization}
\label{app:translation_prompts}

CauSim translates between executable and non-executable causal
representations. We distinguish \textit{informalization}, which converts an
executable SCM into natural language, from \textit{formalization}, which
converts external causal knowledge into an executable SCM.

\subsubsection{Informalization: Executable SCM to Natural Language}
\label{app:informalization_prompt}

We translate a structured representation of an executable SCM into
semantically equivalent natural language:

\begin{lstlisting}[basicstyle=\ttfamily\tiny,breaklines=true]
You will be given a structured dictionary representation
of a Structural Causal Model (SCM).

YOUR TASK:
Translate the SCM into unstructured natural language
that is SEMANTICALLY EQUIVALENT and LOSSLESS.

NAME PRESERVATION:
All names MUST be preserved EXACTLY.

WHAT TO DESCRIBE:

1. Exogenous variables:
- State the variable's instantiation in the driver.
- Describe exactly how its value is generated.

2. Structural variables:
- State the variable's instantiation in the driver.
- Explicitly state all inputs in their original order.
- Describe exactly how its value is generated.

3. Driver output:
- Describe the return value of the driver.
- State every output name and the variable to which
  it corresponds.

OUTPUT RULES:
- One paragraph per variable in topological order.
- Output only unstructured natural language.
- Do not output code, equations, bullet points, or lists.
- Do not introduce assumptions or mechanisms not
  present in the SCM.

Dictionary:
[STRUCTURED SCM REPRESENTATION]
\end{lstlisting}

\subsubsection{One-Shot Natural-Language Formalization}
\label{app:oneshot_formalization_prompt}

The one-shot baseline receives the full natural-language SCM in a single call:

\begin{lstlisting}[basicstyle=\ttfamily\tiny,breaklines=true]
You are given an unstructured natural language description
of a Structural Causal Model (SCM).

The description is guaranteed to be LOSSLESS and COMPLETE.

YOUR TASK:
Write the SCM as executable Python code that exactly
matches the described causal structure.

HARD REQUIREMENTS:
- Add imports: import random, import typing. No others.
- Use docstrings ONLY.
- Define one exogenous sampler per structural assignment.
- Each exogenous sampler must be independent.
- Define structural assignments named f_*.
- Each structural assignment must take all parent
  variables explicitly.
- Each structural assignment must have one
  exogenous parent.
- Structural assignments are deterministic given inputs.
- The SCM must be acyclic.
- Do not define additional functions.

OUTPUT FORMAT:
1) imports
2) all exogenous sampler definitions
3) all structural mechanism definitions
4) driver

Natural Language Description:
[NATURAL-LANGUAGE SCM]
\end{lstlisting}

\subsubsection{Step-by-Step Formalization}
\label{app:stepwise_formalization_prompt}

Our step-by-step formalization uses a two-stage decision--action procedure.

\paragraph{Decision Stage.}

\begin{lstlisting}[basicstyle=\ttfamily\tiny,breaklines=true]
You are given:

1. SEMANTIC INFORMATION FOR THE EXISTING SCM.
2. NEW ITEM that needs to be formalized.

YOUR TASK:
Insert the NEW ITEM into the existing SCM.

Specifically:
1. Choose the new structural function.
2. Choose its exogenous sampler.
3. Identify direct parents already in the SCM.
4. Identify direct children already in the SCM.

IMPORTANT:
If the NEW ITEM does not mention direct parents or
children in the current SCM, leave them empty.

Return Strict JSON:
{
  "new_function": "f_<item_name>",
  "new_exo_noise": "U_<item_name>",
  "direct_parents": ["f_<parent1>", ...],
  "direct_children": ["f_<child1>", ...]
}

SEMANTIC INFORMATION FOR THE EXISTING SCM:
[CURRENT SCM SEMANTIC INFORMATION]

NEW ITEM:
[NEW ITEM DESCRIPTION]
\end{lstlisting}

\paragraph{Action Stage.}

\begin{lstlisting}[basicstyle=\ttfamily\tiny,breaklines=true]
You will IMPLEMENT a small update to a structural causal
model in which ONE new structural function is added.

You are given:
1. A DECISION JSON.
2. The specification of the NEW FUNCTION.
3. CODE CONTEXT:
   - full code for each direct child
   - semantic information for each direct parent

TASK:
Implement exactly the update specified by the decision.

1. Define the new exogenous sampler.
2. Define the new structural function.
3. Modify each direct child if required.

OUTPUT FORMAT:
- Full definition of the new function.
- Full definition of the new exogenous sampler.
- Full definitions of modified direct children.
- Do not output prose.

DECISION JSON:
[DECISION JSON]

NEW FUNCTION DESCRIPTION:
[NEW ITEM]

CODE CONTEXT:
[LOCAL CODE CONTEXT]
\end{lstlisting}

\subsubsection{NIH Stroke Scale Formalization}
\label{app:nihss_formalization}

For NIHSS, the same decision--action procedure is augmented with the
following domain constraints:

\begin{lstlisting}[basicstyle=\ttfamily\tiny,breaklines=true]
1. Treat each scored NIHSS item as a structural variable.
   Each score should be equally likely.

2. If an item explicitly depends on another item's value,
   that dependency MUST be represented in the function
   signatures.

3. Do not use "UN" unless specified in the item.

4. Do NOT rename, simplify, or reinterpret labels.

5. Update total_score to be the sum of ALL inputs.
   If an input is "UN", ignore it in the summation.
\end{lstlisting}

For example,

\begin{lstlisting}[basicstyle=\ttfamily\footnotesize,breaklines=true]
8. f_sensory
Test pinprick sensation.

Score:
0 = Normal; no sensory loss.
1 = Mild-to-moderate sensory loss.
2 = Severe or total sensory loss.

Patients with f_level_of_consciousness = 3
automatically score 2.
\end{lstlisting}

is formalized as

\begin{lstlisting}[basicstyle=\ttfamily\footnotesize,breaklines=true]
def f_sensory(f_level_of_consciousness, u_sensory):
    if f_level_of_consciousness == 3:
        return 2
    return u_sensory
\end{lstlisting}

\subsubsection{Program-to-SCM Formalization for GSM8K}
\label{app:gsm8k_formalization}

For GSM8K-derived tasks, the source representation is an executable reasoning
program. We convert it into an SCM using:

\begin{lstlisting}[basicstyle=\ttfamily\tiny,breaklines=true]
You are given a Python program that solves a
reasoning problem.

Your task is to convert this program into a
Structural Causal Model written as executable Python code.

CORE PRINCIPLE:
Each variable is represented by an exogenous variable
and a deterministic structural function.

EXOGENOUS VARIABLES:
- Every directly assigned constant must have an
  exogenous variable.
- Its support must include the original value.

STRUCTURAL FUNCTIONS:
- Define one function per variable.
- Include all causal parents explicitly.
- Preserve the exact program logic.

DEPENDENCIES:
- The dependency structure must match the program exactly.

CONDITIONALS:
- Preserve all if/else logic.

LOOPS:
- Rewrite explicitly where possible or preserve them.

DRIVER:
- Execute variables in causal order.
- Return all endogenous variables.

INPUT PROGRAM:
[PROGRAM]
\end{lstlisting}

\subsubsection{Bayesian-Network Formalization}
\label{app:bn_formalization}

For ASIA, CANCER, and SURVEY, the source representation is a fully specified
Bayesian network with conditional probability distributions.

\paragraph{One-Shot BN Formalization.}

\begin{lstlisting}[basicstyle=\ttfamily\tiny,breaklines=true]
You are given a fully specified Bayesian Network in JSON.

The BN defines:
- variables
- conditional probability distributions (CPDs)

Your task is to convert the BN into an equivalent
executable Structural Causal Model.

- Use independent exogenous noise for stochastic variables.
- Structural functions are deterministic given their inputs.
- Select the correct CPD according to parent values.
- Implement stochastic CPDs using inverse-CDF sampling.
- Deterministic CPDs do not require exogenous noise.
- Execute variables in topological order.
- Return all endogenous variables.
- The SCM must reproduce the BN distribution.

Input BN:
[BAYESIAN NETWORK]
\end{lstlisting}

\paragraph{Step-by-Step BN Formalization.}

The decision stage is the same as above. The action stage is specialized
to the supplied CPD:

\begin{lstlisting}[basicstyle=\ttfamily\tiny,breaklines=true]
You are given:

1. A DECISION JSON.
2. A NEW BN NODE SPECIFICATION containing the variable
   and its conditional probability table.
3. LOCAL CODE CONTEXT.

Implement exactly the specified graph update.

- Use independent exogenous noise for stochastic nodes.
- Use inverse-CDF sampling.
- Include every structural parent explicitly.
- Deterministic CPDs require no exogenous noise.
- Preserve the supplied CPD exactly.

OUTPUT:
1. New exogenous function, if needed
2. New structural function
3. Modified direct-child functions, if needed

DECISION JSON:
[DECISION JSON]

NEW BN NODE SPECIFICATION:
[BN NODE AND CPD]

CODE CONTEXT:
[LOCAL CODE CONTEXT]
\end{lstlisting}

\subsection{Formalization Results}
\label{app:formalization_results}

Step-by-step formalization produces more valid executable SCMs than one-shot
formalization for NIHSS and both Bayesian-network datasets:

\begin{table}[t]
\centering
\small
\begin{tabular}{lcc}
\toprule
Dataset & Step-by-Step & One-Shot \\
\midrule
NIHSS & \textbf{52/100} & 20/100 \\
ASIA & \textbf{84/100} & 69/100 \\
CANCER & \textbf{74/100} & 51/100 \\
\bottomrule
\end{tabular}
\captionsetup{font=small}
\caption{Step-by-step versus one-shot formalization accuracy (successful trials out of 100).}
\label{tab:formalization_bn}
\end{table}

\subsection{Experimental Configurations}
\label{app:experiment_configurations}

\subsubsection{Generalization}

\begin{table*}[t]
\centering
\scriptsize
\resizebox{\textwidth}{!}{
\begin{tabular}{llll}
Experiment & Model & Training Data & Evaluation Data \\
In-distribution &
Qwen2.5-3B-Instruct, BLOCK, 135 queries/type &
GPT-5 \textsc{nonsense}-\textit{code} &
6,000 held-out \textsc{nonsense}-\textit{code} queries \\
Domain shift &
Same &
Same &
6,000 held-out \textsc{medical}-\textit{code} queries \\
Domain + representation &
Three Qwen2.5-3B-Instruct variants &
\textsc{nonsense}-\textit{code} and \textsc{nonsense}-\textit{nl} &
6,000 held-out \textsc{medical}-\textit{nl} queries \\
External &
Qwen2.5-3B-Instruct, trained &
\textsc{nonsense}-\textit{code} &
GSM8K, ASIA, CANCER, and SURVEY \\
Partially specified &
Qwen2.5-3B-Instruct, trained &
\textsc{nonsense}-\textit{code} &
262 GSM8K SCMs, 500 total queries \\
\end{tabular}
}
\captionsetup{font=small}
\caption{Experimental configurations for generalization.}
\label{tab:generalization_setup}
\end{table*}

For synthetic evaluation, each 6,000-query set contains three topologies,
20 SCM IDs per topology, 10 SCM sizes, five queries per type, and two query
types. For the domain-and-representation experiment, we compare a
\textsc{nonsense}-\textit{code}-trained model, a directly
\textsc{nonsense}-\textit{nl}-trained model, and a code-trained model further
fine-tuned on \textsc{nonsense}-\textit{nl}. Formalized GSM8K contains
270 SCMs with 270 intervention and 264 counterfactual queries. ASIA, CANCER,
and SURVEY each contain 100 intervention and 100 counterfactual queries.

\subsubsection{Curriculum and Dataset Size}

\begin{table*}[t]
\centering
\scriptsize
\resizebox{\textwidth}{!}{
\begin{tabular}{llll}
Experiment & Model & Training Configuration & Evaluation \\
Curriculum &
Qwen2.5-3B-Instruct &
BLOCK, AT-ONCE, EASYONLY, HARDONLY; 64,800 queries each &
Held-out \textsc{nonsense}-\textit{code} \\
Dataset size &
Qwen2.5-3B-Instruct &
BLOCK with 5, 15, 45, or 135 queries/type &
Held-out \textsc{nonsense}-\textit{code} \\
\end{tabular}
}
\captionsetup{font=small}
\caption{Curriculum and dataset-size configurations.}
\label{tab:scaling_setup}
\end{table*}

Training uses GPT-5-generated \textsc{nonsense}-\textit{code} SCMs with
chain, inverted-star, and layered topologies and 180 SCM IDs per topology.
Training levels are $l_2$, $l_4$, $l_6$, and $l_8$. BLOCK and AT-ONCE use
15 queries per type at each level. EASYONLY uses 60 queries per type at
$l_2$, while HARDONLY uses 60 queries per type at $l_8$.

\subsubsection{Self-Improvement}

\begin{table*}[t]
\centering
\scriptsize
\resizebox{\textwidth}{!}{
\begin{tabular}{llll}
Experiment & Model & Training Data & Evaluation \\
Generator diversity &
Qwen2.5-7B-Instruct &
SCMs from GPT-5, GPT-5-mini, or Qwen2.5-7B-Instruct &
Held-out GPT-5 \textsc{nonsense}-\textit{code} \\
\end{tabular}
}
\captionsetup{font=small}
\caption{Self-improvement experimental configuration.}
\label{tab:self_improvement_setup}
\end{table*}

Each generator produces 180 SCM IDs with incremental versions spanning
1--10 nodes. Training uses BLOCK over $l_2$, $l_4$, $l_6$, and $l_8$,
with 15 intervention and 15 counterfactual queries per level.

\subsubsection{NIHSS Data Augmentation}

\begin{table*}[t]
\centering
\scriptsize
\resizebox{\textwidth}{!}{
\begin{tabular}{llll}
Model & Small Training Set & Augmented Training Set & Evaluation \\
Qwen2.5-3B-Instruct &
50 queries/type/size; 400 total &
1,000 queries/type/size; 8,000 total &
$l_1$--$l_{10}$; 8,000 held-out queries \\
\end{tabular}
}
\captionsetup{font=small}
\caption{NIHSS augmentation configuration.}
\label{tab:nihss_setup}
\end{table*}

Both training conditions use BLOCK over $l_2$, $l_4$, $l_6$, and $l_8$.
The evaluation set contains 400 queries per type at each size from
$l_1$ through $l_{10}$.

\subsection{Additional Experimental Analyses}
\label{app:additional_analyses}

The supplementary analyses are reported in dedicated appendices to avoid duplicating results. Appendix~\ref{app:performance_anchoring} compares absolute performance against larger, newer, and frontier models and gives the query-type breakdown in Table~\ref{tab:performance_anchoring}. Appendix~\ref{app:model_family} tests transfer of the training procedure to Llama~3.2~3B (Table~\ref{tab:llama_family}). Appendix~\ref{app:scm_query_ablation} compares allocations of a fixed supervision budget between distinct SCMs and queries per SCM (Table~\ref{tab:scm_query_ablation}).

\subsection{Representative Dataset Examples}
\label{app:dataset_examples}

\subsubsection{\textsc{nonsense}-\textit{code}}
\label{app:nonsense_code_example}

\begin{lstlisting}[basicstyle=\ttfamily\footnotesize,breaklines=true]
def f_XQ(u_xq: str) -> str:
    if u_xq in ('aa', 'bb'):
        return 'XQ'
    return 'xq'
\end{lstlisting}

\subsubsection{\textsc{nonsense}-\textit{nl}}

\begin{lstlisting}[basicstyle=\ttfamily\footnotesize,breaklines=true]
mevu is set as the returned output of f_MEVU.
The inputs to f_MEVU are u_mevu, BELU.

The Generation Rule for f_MEVU is: when u_mevu equals
"m", map BELU to "T" if BELU is True and to "F" if
BELU is False; when u_mevu equals "n", map BELU to
"Y" if BELU is True and to "N" if BELU is False.
\end{lstlisting}

\subsubsection{\textsc{medical}-\textit{code}}

\begin{lstlisting}[basicstyle=\ttfamily\footnotesize,breaklines=true]
def f_infection_status(U_infection_exposure):
    if U_infection_exposure == 'none':
        return 'no_infection'
    elif U_infection_exposure == 'low':
        return 'colonized'
    else:
        return 'infected'
\end{lstlisting}

\subsubsection{\textsc{medical}-\textit{nl}}

\begin{lstlisting}[basicstyle=\ttfamily\footnotesize,breaklines=true]
adaptive_status is set as the returned output of
f_adaptive_status. The input to f_adaptive_status is
U_vaccination_history.

The generation rule is: if U_vaccination_history equals
"none", return "naive"; if it equals "partial", return
"primed"; otherwise return "boosted".
\end{lstlisting}

\subsubsection{Bayesian Networks from \texttt{bnlearn}}

\begin{lstlisting}[basicstyle=\ttfamily\footnotesize,breaklines=true]
def f_Cancer(smoker, pollution, u_cancer):
    states = ['True', 'False']
    poll_index = 0 if pollution == 'low' else 1
    smoke_index = 0 if smoker == 'True' else 1

    p_true = [
        [0.03, 0.001],
        [0.05, 0.02]
    ][poll_index][smoke_index]

    p_false = [
        [0.97, 0.999],
        [0.95, 0.98]
    ][poll_index][smoke_index]

    probs = [p_true, p_false]
    cum = 0.0

    for p, s in zip(probs, states):
        cum += p
        if u_cancer < cum:
            return s

    return states[-1]
\end{lstlisting}

\subsubsection{Formalized GSM8K}

\begin{lstlisting}[basicstyle=\ttfamily\footnotesize,breaklines=true]
def f_calories_remaining_for_dinner(
    calories_per_day,
    calories_at_breakfast,
    calories_at_lunch,
    calories_at_snack,
):
    return (
        calories_per_day
        - calories_at_breakfast
        - calories_at_lunch
        - calories_at_snack
    )
\end{lstlisting}

\subsubsection{Partially Specified GSM8K}
\label{app:partial_gsm8k_example}

\begin{lstlisting}[basicstyle=\ttfamily\footnotesize,breaklines=true]
A cafe owner records 1 sandwich on Saturday and
2 on Sunday. Each sandwich uses an unknown 2 or
3 bread pieces; the factual record says 3.

Set the total number of bread pieces to 9 while
retaining the inferred facts. Return:

["num_of_pieces_of_bread_per_sandwich"]
\end{lstlisting}

\subsubsection{NIH Stroke Scale}

The NIHSS data begin from non-executable clinical scoring rules that are
incrementally formalized into executable structural mechanisms. A
representative rule and its formalized mechanism are given in
Sec.~\ref{app:nihss_formalization}.
\newpage
\section{Performance Anchoring Against Larger and Frontier Models}
\label{app:performance_anchoring}

To contextualize the absolute performance reported in Sec.~\ref{generalization} and the magnitude of improvement from training, we compare our trained \texttt{Qwen2.5-3B-Instruct} model against several untrained reference models. We include: (i) its corresponding untrained \texttt{Qwen2.5-3B-Instruct} base model, (ii) the larger \texttt{Qwen2.5-7B-Instruct} model from the same model family, (iii) the newer-generation \texttt{Qwen3-4B-Instruct-2507}, and (iv) \texttt{GPT-5.5} as a frontier-model reference.

\subsection{Evaluation Protocol and Query-Type Results}
\label{app:performance_anchoring_details}

\textbf{Trained model.} We use the same \texttt{Qwen2.5-3B-Instruct} checkpoint as in Sec.~\ref{generalization}: \textsc{Block} training on \textsc{nonsense}-\textit{code}, with 135 queries per type at each training size $l_2$, $l_4$, $l_6$, and $l_8$ (Appendix~\ref{app:experiment_configurations}). Here, ``base'' denotes the corresponding instruction-tuned checkpoint before our causal-reasoning training, not a separate pretrained-only model.

\textbf{Evaluation sets and metric.} All five models are evaluated on the same held-out \textsc{nonsense}-\textit{code} and \textsc{medical}-\textit{code} sets. Each set contains 6,000 queries: three topologies, 20 SCM IDs per topology, 10 incremental sizes, and five queries per type and size. Intervention and counterfactual queries therefore contribute 3,000 examples each. Table~\ref{tab:performance_anchoring} reports Pass@1 (\%) separately by query type and their average; Table~\ref{tab:absolute-performance} presents the average scores in the main paper. Correctness is determined using the executable verification procedure described in Sec.~\ref{subsec:causal-query-generation}.

\textbf{Scope of the comparison.} These are absolute-performance reference points, not compute-matched comparisons: model scale, pretraining, and inference configuration are not controlled across model families. The results therefore quantify the observed performance gaps without isolating a compute-normalized training advantage.

\begin{table}[H]
\centering
\small
\setlength{\tabcolsep}{4pt}
\renewcommand{\arraystretch}{1.08}
\begin{tabularx}{\linewidth}{@{}Xrrr@{}}
\toprule
\textbf{Model} & \textbf{Avg. Pass@1} & \textbf{Counterfactual} & \textbf{Intervention} \\
\midrule
\multicolumn{4}{@{}l}{\textsc{nonsense}-\textit{code}} \\
Qwen2.5-3B-Instruct, base     & 14.73 & 13.93 & 15.53 \\
\rowcolor{myPurple!15}
Qwen2.5-3B-Instruct, trained  & 39.18 & 43.47 & 34.90 \\
Qwen2.5-7B-Instruct, base     & 22.48 & 22.33 & 22.63 \\
Qwen3-4B-Instruct-2507, base  & 25.83 & 23.00 & 28.67 \\
GPT-5.5                     & \textbf{44.73} & \textbf{44.43} & \textbf{45.03} \\
\midrule
\multicolumn{4}{@{}l}{\textsc{medical}-\textit{code}} \\
Qwen2.5-3B-Instruct, base     & 18.26 & 17.90 & 18.62 \\
\rowcolor{myPurple!15}
Qwen2.5-3B-Instruct, trained  & 31.07 & 31.34 & 30.79 \\
Qwen2.5-7B-Instruct, base     & 28.90 & 29.24 & 28.55 \\
Qwen3-4B-Instruct-2507, base  & 39.59 & 38.86 & 40.31 \\
GPT-5.5                     & \textbf{60.88} & \textbf{60.41} & \textbf{61.34} \\
\bottomrule
\end{tabularx}
\captionsetup{font=small}
\caption{Performance anchoring against larger, newer, and frontier models. Pass@1 (\%) on the same held-out \textsc{nonsense}-\textit{code} and \textsc{medical}-\textit{code} sets.}
\label{tab:performance_anchoring}
\end{table}

\textbf{Results.} Training substantially improves the Qwen2.5-3B model over its untrained counterpart in both domains. The trained 3B model also outperforms the larger untrained Qwen2.5-7B model on both \textsc{nonsense}-\textit{code} (39.18\% vs.\ 22.48\%) and \textsc{medical}-\textit{code} (31.07\% vs.\ 28.90\%), showing that the improvement from causal training is not explained simply by model scale.

Second, relative to the newer Qwen3-4B model, the trained Qwen2.5-3B model performs better on the training-domain representation, \textsc{nonsense}-\textit{code} (39.18\% vs.\ 25.83\%), but not on the unseen \textsc{medical}-\textit{code} domain (31.07\% vs.\ 39.59\%). This comparison provides a reference point for the gains achieved under the current training curriculum and data scale, particularly under domain shift.

Finally, GPT-5.5 provides an estimate of frontier-model performance on these tasks. On \textsc{nonsense}-\textit{code}, the trained 3B model reaches 39.18\% Pass@1 compared with 44.73\% for GPT-5.5. The gap is larger on the unseen \textsc{medical}-\textit{code} domain, where the respective scores are 31.07\% and 60.88\%.

\newpage

\section{Generalization Across Model Families}
\label{app:model_family}

Our main experiments use Qwen2.5 models: Qwen2.5-3B-Instruct in Secs.~\ref{generalization}--\ref{scaling} and Qwen2.5-7B-Instruct in Sec.~\ref{self-improvement}, demonstrating that the training procedure can be applied across model sizes. We additionally test whether the same procedure transfers to a different model family by training Llama~3.2~3B on \textsc{nonsense}-\textit{code}.

We train on 180 SCM IDs spanning 1--10 endogenous variables, with five queries per SCM ID at each complexity level. Evaluation uses 20 held-out SCM IDs over the same range of graph sizes and query counts. We report performance on \textsc{nonsense}-\textit{code}, \textsc{nonsense}-\textit{nl}, and \textsc{medical}-\textit{code}.

\begin{table}[t]
\centering
\small
\begin{tabular}{llccc}
\toprule
\textbf{Domain} & \textbf{Model} & \textbf{Avg. Pass@1} &
\textbf{Counterfactual} & \textbf{Intervention} \\
\midrule

\textsc{nonsense}-\textit{code}
& Base
& 10.57\%
& 10.20\%
& 10.93\% \\

& Trained
& \textbf{40.55\%}
& \textbf{43.20\%}
& \textbf{37.90\%} \\

\midrule

\textsc{nonsense}-\textit{nl}
& Base
& 11.40\%
& 11.30\%
& 11.50\% \\

& Trained
& \textbf{37.03\%}
& \textbf{38.80\%}
& \textbf{35.27\%} \\

\midrule

\textsc{medical}-\textit{code}
& Base
& 9.53\%
& 8.93\%
& 10.14\% \\

& Trained
& \textbf{26.50\%}
& \textbf{28.14\%}
& \textbf{24.86\%} \\

\bottomrule
\end{tabular}
\captionsetup{font=small}
\caption{Generalization of the training procedure to Llama~3.2~3B.
Training improves Pass@1 across in-distribution, representation-shift,
and domain-shift evaluations.
}
\label{tab:llama_family}
\end{table}

Training substantially improves Llama~3.2~3B across all three evaluation domains. Average Pass@1 increases from 10.57\% to 40.55\% on \textsc{nonsense}-\textit{code}, from 11.40\% to 37.03\% on \textsc{nonsense}-\textit{nl}, and from 9.53\% to 26.50\% on \textsc{medical}-\textit{code}. These results show that the gains from our training procedure are not specific to the Qwen model family.

\section{7B Generator Comparisons}
\label{app:7b-comparisons}

This section reports the full generator-ablation results for Qwen2.5-7B-Instruct. The target model is fine-tuned on \textsc{Formal Nonsense} SCMs generated by different sources: the target model itself, GPT-5-mini, and GPT-5. All models are evaluated on the same held-out GPT-5-generated benchmark.

\begin{table}[H]
\centering
\caption{\textbf{Pass@1 performance on counterfactual queries with 5 training instances per SCM.}
Qwen2.5-7B-Instruct is fine-tuned on \textbf{5} \textsc{Formal Nonsense} instances per SCM, using 250 SCMs in total. Evaluation uses a held-out GPT-5-generated benchmark with 20 SCMs and 5 instances per SCM. Columns indicate the number of nodes.}
\label{tab:7b_pass1_5_counterfactual}
\resizebox{\textwidth}{!}{%
\begin{tabular}{lcccccccccc}
\hline
Training data & 1 & 2 & 3 & 4 & 5 & 6 & 7 & 8 & 9 & 10 \\
\hline
Base & 0.878 & 0.635 & 0.450 & 0.298 & 0.232 & 0.157 & 0.097 & 0.082 & 0.060 & 0.052 \\
7B (self-generation) & 0.878 & 0.660 & \textbf{0.497} & 0.357 & 0.265 & 0.213 & 0.162 & 0.113 & \textbf{0.087} & 0.072 \\
GPT-5-mini & 0.905 & \textbf{0.720} & 0.468 & \textbf{0.360} & \textbf{0.295} & \textbf{0.218} & 0.160 & \textbf{0.115} & 0.078 & 0.072 \\
GPT-5 & \textbf{0.925} & 0.692 & 0.482 & 0.352 & 0.268 & 0.188 & \textbf{0.178} & 0.110 & 0.085 & \textbf{0.075} \\
\hline
\end{tabular}%
}
\end{table}

\begin{table}[H]
\centering
\caption{\textbf{Pass@1 performance on intervention queries with 5 training instances per SCM.}
Qwen2.5-7B-Instruct is fine-tuned on \textbf{5} \textsc{Formal Nonsense} instances per SCM, using 250 SCMs in total. Evaluation uses a held-out GPT-5-generated benchmark with 20 SCMs and 5 instances per SCM. Columns indicate the number of nodes.}
\label{tab:7b_pass1_5_intervention}
\resizebox{\textwidth}{!}{%
\begin{tabular}{lcccccccccc}
\hline
Training data & 1 & 2 & 3 & 4 & 5 & 6 & 7 & 8 & 9 & 10 \\
\hline
Base & 0.898 & 0.625 & 0.438 & 0.308 & 0.228 & 0.185 & 0.130 & 0.108 & 0.065 & 0.052 \\
7B (self-generation) & 0.932 & 0.647 & \textbf{0.475} & 0.310 & 0.283 & 0.225 & 0.180 & 0.128 & \textbf{0.108} & \textbf{0.078} \\
GPT-5-mini & \textbf{0.960} & 0.655 & 0.472 & \textbf{0.342} & \textbf{0.287} & \textbf{0.242} & 0.175 & \textbf{0.132} & 0.105 & 0.077 \\
GPT-5 & 0.958 & \textbf{0.660} & 0.460 & 0.300 & 0.280 & 0.223 & \textbf{0.183} & 0.112 & 0.093 & 0.062 \\
\hline
\end{tabular}%
}
\end{table}

\begin{table}[H]
\centering
\caption{\textbf{Pass@1 performance on counterfactual queries with 15 training instances per SCM.}
Qwen2.5-7B-Instruct is fine-tuned on \textbf{15} \textsc{Formal Nonsense} instances per SCM, using 250 SCMs in total. Evaluation uses a held-out GPT-5-generated benchmark with 20 SCMs and 5 instances per SCM. Columns indicate the number of nodes.}
\label{tab:7b_pass1_15_counterfactual}
\resizebox{\textwidth}{!}{%
\begin{tabular}{lcccccccccc}
\hline
Training data & 1 & 2 & 3 & 4 & 5 & 6 & 7 & 8 & 9 & 10 \\
\hline
Base & 0.858 & 0.613 & 0.428 & 0.298 & 0.195 & 0.148 & 0.135 & 0.100 & 0.052 & 0.030 \\
7B (self-generation) & 0.900 & 0.653 & 0.533 & 0.347 & 0.295 & 0.197 & 0.175 & 0.132 & \textbf{0.115} & 0.082 \\
GPT-5-mini & 0.928 & 0.715 & \textbf{0.540} & 0.380 & \textbf{0.318} & 0.227 & \textbf{0.180} & 0.117 & 0.095 & 0.077 \\
GPT-5 & \textbf{0.945} & \textbf{0.728} & 0.523 & \textbf{0.395} & 0.292 & \textbf{0.233} & 0.168 & \textbf{0.147} & 0.110 & \textbf{0.092} \\
\hline
\end{tabular}%
}
\end{table}

\begin{table}[H]
\centering
\caption{\textbf{Pass@1 performance on intervention queries with 15 training instances per SCM.}
Qwen2.5-7B-Instruct is fine-tuned on \textbf{15} \textsc{Formal Nonsense} instances per SCM, using 250 SCMs in total. Evaluation uses a held-out GPT-5-generated benchmark with 20 SCMs and 5 instances per SCM. Columns indicate the number of nodes.}
\label{tab:7b_pass1_15_intervention}
\resizebox{\textwidth}{!}{%
\begin{tabular}{lcccccccccc}
\hline
Training data & 1 & 2 & 3 & 4 & 5 & 6 & 7 & 8 & 9 & 10 \\
\hline
Base & 0.895 & 0.625 & 0.415 & 0.293 & 0.257 & 0.148 & 0.130 & 0.092 & 0.060 & 0.043 \\
7B (self-generation) & 0.918 & 0.660 & \textbf{0.508} & \textbf{0.352} & \textbf{0.322} & 0.243 & \textbf{0.195} & 0.140 & 0.107 & 0.087 \\
GPT-5-mini & \textbf{0.972} & 0.675 & 0.480 & 0.343 & 0.293 & \textbf{0.253} & 0.165 & \textbf{0.152} & 0.098 & \textbf{0.090} \\
GPT-5 & 0.958 & \textbf{0.697} & 0.462 & 0.350 & 0.277 & 0.228 & 0.185 & 0.140 & \textbf{0.108} & 0.075 \\
\hline
\end{tabular}%
}
\end{table}

\clearpage

\subsection{Delta-Highlighted 7B Generator Comparisons}
\label{app:7b-green}

The following tables repeat the 7B generator-ablation results while additionally reporting absolute improvements over the base model in parentheses.

\begin{table}[H]
\centering
\caption{\textbf{SCM generator ablation: Pass@1 performance on counterfactual queries with 5 training instances per SCM.}
Values in parentheses report absolute improvement over the base model.}
\label{tab:7b_pass1_5_counterfactual_delta}
\resizebox{\textwidth}{!}{%
\begin{tabular}{lcccccccccc}
\hline
Training data & 1 & 2 & 3 & 4 & 5 & 6 & 7 & 8 & 9 & 10 \\
\hline
Base & 0.88\basedelta & 0.64\basedelta & 0.45\basedelta & 0.30\basedelta & 0.23\basedelta & 0.16\basedelta & 0.10\basedelta & 0.08\basedelta & 0.06\basedelta & 0.05\basedelta \\
7B (self-generation) & 0.88 \textcolor{green!60!black}{(+0.00)} & 0.66 \textcolor{green!60!black}{(+0.03)} & \textbf{0.50} \textcolor{green!60!black}{(+0.05)} & 0.36 \textcolor{green!60!black}{(+0.06)} & 0.27 \textcolor{green!60!black}{(+0.03)} & 0.21 \textcolor{green!60!black}{(+0.06)} & 0.16 \textcolor{green!60!black}{(+0.07)} & 0.11 \textcolor{green!60!black}{(+0.03)} & \textbf{0.09} \textcolor{green!60!black}{(+0.03)} & 0.07 \textcolor{green!60!black}{(+0.02)} \\
GPT-5-mini & 0.91 \textcolor{green!60!black}{(+0.03)} & \textbf{0.72} \textcolor{green!60!black}{(+0.09)} & 0.47 \textcolor{green!60!black}{(+0.02)} & \textbf{0.36} \textcolor{green!60!black}{(+0.06)} & \textbf{0.29} \textcolor{green!60!black}{(+0.06)} & \textbf{0.22} \textcolor{green!60!black}{(+0.06)} & 0.16 \textcolor{green!60!black}{(+0.06)} & \textbf{0.12} \textcolor{green!60!black}{(+0.03)} & 0.08 \textcolor{green!60!black}{(+0.02)} & 0.07 \textcolor{green!60!black}{(+0.02)} \\
GPT-5 & \textbf{0.93} \textcolor{green!60!black}{(+0.05)} & 0.69 \textcolor{green!60!black}{(+0.06)} & 0.48 \textcolor{green!60!black}{(+0.03)} & 0.35 \textcolor{green!60!black}{(+0.05)} & 0.27 \textcolor{green!60!black}{(+0.04)} & 0.19 \textcolor{green!60!black}{(+0.03)} & \textbf{0.18} \textcolor{green!60!black}{(+0.08)} & 0.11 \textcolor{green!60!black}{(+0.03)} & 0.09 \textcolor{green!60!black}{(+0.03)} & \textbf{0.08} \textcolor{green!60!black}{(+0.02)} \\
\hline
\end{tabular}%
}
\end{table}

\begin{table}[H]
\centering
\caption{\textbf{SCM generator ablation: Pass@1 performance on intervention queries with 5 training instances per SCM.}
Values in parentheses report absolute improvement over the base model.}
\label{tab:7b_pass1_5_intervention_delta}
\resizebox{\textwidth}{!}{%
\begin{tabular}{lcccccccccc}
\hline
Training data & 1 & 2 & 3 & 4 & 5 & 6 & 7 & 8 & 9 & 10 \\
\hline
Base & 0.90\basedelta & 0.62\basedelta & 0.44\basedelta & 0.31\basedelta & 0.23\basedelta & 0.18\basedelta & 0.13\basedelta & 0.11\basedelta & 0.07\basedelta & 0.05\basedelta \\
7B (self-generation) & 0.93 \textcolor{green!60!black}{(+0.03)} & 0.65 \textcolor{green!60!black}{(+0.02)} & \textbf{0.47} \textcolor{green!60!black}{(+0.04)} & 0.31 \textcolor{green!60!black}{(+0.00)} & 0.28 \textcolor{green!60!black}{(+0.06)} & 0.23 \textcolor{green!60!black}{(+0.04)} & 0.18 \textcolor{green!60!black}{(+0.05)} & 0.13 \textcolor{green!60!black}{(+0.02)} & \textbf{0.11} \textcolor{green!60!black}{(+0.04)} & \textbf{0.08} \textcolor{green!60!black}{(+0.03)} \\
GPT-5-mini & \textbf{0.96} \textcolor{green!60!black}{(+0.06)} & 0.66 \textcolor{green!60!black}{(+0.03)} & 0.47 \textcolor{green!60!black}{(+0.03)} & \textbf{0.34} \textcolor{green!60!black}{(+0.03)} & \textbf{0.29} \textcolor{green!60!black}{(+0.06)} & \textbf{0.24} \textcolor{green!60!black}{(+0.06)} & 0.18 \textcolor{green!60!black}{(+0.04)} & \textbf{0.13} \textcolor{green!60!black}{(+0.02)} & 0.10 \textcolor{green!60!black}{(+0.04)} & 0.08 \textcolor{green!60!black}{(+0.03)} \\
GPT-5 & 0.96 \textcolor{green!60!black}{(+0.06)} & \textbf{0.66} \textcolor{green!60!black}{(+0.04)} & 0.46 \textcolor{green!60!black}{(+0.02)} & 0.30 \textcolor{green!60!black}{(-0.01)} & 0.28 \textcolor{green!60!black}{(+0.05)} & 0.22 \textcolor{green!60!black}{(+0.04)} & \textbf{0.18} \textcolor{green!60!black}{(+0.05)} & 0.11 \textcolor{green!60!black}{(+0.00)} & 0.09 \textcolor{green!60!black}{(+0.03)} & 0.06 \textcolor{green!60!black}{(+0.01)} \\
\hline
\end{tabular}%
}
\end{table}

\begin{table}[H]
\centering
\caption{\textbf{SCM generator ablation: Pass@1 performance on counterfactual queries with 15 training instances per SCM.}
Values in parentheses report absolute improvement over the base model.}
\label{tab:7b_pass1_15_counterfactual_delta}
\resizebox{\textwidth}{!}{%
\begin{tabular}{lcccccccccc}
\hline
Training data & 1 & 2 & 3 & 4 & 5 & 6 & 7 & 8 & 9 & 10 \\
\hline
Base & 0.86\basedelta & 0.61\basedelta & 0.43\basedelta & 0.30\basedelta & 0.20\basedelta & 0.15\basedelta & 0.14\basedelta & 0.10\basedelta & 0.05\basedelta & 0.03\basedelta \\
7B (self-generation) & 0.90 \textcolor{green!60!black}{(+0.04)} & 0.65 \textcolor{green!60!black}{(+0.04)} & 0.53 \textcolor{green!60!black}{(+0.10)} & 0.35 \textcolor{green!60!black}{(+0.05)} & 0.29 \textcolor{green!60!black}{(+0.10)} & 0.20 \textcolor{green!60!black}{(+0.05)} & 0.18 \textcolor{green!60!black}{(+0.04)} & 0.13 \textcolor{green!60!black}{(+0.03)} & \textbf{0.12} \textcolor{green!60!black}{(+0.06)} & 0.08 \textcolor{green!60!black}{(+0.05)} \\
GPT-5-mini & 0.93 \textcolor{green!60!black}{(+0.07)} & 0.72 \textcolor{green!60!black}{(+0.10)} & \textbf{0.54} \textcolor{green!60!black}{(+0.11)} & 0.38 \textcolor{green!60!black}{(+0.08)} & \textbf{0.32} \textcolor{green!60!black}{(+0.12)} & 0.23 \textcolor{green!60!black}{(+0.08)} & \textbf{0.18} \textcolor{green!60!black}{(+0.04)} & 0.12 \textcolor{green!60!black}{(+0.02)} & 0.10 \textcolor{green!60!black}{(+0.04)} & 0.08 \textcolor{green!60!black}{(+0.05)} \\
GPT-5 & \textbf{0.94} \textcolor{green!60!black}{(+0.09)} & \textbf{0.73} \textcolor{green!60!black}{(+0.12)} & 0.52 \textcolor{green!60!black}{(+0.10)} & \textbf{0.40} \textcolor{green!60!black}{(+0.10)} & 0.29 \textcolor{green!60!black}{(+0.10)} & \textbf{0.23} \textcolor{green!60!black}{(+0.08)} & 0.17 \textcolor{green!60!black}{(+0.03)} & \textbf{0.15} \textcolor{green!60!black}{(+0.05)} & 0.11 \textcolor{green!60!black}{(+0.06)} & \textbf{0.09} \textcolor{green!60!black}{(+0.06)} \\
\hline
\end{tabular}%
}
\end{table}

\begin{table}[H]
\centering
\caption{\textbf{SCM generator ablation: Pass@1 performance on intervention queries with 15 training instances per SCM.}
Values in parentheses report absolute improvement over the base model.}
\label{tab:7b_pass1_15_intervention_delta}
\resizebox{\textwidth}{!}{%
\begin{tabular}{lcccccccccc}
\hline
Training data & 1 & 2 & 3 & 4 & 5 & 6 & 7 & 8 & 9 & 10 \\
\hline
Base & 0.90\basedelta & 0.62\basedelta & 0.41\basedelta & 0.29\basedelta & 0.26\basedelta & 0.15\basedelta & 0.13\basedelta & 0.09\basedelta & 0.06\basedelta & 0.04\basedelta \\
7B (self-generation) & 0.92 \textcolor{green!60!black}{(+0.02)} & 0.66 \textcolor{green!60!black}{(+0.04)} & \textbf{0.51} \textcolor{green!60!black}{(+0.09)} & \textbf{0.35} \textcolor{green!60!black}{(+0.06)} & \textbf{0.32} \textcolor{green!60!black}{(+0.07)} & 0.24 \textcolor{green!60!black}{(+0.09)} & \textbf{0.20} \textcolor{green!60!black}{(+0.07)} & 0.14 \textcolor{green!60!black}{(+0.05)} & 0.11 \textcolor{green!60!black}{(+0.05)} & 0.09 \textcolor{green!60!black}{(+0.04)} \\
GPT-5-mini & \textbf{0.97} \textcolor{green!60!black}{(+0.08)} & 0.68 \textcolor{green!60!black}{(+0.05)} & 0.48 \textcolor{green!60!black}{(+0.07)} & 0.34 \textcolor{green!60!black}{(+0.05)} & 0.29 \textcolor{green!60!black}{(+0.04)} & \textbf{0.25} \textcolor{green!60!black}{(+0.10)} & 0.17 \textcolor{green!60!black}{(+0.04)} & \textbf{0.15} \textcolor{green!60!black}{(+0.06)} & 0.10 \textcolor{green!60!black}{(+0.04)} & \textbf{0.09} \textcolor{green!60!black}{(+0.05)} \\
GPT-5 & 0.96 \textcolor{green!60!black}{(+0.06)} & \textbf{0.70} \textcolor{green!60!black}{(+0.07)} & 0.46 \textcolor{green!60!black}{(+0.05)} & 0.35 \textcolor{green!60!black}{(+0.06)} & 0.28 \textcolor{green!60!black}{(+0.02)} & 0.23 \textcolor{green!60!black}{(+0.08)} & 0.18 \textcolor{green!60!black}{(+0.06)} & 0.14 \textcolor{green!60!black}{(+0.05)} & \textbf{0.11} \textcolor{green!60!black}{(+0.05)} & 0.07 \textcolor{green!60!black}{(+0.03)} \\
\hline
\end{tabular}%
}
\end{table}

\clearpage

\section{Informal Representation Experiments}
\label{app:informal-experiments}

This section reports additional experiments evaluating transfer to informal SCM representations. Unless otherwise specified, all entries are pass@1 accuracies and columns indicate the number of nodes.

\begin{table}[H]
\centering
\caption{\textbf{Pass@1 performance on \textsc{Informal-Medical} causal queries, averaged over selected prompt modes.}}
\label{tab:informal_medical_avg_raw}
\small
\setlength{\tabcolsep}{4pt}
\renewcommand{\arraystretch}{1.1}
\resizebox{0.9\textwidth}{!}{%
\begin{tabular}{p{5.7cm}cccccccccc}
\hline
Variant & 1 & 2 & 3 & 4 & 5 & 6 & 7 & 8 & 9 & 10 \\
\hline
Base & 0.310 & 0.230 & 0.175 & 0.100 & 0.075 & 0.075 & 0.105 & 0.085 & 0.055 & 0.020 \\
Train-135 (informal, non-sense) & 0.375 & 0.305 & 0.285 & 0.225 & 0.170 & 0.145 & 0.120 & 0.140 & 0.090 & 0.085 \\
Train-135 (formal, non-sense) & \textbf{0.535} & 0.400 & \textbf{0.470} & \textbf{0.305} & 0.270 & 0.240 & 0.270 & 0.225 & 0.220 & 0.155 \\
Train-135 (formal $\rightarrow$ informal, non-sense) & 0.500 & \textbf{0.410} & 0.420 & 0.285 & \textbf{0.395} & \textbf{0.305} & \textbf{0.340} & \textbf{0.310} & \textbf{0.270} & \textbf{0.205} \\
\hline
\end{tabular}%
}
\end{table}

\begin{table}[H]
\centering
\caption{\textbf{Pass@1 performance on \textsc{Informal-Medical} counterfactual queries.}}
\label{tab:informal_medical_cf_raw}
\small
\setlength{\tabcolsep}{4pt}
\renewcommand{\arraystretch}{1.1}
\resizebox{0.9\textwidth}{!}{%
\begin{tabular}{p{5.7cm}cccccccccc}
\hline
Variant & 1 & 2 & 3 & 4 & 5 & 6 & 7 & 8 & 9 & 10 \\
\hline
Base & 0.310 & 0.190 & 0.140 & 0.070 & 0.080 & 0.080 & 0.090 & 0.020 & 0.040 & 0.020 \\
Train-135 (informal, non-sense) & 0.380 & 0.320 & 0.340 & 0.230 & 0.210 & 0.120 & 0.130 & 0.150 & 0.070 & 0.100 \\
Train-135 (formal, non-sense) & \textbf{0.530} & 0.450 & 0.480 & \textbf{0.330} & 0.340 & 0.260 & 0.290 & 0.240 & 0.240 & 0.170 \\
Train-135 (formal $\rightarrow$ informal, non-sense) & 0.480 & \textbf{0.490} & \textbf{0.490} & 0.320 & \textbf{0.440} & \textbf{0.310} & \textbf{0.380} & \textbf{0.340} & \textbf{0.290} & \textbf{0.250} \\
\hline
\end{tabular}%
}
\end{table}

\begin{table}[H]
\centering
\caption{\textbf{Pass@1 performance on \textsc{Informal-Medical} intervention queries.}}
\label{tab:informal_medical_intervention_raw}
\small
\setlength{\tabcolsep}{4pt}
\renewcommand{\arraystretch}{1.1}
\resizebox{0.9\textwidth}{!}{%
\begin{tabular}{p{5.7cm}cccccccccc}
\hline
Variant & 1 & 2 & 3 & 4 & 5 & 6 & 7 & 8 & 9 & 10 \\
\hline
Base & 0.310 & 0.270 & 0.210 & 0.130 & 0.070 & 0.070 & 0.120 & 0.150 & 0.070 & 0.020 \\
Train-135 (informal, non-sense) & 0.370 & 0.290 & 0.230 & 0.220 & 0.130 & 0.170 & 0.110 & 0.130 & 0.110 & 0.070 \\
Train-135 (formal, non-sense) & \textbf{0.540} & \textbf{0.350} & \textbf{0.460} & \textbf{0.280} & 0.200 & 0.220 & 0.250 & 0.210 & 0.200 & 0.140 \\
Train-135 (formal $\rightarrow$ informal, non-sense) & 0.520 & 0.330 & 0.350 & 0.250 & \textbf{0.350} & \textbf{0.300} & \textbf{0.300} & \textbf{0.280} & \textbf{0.250} & \textbf{0.160} \\
\hline
\end{tabular}%
}
\end{table}

\clearpage

\subsection{Delta-Highlighted \textsc{Informal-Medical} Out-of-Distribution Results}
\label{app:informal-medical-green}

\begin{table}[H]
\centering
\caption{\textbf{Pass@1 performance on \textsc{Informal-Medical} out-of-distribution causal queries, averaged over selected prompt modes.}
All models use Qwen2.5-3B-Instruct. Train-135 (formal) and Train-135 (informal) are fine-tuned on 135 instances per SCM, using 180 SCMs in total. Train-135 (formal $\rightarrow$ informal) is trained sequentially on 135 formal and then 135 informal instances per SCM. Values in parentheses report absolute improvement over the base model.}
\label{tab:informal_medical_avg_delta}
\resizebox{\textwidth}{!}{%
\begin{tabular}{lcccccccccc}
\hline
Training data & 1 & 2 & 3 & 4 & 5 & 6 & 7 & 8 & 9 & 10 \\
\hline
Base & 0.31\basedelta & 0.23\basedelta & 0.18\basedelta & 0.10\basedelta & 0.08\basedelta & 0.08\basedelta & 0.11\basedelta & 0.09\basedelta & 0.06\basedelta & 0.02\basedelta \\
Train-135 (informal, non-sense) & 0.38 \textcolor{green!60!black}{(+0.07)} & 0.31 \textcolor{green!60!black}{(+0.08)} & 0.29 \textcolor{green!60!black}{(+0.11)} & 0.23 \textcolor{green!60!black}{(+0.13)} & 0.17 \textcolor{green!60!black}{(+0.10)} & 0.15 \textcolor{green!60!black}{(+0.07)} & 0.12 \textcolor{green!60!black}{(+0.02)} & 0.14 \textcolor{green!60!black}{(+0.06)} & 0.09 \textcolor{green!60!black}{(+0.04)} & 0.09 \textcolor{green!60!black}{(+0.07)} \\
Train-135 (formal, non-sense) & \textbf{0.54} \textcolor{green!60!black}{(+0.23)} & 0.40 \textcolor{green!60!black}{(+0.17)} & \textbf{0.47} \textcolor{green!60!black}{(+0.30)} & \textbf{0.31} \textcolor{green!60!black}{(+0.21)} & 0.27 \textcolor{green!60!black}{(+0.20)} & 0.24 \textcolor{green!60!black}{(+0.17)} & 0.27 \textcolor{green!60!black}{(+0.17)} & 0.23 \textcolor{green!60!black}{(+0.14)} & 0.22 \textcolor{green!60!black}{(+0.17)} & 0.16 \textcolor{green!60!black}{(+0.14)} \\
Train-135 (formal $\rightarrow$ informal, non-sense) & 0.50 \textcolor{green!60!black}{(+0.19)} & \textbf{0.41} \textcolor{green!60!black}{(+0.18)} & 0.42 \textcolor{green!60!black}{(+0.25)} & 0.29 \textcolor{green!60!black}{(+0.19)} & \textbf{0.40} \textcolor{green!60!black}{(+0.32)} & \textbf{0.31} \textcolor{green!60!black}{(+0.23)} & \textbf{0.34} \textcolor{green!60!black}{(+0.24)} & \textbf{0.31} \textcolor{green!60!black}{(+0.23)} & \textbf{0.27} \textcolor{green!60!black}{(+0.22)} & \textbf{0.21} \textcolor{green!60!black}{(+0.19)} \\
\hline
\end{tabular}%
}
\end{table}

\begin{table}[H]
\centering
\caption{\textbf{Pass@1 performance on \textsc{Informal-Medical} out-of-distribution counterfactual queries.}
Values in parentheses report absolute improvement over the base model.}
\label{tab:informal_medical_cf_delta}
\resizebox{\textwidth}{!}{%
\begin{tabular}{lcccccccccc}
\hline
Training data & 1 & 2 & 3 & 4 & 5 & 6 & 7 & 8 & 9 & 10 \\
\hline
Base & 0.31\basedelta & 0.19\basedelta & 0.14\basedelta & 0.07\basedelta & 0.08\basedelta & 0.08\basedelta & 0.09\basedelta & 0.02\basedelta & 0.04\basedelta & 0.02\basedelta \\
Train-135 (informal, non-sense) & 0.38 \textcolor{green!60!black}{(+0.07)} & 0.32 \textcolor{green!60!black}{(+0.13)} & 0.34 \textcolor{green!60!black}{(+0.20)} & 0.23 \textcolor{green!60!black}{(+0.16)} & 0.21 \textcolor{green!60!black}{(+0.13)} & 0.12 \textcolor{green!60!black}{(+0.04)} & 0.13 \textcolor{green!60!black}{(+0.04)} & 0.15 \textcolor{green!60!black}{(+0.13)} & 0.07 \textcolor{green!60!black}{(+0.03)} & 0.10 \textcolor{green!60!black}{(+0.08)} \\
Train-135 (formal, non-sense) & \textbf{0.53} \textcolor{green!60!black}{(+0.22)} & 0.45 \textcolor{green!60!black}{(+0.26)} & 0.48 \textcolor{green!60!black}{(+0.34)} & \textbf{0.33} \textcolor{green!60!black}{(+0.26)} & 0.34 \textcolor{green!60!black}{(+0.26)} & 0.26 \textcolor{green!60!black}{(+0.18)} & 0.29 \textcolor{green!60!black}{(+0.20)} & 0.24 \textcolor{green!60!black}{(+0.22)} & 0.24 \textcolor{green!60!black}{(+0.20)} & 0.17 \textcolor{green!60!black}{(+0.15)} \\
Train-135 (formal $\rightarrow$ informal, non-sense) & 0.48 \textcolor{green!60!black}{(+0.17)} & \textbf{0.49} \textcolor{green!60!black}{(+0.30)} & \textbf{0.49} \textcolor{green!60!black}{(+0.35)} & 0.32 \textcolor{green!60!black}{(+0.25)} & \textbf{0.44} \textcolor{green!60!black}{(+0.36)} & \textbf{0.31} \textcolor{green!60!black}{(+0.23)} & \textbf{0.38} \textcolor{green!60!black}{(+0.29)} & \textbf{0.34} \textcolor{green!60!black}{(+0.32)} & \textbf{0.29} \textcolor{green!60!black}{(+0.25)} & \textbf{0.25} \textcolor{green!60!black}{(+0.23)} \\
\hline
\end{tabular}%
}
\end{table}

\begin{table}[H]
\centering
\caption{\textbf{Pass@1 performance on \textsc{Informal-Medical} out-of-distribution intervention queries.}
Values in parentheses report absolute improvement over the base model.}
\label{tab:informal_medical_intervention_delta}
\resizebox{\textwidth}{!}{%
\begin{tabular}{lcccccccccc}
\hline
Training data & 1 & 2 & 3 & 4 & 5 & 6 & 7 & 8 & 9 & 10 \\
\hline
Base & 0.31\basedelta & 0.27\basedelta & 0.21\basedelta & 0.13\basedelta & 0.07\basedelta & 0.07\basedelta & 0.12\basedelta & 0.15\basedelta & 0.07\basedelta & 0.02\basedelta \\
Train-135 (informal, non-sense) & 0.37 \textcolor{green!60!black}{(+0.06)} & 0.29 \textcolor{green!60!black}{(+0.02)} & 0.23 \textcolor{green!60!black}{(+0.02)} & 0.22 \textcolor{green!60!black}{(+0.09)} & 0.13 \textcolor{green!60!black}{(+0.06)} & 0.17 \textcolor{green!60!black}{(+0.10)} & 0.11 \textcolor{green!60!black}{(-0.01)} & 0.13 \textcolor{green!60!black}{(-0.02)} & 0.11 \textcolor{green!60!black}{(+0.04)} & 0.07 \textcolor{green!60!black}{(+0.05)} \\
Train-135 (formal, non-sense) & \textbf{0.54} \textcolor{green!60!black}{(+0.23)} & \textbf{0.35} \textcolor{green!60!black}{(+0.08)} & \textbf{0.46} \textcolor{green!60!black}{(+0.25)} & \textbf{0.28} \textcolor{green!60!black}{(+0.15)} & 0.20 \textcolor{green!60!black}{(+0.13)} & 0.22 \textcolor{green!60!black}{(+0.15)} & 0.25 \textcolor{green!60!black}{(+0.13)} & 0.21 \textcolor{green!60!black}{(+0.06)} & 0.20 \textcolor{green!60!black}{(+0.13)} & 0.14 \textcolor{green!60!black}{(+0.12)} \\
Train-135 (formal $\rightarrow$ informal, non-sense) & 0.52 \textcolor{green!60!black}{(+0.21)} & 0.33 \textcolor{green!60!black}{(+0.06)} & 0.35 \textcolor{green!60!black}{(+0.14)} & 0.25 \textcolor{green!60!black}{(+0.12)} & \textbf{0.35} \textcolor{green!60!black}{(+0.28)} & \textbf{0.30} \textcolor{green!60!black}{(+0.23)} & \textbf{0.30} \textcolor{green!60!black}{(+0.18)} & \textbf{0.28} \textcolor{green!60!black}{(+0.13)} & \textbf{0.25} \textcolor{green!60!black}{(+0.18)} & \textbf{0.16} \textcolor{green!60!black}{(+0.14)} \\
\hline
\end{tabular}%
}
\end{table}

\clearpage

\subsection{Informal-to-Informal Transfer on In-Domain Nonsensical SCMs: Inverted-Star Topology}
\label{app:informal-in-domain-inverted}

\begin{table}[H]
\centering
\caption{\textbf{Pass@1 performance on in-domain informal nonsensical SCMs with inverted-star topology, averaged over selected prompt modes.}
Values in parentheses report absolute improvement over the base model.}
\label{tab:informal_indomain_inverted_avg_delta}
\resizebox{\textwidth}{!}{%
\begin{tabular}{lcccccccccc}
\hline
Training data & 1 & 2 & 3 & 4 & 5 & 6 & 7 & 8 & 9 & 10 \\
\hline
Base & 0.67\basedelta & 0.41\basedelta & 0.21\basedelta & 0.15\basedelta & 0.10\basedelta & 0.09\basedelta & 0.04\basedelta & 0.04\basedelta & 0.03\basedelta & 0.02\basedelta \\
Train-135 (informal, non-sense) & \textbf{1.00} \textcolor{green!60!black}{(+0.33)} & 0.63 \textcolor{green!60!black}{(+0.22)} & 0.48 \textcolor{green!60!black}{(+0.27)} & 0.35 \textcolor{green!60!black}{(+0.20)} & 0.27 \textcolor{green!60!black}{(+0.17)} & 0.21 \textcolor{green!60!black}{(+0.12)} & 0.17 \textcolor{green!60!black}{(+0.13)} & 0.16 \textcolor{green!60!black}{(+0.12)} & 0.05 \textcolor{green!60!black}{(+0.03)} & 0.07 \textcolor{green!60!black}{(+0.05)} \\
Train-135 (formal, non-sense) & 0.93 \textcolor{green!60!black}{(+0.26)} & 0.67 \textcolor{green!60!black}{(+0.26)} & 0.59 \textcolor{green!60!black}{(+0.39)} & 0.54 \textcolor{green!60!black}{(+0.39)} & 0.43 \textcolor{green!60!black}{(+0.33)} & 0.43 \textcolor{green!60!black}{(+0.35)} & 0.40 \textcolor{green!60!black}{(+0.36)} & 0.31 \textcolor{green!60!black}{(+0.28)} & 0.20 \textcolor{green!60!black}{(+0.17)} & 0.20 \textcolor{green!60!black}{(+0.18)} \\
Train-135 (formal $\rightarrow$ informal, non-sense) & 0.98 \textcolor{green!60!black}{(+0.31)} & \textbf{0.75} \textcolor{green!60!black}{(+0.34)} & \textbf{0.71} \textcolor{green!60!black}{(+0.51)} & \textbf{0.68} \textcolor{green!60!black}{(+0.53)} & \textbf{0.59} \textcolor{green!60!black}{(+0.49)} & \textbf{0.59} \textcolor{green!60!black}{(+0.51)} & \textbf{0.55} \textcolor{green!60!black}{(+0.51)} & \textbf{0.52} \textcolor{green!60!black}{(+0.49)} & \textbf{0.38} \textcolor{green!60!black}{(+0.35)} & \textbf{0.41} \textcolor{green!60!black}{(+0.39)} \\
\hline
\end{tabular}%
}
\end{table}

\begin{table}[H]
\centering
\caption{\textbf{Pass@1 performance on in-domain informal nonsensical SCMs with inverted-star topology: counterfactual queries.}
Values in parentheses report absolute improvement over the base model.}
\label{tab:informal_indomain_inverted_cf_delta}
\resizebox{\textwidth}{!}{%
\begin{tabular}{lcccccccccc}
\hline
Training data & 1 & 2 & 3 & 4 & 5 & 6 & 7 & 8 & 9 & 10 \\
\hline
Base & 0.72\basedelta & 0.42\basedelta & 0.18\basedelta & 0.15\basedelta & 0.09\basedelta & 0.06\basedelta & 0.01\basedelta & 0.04\basedelta & 0.03\basedelta & 0.01\basedelta \\
Train-135 (informal, non-sense) & \textbf{1.00} \textcolor{green!60!black}{(+0.28)} & 0.70 \textcolor{green!60!black}{(+0.28)} & 0.40 \textcolor{green!60!black}{(+0.22)} & 0.38 \textcolor{green!60!black}{(+0.23)} & 0.22 \textcolor{green!60!black}{(+0.13)} & 0.13 \textcolor{green!60!black}{(+0.07)} & 0.20 \textcolor{green!60!black}{(+0.19)} & 0.13 \textcolor{green!60!black}{(+0.09)} & 0.03 \textcolor{green!60!black}{(+0.00)} & 0.06 \textcolor{green!60!black}{(+0.05)} \\
Train-135 (formal, non-sense) & 0.93 \textcolor{green!60!black}{(+0.21)} & 0.71 \textcolor{green!60!black}{(+0.29)} & 0.64 \textcolor{green!60!black}{(+0.46)} & 0.55 \textcolor{green!60!black}{(+0.40)} & 0.39 \textcolor{green!60!black}{(+0.30)} & 0.46 \textcolor{green!60!black}{(+0.40)} & 0.41 \textcolor{green!60!black}{(+0.40)} & 0.29 \textcolor{green!60!black}{(+0.25)} & 0.23 \textcolor{green!60!black}{(+0.20)} & 0.18 \textcolor{green!60!black}{(+0.17)} \\
Train-135 (formal $\rightarrow$ informal, non-sense) & \textbf{1.00} \textcolor{green!60!black}{(+0.28)} & \textbf{0.77} \textcolor{green!60!black}{(+0.35)} & \textbf{0.70} \textcolor{green!60!black}{(+0.52)} & \textbf{0.71} \textcolor{green!60!black}{(+0.56)} & \textbf{0.59} \textcolor{green!60!black}{(+0.50)} & \textbf{0.65} \textcolor{green!60!black}{(+0.59)} & \textbf{0.55} \textcolor{green!60!black}{(+0.54)} & \textbf{0.57} \textcolor{green!60!black}{(+0.53)} & \textbf{0.36} \textcolor{green!60!black}{(+0.33)} & \textbf{0.41} \textcolor{green!60!black}{(+0.40)} \\
\hline
\end{tabular}%
}
\end{table}

\begin{table}[H]
\centering
\caption{\textbf{Pass@1 performance on in-domain informal nonsensical SCMs with inverted-star topology: intervention queries.}
Values in parentheses report absolute improvement over the base model.}
\label{tab:informal_indomain_inverted_intervention_delta}
\resizebox{\textwidth}{!}{%
\begin{tabular}{lcccccccccc}
\hline
Training data & 1 & 2 & 3 & 4 & 5 & 6 & 7 & 8 & 9 & 10 \\
\hline
Base & 0.62\basedelta & 0.40\basedelta & 0.23\basedelta & 0.15\basedelta & 0.11\basedelta & 0.11\basedelta & 0.07\basedelta & 0.03\basedelta & 0.02\basedelta & 0.03\basedelta \\
Train-135 (informal, non-sense) & \textbf{0.99} \textcolor{green!60!black}{(+0.37)} & 0.55 \textcolor{green!60!black}{(+0.15)} & 0.55 \textcolor{green!60!black}{(+0.32)} & 0.32 \textcolor{green!60!black}{(+0.17)} & 0.32 \textcolor{green!60!black}{(+0.21)} & 0.28 \textcolor{green!60!black}{(+0.17)} & 0.14 \textcolor{green!60!black}{(+0.07)} & 0.18 \textcolor{green!60!black}{(+0.15)} & 0.07 \textcolor{green!60!black}{(+0.05)} & 0.08 \textcolor{green!60!black}{(+0.05)} \\
Train-135 (formal, non-sense) & 0.92 \textcolor{green!60!black}{(+0.30)} & 0.63 \textcolor{green!60!black}{(+0.23)} & 0.54 \textcolor{green!60!black}{(+0.31)} & 0.52 \textcolor{green!60!black}{(+0.37)} & 0.47 \textcolor{green!60!black}{(+0.36)} & 0.40 \textcolor{green!60!black}{(+0.29)} & 0.39 \textcolor{green!60!black}{(+0.32)} & 0.33 \textcolor{green!60!black}{(+0.30)} & 0.16 \textcolor{green!60!black}{(+0.14)} & 0.22 \textcolor{green!60!black}{(+0.19)} \\
Train-135 (formal $\rightarrow$ informal, non-sense) & 0.96 \textcolor{green!60!black}{(+0.34)} & \textbf{0.73} \textcolor{green!60!black}{(+0.33)} & \textbf{0.72} \textcolor{green!60!black}{(+0.49)} & \textbf{0.65} \textcolor{green!60!black}{(+0.50)} & \textbf{0.59} \textcolor{green!60!black}{(+0.48)} & \textbf{0.53} \textcolor{green!60!black}{(+0.42)} & \textbf{0.54} \textcolor{green!60!black}{(+0.47)} & \textbf{0.47} \textcolor{green!60!black}{(+0.44)} & \textbf{0.39} \textcolor{green!60!black}{(+0.37)} & \textbf{0.40} \textcolor{green!60!black}{(+0.37)} \\
\hline
\end{tabular}%
}
\end{table}

\clearpage

\subsection{Informal-to-Informal Transfer on In-Domain Nonsensical SCMs: Combined Topologies}
\label{app:informal-in-domain-combined}

\begin{table}[H]
\centering
\caption{\textbf{Pass@1 performance on in-domain informal nonsensical SCMs, combined across all prompt modes.}
Values in parentheses report absolute improvement over the base model.}
\label{tab:informal_indomain_combined_all_modes_delta}
\resizebox{\textwidth}{!}{%
\begin{tabular}{lcccccccccc}
\hline
Training data & 1 & 2 & 3 & 4 & 5 & 6 & 7 & 8 & 9 & 10 \\
\hline
Base & 0.58\basedelta & 0.33\basedelta & 0.18\basedelta & 0.11\basedelta & 0.07\basedelta & 0.05\basedelta & 0.05\basedelta & 0.04\basedelta & 0.03\basedelta & 0.02\basedelta \\
Train-135 (informal, non-sense) & \textbf{0.78} \textcolor{green!60!black}{(+0.20)} & 0.51 \textcolor{green!60!black}{(+0.19)} & 0.34 \textcolor{green!60!black}{(+0.16)} & 0.24 \textcolor{green!60!black}{(+0.13)} & 0.19 \textcolor{green!60!black}{(+0.11)} & 0.15 \textcolor{green!60!black}{(+0.09)} & 0.11 \textcolor{green!60!black}{(+0.06)} & 0.10 \textcolor{green!60!black}{(+0.06)} & 0.06 \textcolor{green!60!black}{(+0.03)} & 0.07 \textcolor{green!60!black}{(+0.05)} \\
Train-135 (formal, non-sense) & 0.68 \textcolor{green!60!black}{(+0.10)} & 0.49 \textcolor{green!60!black}{(+0.16)} & 0.38 \textcolor{green!60!black}{(+0.19)} & 0.27 \textcolor{green!60!black}{(+0.16)} & 0.21 \textcolor{green!60!black}{(+0.14)} & 0.20 \textcolor{green!60!black}{(+0.14)} & 0.18 \textcolor{green!60!black}{(+0.13)} & 0.15 \textcolor{green!60!black}{(+0.11)} & 0.12 \textcolor{green!60!black}{(+0.09)} & 0.11 \textcolor{green!60!black}{(+0.09)} \\
Train-135 (formal $\rightarrow$ informal, non-sense) & 0.73 \textcolor{green!60!black}{(+0.14)} & \textbf{0.55} \textcolor{green!60!black}{(+0.22)} & \textbf{0.41} \textcolor{green!60!black}{(+0.23)} & \textbf{0.32} \textcolor{green!60!black}{(+0.21)} & \textbf{0.31} \textcolor{green!60!black}{(+0.23)} & \textbf{0.27} \textcolor{green!60!black}{(+0.22)} & \textbf{0.25} \textcolor{green!60!black}{(+0.20)} & \textbf{0.22} \textcolor{green!60!black}{(+0.18)} & \textbf{0.18} \textcolor{green!60!black}{(+0.15)} & \textbf{0.18} \textcolor{green!60!black}{(+0.15)} \\
\hline
\end{tabular}%
}
\end{table}

\begin{table}[H]
\centering
\caption{\textbf{Pass@1 performance on in-domain informal nonsensical SCMs: counterfactual queries.}
Values in parentheses report absolute improvement over the base model.}
\label{tab:informal_indomain_combined_cf_delta}
\resizebox{\textwidth}{!}{%
\begin{tabular}{lcccccccccc}
\hline
Training data & 1 & 2 & 3 & 4 & 5 & 6 & 7 & 8 & 9 & 10 \\
\hline
Base & 0.59\basedelta & 0.34\basedelta & 0.18\basedelta & 0.10\basedelta & 0.07\basedelta & 0.05\basedelta & 0.04\basedelta & 0.02\basedelta & 0.02\basedelta & 0.02\basedelta \\
Train-135 (informal, non-sense) & \textbf{0.77} \textcolor{green!60!black}{(+0.18)} & 0.55 \textcolor{green!60!black}{(+0.21)} & 0.34 \textcolor{green!60!black}{(+0.17)} & 0.27 \textcolor{green!60!black}{(+0.16)} & 0.19 \textcolor{green!60!black}{(+0.13)} & 0.14 \textcolor{green!60!black}{(+0.09)} & 0.12 \textcolor{green!60!black}{(+0.08)} & 0.10 \textcolor{green!60!black}{(+0.07)} & 0.06 \textcolor{green!60!black}{(+0.04)} & 0.08 \textcolor{green!60!black}{(+0.05)} \\
Train-135 (formal, non-sense) & 0.68 \textcolor{green!60!black}{(+0.09)} & 0.52 \textcolor{green!60!black}{(+0.18)} & 0.40 \textcolor{green!60!black}{(+0.23)} & 0.29 \textcolor{green!60!black}{(+0.18)} & 0.23 \textcolor{green!60!black}{(+0.17)} & 0.20 \textcolor{green!60!black}{(+0.16)} & 0.19 \textcolor{green!60!black}{(+0.15)} & 0.15 \textcolor{green!60!black}{(+0.13)} & 0.14 \textcolor{green!60!black}{(+0.12)} & 0.12 \textcolor{green!60!black}{(+0.09)} \\
Train-135 (formal $\rightarrow$ informal, non-sense) & 0.74 \textcolor{green!60!black}{(+0.15)} & \textbf{0.59} \textcolor{green!60!black}{(+0.26)} & \textbf{0.44} \textcolor{green!60!black}{(+0.26)} & \textbf{0.35} \textcolor{green!60!black}{(+0.24)} & \textbf{0.32} \textcolor{green!60!black}{(+0.26)} & \textbf{0.30} \textcolor{green!60!black}{(+0.25)} & \textbf{0.26} \textcolor{green!60!black}{(+0.22)} & \textbf{0.24} \textcolor{green!60!black}{(+0.22)} & \textbf{0.19} \textcolor{green!60!black}{(+0.17)} & \textbf{0.19} \textcolor{green!60!black}{(+0.16)} \\
\hline
\end{tabular}%
}
\end{table}

\begin{table}[H]
\centering
\caption{\textbf{Pass@1 performance on in-domain informal nonsensical SCMs: intervention queries.}
Values in parentheses report absolute improvement over the base model.}
\label{tab:informal_indomain_combined_intervention_delta}
\resizebox{\textwidth}{!}{%
\begin{tabular}{lcccccccccc}
\hline
Training data & 1 & 2 & 3 & 4 & 5 & 6 & 7 & 8 & 9 & 10 \\
\hline
Base & 0.58\basedelta & 0.32\basedelta & 0.19\basedelta & 0.11\basedelta & 0.08\basedelta & 0.06\basedelta & 0.06\basedelta & 0.06\basedelta & 0.03\basedelta & 0.02\basedelta \\
Train-135 (informal, non-sense) & \textbf{0.78} \textcolor{green!60!black}{(+0.21)} & 0.48 \textcolor{green!60!black}{(+0.16)} & 0.35 \textcolor{green!60!black}{(+0.15)} & 0.22 \textcolor{green!60!black}{(+0.11)} & 0.18 \textcolor{green!60!black}{(+0.10)} & 0.15 \textcolor{green!60!black}{(+0.09)} & 0.10 \textcolor{green!60!black}{(+0.04)} & 0.09 \textcolor{green!60!black}{(+0.04)} & 0.07 \textcolor{green!60!black}{(+0.03)} & 0.06 \textcolor{green!60!black}{(+0.04)} \\
Train-135 (formal, non-sense) & 0.69 \textcolor{green!60!black}{(+0.12)} & 0.47 \textcolor{green!60!black}{(+0.15)} & 0.35 \textcolor{green!60!black}{(+0.16)} & 0.25 \textcolor{green!60!black}{(+0.14)} & 0.19 \textcolor{green!60!black}{(+0.11)} & 0.19 \textcolor{green!60!black}{(+0.13)} & 0.17 \textcolor{green!60!black}{(+0.11)} & 0.14 \textcolor{green!60!black}{(+0.08)} & 0.10 \textcolor{green!60!black}{(+0.06)} & 0.11 \textcolor{green!60!black}{(+0.09)} \\
Train-135 (formal $\rightarrow$ informal, non-sense) & 0.72 \textcolor{green!60!black}{(+0.14)} & \textbf{0.50} \textcolor{green!60!black}{(+0.18)} & \textbf{0.38} \textcolor{green!60!black}{(+0.19)} & \textbf{0.29} \textcolor{green!60!black}{(+0.18)} & \textbf{0.29} \textcolor{green!60!black}{(+0.21)} & \textbf{0.24} \textcolor{green!60!black}{(+0.18)} & \textbf{0.23} \textcolor{green!60!black}{(+0.17)} & \textbf{0.20} \textcolor{green!60!black}{(+0.14)} & \textbf{0.17} \textcolor{green!60!black}{(+0.13)} & \textbf{0.16} \textcolor{green!60!black}{(+0.14)} \\
\hline
\end{tabular}%
}
\end{table}

\clearpage

\section{Additional Benchmark and Ablation Results}
\label{app:additional-benchmarks}

This section provides additional benchmark results and diagnostic analyses used to support the main empirical claims.

\subsection{Formalized External Benchmarks}
\label{app:external_benchmarks}

We evaluate the \textsc{nonsense}-\textit{code}-trained model on independently constructed SCMs derived from GSM8K and the ASIA, CANCER, and SURVEY Bayesian networks. The dataset configurations are given in Appendix~\ref{app:experiment_configurations}, with formalization prompts in Appendices~\ref{app:gsm8k_formalization} and~\ref{app:bn_formalization}.

\begin{table}[H]
\centering
\caption{\textbf{Re-Imagine-derived benchmark.}
Pass@1 performance on the formalized GSM8K evaluation set (configuration in Table~\ref{tab:generalization_setup}). We compare the base Qwen2.5-3B-Instruct model with a version fine-tuned on 135 \textsc{nonsense} instances per SCM. The training set includes 180 SCMs for counterfactual and intervention queries; we additionally assess generalization to abduction and deduction tasks.}
\label{tab:gsm8k_scm}
\begin{tabular}{lccccc}
\toprule
Model & Overall & Abd. & Ctf. & Ded. & Int. \\
\midrule
Base & 0.331 & 0.654 & 0.231 & 0.393 & 0.141 \\
Train-135 & \textbf{0.396} & \textbf{0.710} & \textbf{0.364} & \textbf{0.448} & \textbf{0.158} \\
\bottomrule
\end{tabular}
\end{table}

\begin{table}[H]
\centering
\caption{\textbf{Real-world causal graphs from bnlearn.}
Pass@1 performance on bnlearn Bayesian networks converted to formal SCMs. We compare the base Qwen2.5-3B-Instruct model with a model fine-tuned on 135 \textsc{nonsense} queries per SCM.}
\label{tab:real_causal_graphs}
\begin{tabular}{lcccccc}
\toprule
& \multicolumn{2}{c}{Overall} & \multicolumn{2}{c}{Counterfactual} & \multicolumn{2}{c}{Intervention} \\
\cmidrule(lr){2-3} \cmidrule(lr){4-5} \cmidrule(lr){6-7}
Graph & Base & Train-135 & Base & Train-135 & Base & Train-135 \\
\midrule
ASIA & 0.040 & \textbf{0.155} & 0.060 & \textbf{0.190} & 0.020 & \textbf{0.120} \\
CANCER & 0.065 & \textbf{0.115} & 0.070 & \textbf{0.150} & 0.060 & \textbf{0.080} \\
SURVEY & 0.010 & \textbf{0.055} & 0.020 & \textbf{0.080} & 0.000 & \textbf{0.030} \\
\bottomrule
\end{tabular}
\end{table}

\subsection{Partially Specified GSM8K}
\label{app:partial_gsm8k}

This evaluation tests transfer beyond fully specified executable inputs. We present GSM8K-derived SCMs as unstructured natural-language problems with partially specified mechanisms; a representative input is provided in Appendix~\ref{app:partial_gsm8k_example}. The evaluation configuration contains 262 SCMs and 500 intervention and counterfactual queries (Table~\ref{tab:generalization_setup}). We compare the base \texttt{Qwen2.5-3B-Instruct} model with the same \textsc{nonsense}-\textit{code}-trained model used in Sec.~\ref{generalization}, without training on this evaluation distribution.

\begin{table}[H]
\centering
\caption{\textbf{Re-Imagine-derived benchmark (Extremely Informal).}
Pass@1 performance on the unstructured, partially specified GSM8K evaluation set. We compare the base Qwen2.5-3B-Instruct model with a version fine-tuned on 135 \textsc{nonsense}-\textit{code} instances per SCM (\textsc{Train-135}). Results are reported for average causal queries, counterfactual queries, and intervention queries.}
\label{tab:causal_queries}
\begin{tabular}{lccc}
\toprule
Model & Avg. Causal & Ctf. & Int. \\
\midrule
Base & 0.1380 & 0.1218 & 0.1527 \\
Train-135 & \textbf{0.1940} & \textbf{0.1891} & \textbf{0.1985} \\
\bottomrule
\label{tab:extremely_informal}
\end{tabular}
\end{table}

Average Pass@1 increases from 13.80\% to 19.40\%; counterfactual performance increases from 12.18\% to 18.91\%, and intervention performance from 15.27\% to 19.85\%. These gains support transfer to partially specified, non-executable inputs while leaving substantial headroom in absolute accuracy.

\subsection{\texorpdfstring{Pass@$k$}{Pass@k} and Majority-Vote Evaluation}
\label{app:passk_majority}

Table~\ref{tab:real_causal_graphs_passk} reports the additional multi-sample metrics for the external benchmarks. Corresponding NIHSS Pass@$k$ results appear in Appendix~\ref{subsec:exps_p3}.

\begin{table}[H]
\centering
\caption{\textbf{Pass@k and majority-vote performance on external benchmarks.}
We compare the base Qwen2.5-3B-Instruct model with the model fine-tuned on 135 \textsc{nonsense} queries per SCM.}
\label{tab:real_causal_graphs_passk}
\begin{tabular}{lcccc}
\toprule
& \multicolumn{2}{c}{Pass@k} & \multicolumn{2}{c}{Majority vote} \\
\cmidrule(lr){2-3} \cmidrule(lr){4-5}
Dataset & Base & Train-135 & Base & Train-135 \\
\midrule
ASIA & 0.220 & \textbf{0.235} & 0.110 & \textbf{0.165} \\
CANCER & 0.300 & \textbf{0.315} & 0.100 & \textbf{0.130} \\
SURVEY & 0.135 & \textbf{0.145} & 0.025 & \textbf{0.055} \\
ReImagine-derived & 0.595 & \textbf{0.623} & 0.461 & \textbf{0.469} \\
\bottomrule
\end{tabular}
\end{table}

\subsection{Failure Modes of One-Shot SCM Generation}
\label{app:one-shot-failure-modes}

As a preamble to our incremental setup, we characterize the dominant failure modes of one-shot SCM generation. Table~\ref{tab:scm-errors} summarizes failures when prompting each model to generate a connected 20-node SCM in one prompt. Both models exhibit substantial failure rates, but the dominant failure mode differs: Qwen frequently produces disconnected graphs, whereas GPT-5 more often violates the requested node count.

\begin{table}[H]
\centering
\caption{\textbf{Dominant error modes for one-shot generation of connected 20-node SCMs.}
Results are measured over 100 trials. One-shot SCM generation becomes brittle at this scale.}
\label{tab:scm-errors}
\small
\begin{tabular}{lcc}
\hline
\textbf{Error type} & \textbf{Qwen} & \textbf{GPT-5} \\
\hline
Graph is disconnected & 75 / 100 & 3 / 100 \\
Wrong number of nodes & 3 / 100 & 44 / 100 \\
Invalid syntax & 14 / 100 & 3 / 100 \\
Exogenous confounding & 3 / 100 & -- \\
Correct & 5 / 100 & 50 / 100 \\
\hline
\end{tabular}
\end{table}

\clearpage

\section{Why Do We Want Causal Simulators?}
\label{appendix:causal_sim}

Table~\ref{tab:simulator_eg} gives examples of domains where a causal simulator can support meaningful causal queries. The table illustrates the same core pattern across domains: a formal simulator encodes mechanisms, an informal representation describes those mechanisms in the format users typically encounter, SCM complexity can be scaled, and causal queries can be answered mechanically once the simulator is executable.

\begin{table}[H]
\centering
\caption{\textbf{Formal and informal representations of SCM-based environments across domains.}
The examples illustrate how simulator complexity and causal-query difficulty can scale together.}
\label{tab:simulator_eg}
\scriptsize
\setlength{\tabcolsep}{3.5pt}
\renewcommand{\arraystretch}{1.05}
\begin{tabularx}{\textwidth}{@{}p{.9cm}p{3.25cm}p{2.55cm}p{3.05cm}X@{}}
\toprule
\textbf{Setting} & \textbf{Formal representation} & \textbf{Informal representation} & \textbf{Scaling SCM complexity} & \textbf{Example causal query} \\
\midrule
Physics & Rigid-body simulator; $U$: initial conditions & Text, frames, or diagram & Add objects, forces, and longer horizons & If we set the initial speed higher, where does the object land? \\
Robot & 3D manipulation simulator; $U$: friction and mass & RGB-D trace plus instruction & Add tools, contacts, and steps & Given a slip and failure, would the task succeed with a tighter grip? \\
Drug response & PK/PD code; $U$: patient type & Clinical note plus lab plot & Add compartments, interactions, and dosing rules & If we set the dose to high, what outcome follows? \\
Circuits & SPICE/HDL simulator; $U$: tolerances & Schematic plus waveforms & Add stages, feedback, and nonidealities & If we replace $R_2$ with $2$k$\Omega$, how does the output voltage change? \\
Math & Executable algebraic program; $U$: parameters & Equations or word problem & Add variables and compositions & If we force $x=x'$, what value(s) can $y$ take? \\
Logic & Rule system or finite-state simulator; $U$: events & Natural-language rules or puzzles & Add rules, relations, and states & The alarm went off. Would it still go off if the sensor were off? \\
\bottomrule
\end{tabularx}
\end{table}

\clearpage

\section{LLM-Based SCM Generation}
\label{app:llm_vs_procedural}
SCMs can also be generated without LLMs by sampling graphs and mechanisms from pre-defined priors. Here, we discuss why we explore LLM-based SCM generation and provide an initial empirical comparison against procedural SCM generators.

\textbf{Why LLM-based SCM generation?}
We view LLM-based SCM generation as a valuable alternative to fixed-prior procedural methods for two main reasons: (1) reduced dependence on manually engineered priors, and (2) the ability to convert non-executable causal knowledge into executable SCMs.

\textbf{Reduced dependence on manually engineered priors.}
To construct SCMs, a procedural generator samples from a manually specified prior over variables, edge probabilities, mechanisms, and exogenous noise distributions. An LLM also induces a prior, but one learned through broad pretraining and then conditioned on through prompts. It can therefore draw on pretrained knowledge to generate domain-specific SCMs, such as the mathematical and medical SCMs in our experiments, without requiring every component of the sampling distribution to be redesigned manually.

This reduced dependence on prior-distribution engineering connects to the motivation of recent work such as Absolute Zero~\cite{zhao2025absolutezero}: as model capabilities improve, requiring humans to construct and enumerate the training distribution may itself become a limiting factor. This makes model-generated environments that are mechanically verifiable a valuable alternative to explore.

\textbf{Conversion of non-executable causal knowledge into executable SCMs.}
A procedural generator can sample executable SCMs once the prior is defined. However, it does not by itself address the representation challenge identified in our introduction: causal knowledge is often available only as partial and non-executable descriptions. LLM-based construction allows this knowledge to be converted into executable SCMs and incorporated incrementally. This means that, as additional causal knowledge becomes available, new variables and mechanisms can be added step by step. A procedural generator could represent the same knowledge, but only after a human had incorporated it into the prior distribution.

Our current results therefore establish LLM-based generation with incremental construction and mechanical verification as one viable approach to SCM construction. A rigorous comparison against fixed-prior methods requires selecting and validating priors over variables, graph structures, mechanisms, and noise distributions, and conclusions may depend strongly on these choices and on the domain being evaluated. We therefore view a systematic comparison across different prior families as an important direction for future work.

\subsection{Procedurally Generated SCM Baselines}
\label{appendix}

To provide an initial empirical comparison against non-LLM data generation, we construct training datasets using procedural SCM generators and compare models trained on these datasets against models trained on our LLM-generated SCMs.

\paragraph{Data generation.}
A non-LLM SCM generator requires explicit specification of the prior from which each component of the SCM is sampled, including the graph structure, exogenous noise distributions, and structural mechanisms. We use the same connected random-DAG structure across both non-LLM settings, constructed incrementally by adding one variable at each step. Each exogenous variable is sampled independently from a Bernoulli distribution.

We evaluate two manually specified mechanism priors over \textsc{Nonsense} variable names. Root variables copy their exogenous value, while each non-root variable applies either OR or XOR to its exogenous value and the values of all its parents. We generate a separate training dataset for each mechanism family, yielding \textbf{No-LLM: OR} and \textbf{No-LLM: XOR} baselines, and compare them against training on our LLM-generated \textsc{Nonsense}-code SCMs.

\paragraph{Training setup.}
For each setting, we train Qwen2.5-3B-Instruct using the same number of SCMs and the same training setup.

\paragraph{Evaluation setup.}
We evaluate on held-out external datasets: GSM8K and the bnlearn Bayesian networks ASIA, CANCER, and SURVEY. GSM8K contains 270 independently constructed SCMs, with 270 intervention and 264 counterfactual queries. ASIA, CANCER, and SURVEY each correspond to a single SCM and contain 200 queries: 100 intervention and 100 counterfactual queries. We report Pass@1 on counterfactual queries.

\begin{table}[t]
\centering
\small
\begin{tabular}{lcccc}
\toprule
\textbf{Evaluation Dataset} &
\textbf{Base LLM} &
\textbf{LLM} &
\textbf{No-LLM: OR} &
\textbf{No-LLM: XOR} \\
\midrule
GSM8K  & 25.38 & \textbf{32.95} & 30.68 & 31.06 \\
ASIA   & 6.00  & \textbf{23.00} & 4.00  & 5.00  \\
CANCER & 8.00  & \textbf{11.00} & 5.00  & 10.00 \\
SURVEY & 4.00  & \textbf{6.00}  & 5.00  & 2.00  \\
\midrule
\textbf{Avg. over Datasets} &
10.84 & \textbf{18.24} & 11.17 & 12.02 \\
\bottomrule
\end{tabular}
\caption{Counterfactual Pass@1 (\%) when training on LLM-generated SCMs versus procedurally generated SCMs.}
\label{tab:procedural_scm_baselines}
\end{table}

\paragraph{Results.}
Training on LLM-generated SCMs achieves the best performance on all four evaluation datasets and the highest average performance overall. Both non-LLM baselines improve over the base model on average, while performance also varies across the manually specified mechanism priors.

\paragraph{Discussion.}
These results support our motivation of reducing dependence on manually engineered priors. Procedurally generated SCMs provide useful training signal, but their performance varies with the specified priors over graph structure, exogenous noise, and mechanisms, as well as with how well these choices align with the evaluation distribution. In contrast, the LLM provides a broader learned prior without requiring each component of the SCM-generating distribution to be specified manually.

These experiments provide an initial empirical comparison rather than a comprehensive evaluation of LLM-based versus procedural SCM generation. A more systematic comparison across procedural priors, including different graph families, mechanism families, exogenous distributions, and domains, remains important future work.

\clearpage
\section{Discrete SCMs}
\label{app:discrete_scms}
We use discrete exogenous variables in our framework. This design choice serves two purposes: (1) discrete supports still require correct counterfactual reasoning, and (2) the discrete setting isolates causal reasoning from the additional mathematical demands of continuous probabilistic inference.

\paragraph{Discrete Supports Still Require Correct Counterfactual Reasoning.}
For any counterfactual query, the verifier first identifies the exogenous assignments that are compatible with the observed variables, applies the intervention under each of those assignments, and propagates its effects through the structural equations. The resulting set of possible outcomes forms the support of valid counterfactual answers, and the model is rewarded when its prediction matches any value in this support.

Importantly, the use of discrete variables does not make the counterfactual task trivial: producing a valid answer still requires the three steps of counterfactual reasoning:
\begin{enumerate}
\item \textbf{Abduction.} Reasoning from the observed variables to the exogenous assignments that could have generated them;
\item \textbf{Action.} Modifying the relevant structural equation through $\operatorname{do}(\cdot)$; and
\item \textbf{Prediction.} Propagating the exogenous assignments forward through the intervened SCM to obtain the corresponding counterfactual outcome.
\end{enumerate}

Thus, in our discrete exogenous-noise setting, an LLM must correctly perform counterfactual reasoning to receive a reward. This aligns the training objective with our goal of improving causal reasoning.

\paragraph{The Discrete Setting Isolates Causal from Mathematical Reasoning.}
In a continuous setting, the same causal operations require additional probabilistic inference:
\begin{enumerate}
\item Abduction requires computing or approximating a posterior density over continuous exogenous variables.
\item Prediction requires propagating this density through the modified SCM and performing integration or density estimation.
\item Verification must account for the fact that exact equality to a single value is generally not an appropriate correctness criterion.
\end{enumerate}

By using finite supports and exact executable verification, our framework allows us to test whether a model has learned the do-operator and the abduction--action--prediction procedure without conflating this with its ability to perform continuous probabilistic inference.

\paragraph{Future Work.}
Extending the framework to continuous-valued exogenous noise and stochastic mechanisms is an important direction for future work. We view this extension as distinct from the primary objective of the present work: designing a verifiable training environment for causal reasoning.

\section{Related Work Comparison}
\label{app:appendix-related-work}

Table~\ref{tab:scm_comparison} compares our approach with prior SCM-based causal-reasoning work. The comparison highlights that our work focuses on incrementally constructing scalable executable SCMs and using them to improve causal reasoning through generated supervision.

\newcommand{\vcenthdr}[1]{\raisebox{0.5\normalbaselineskip}{#1}}

\begin{table}[H]
\centering
\caption{\textbf{Comparison of SCM-based approaches for causal reasoning.}}
\label{tab:scm_comparison}
\small
\scalebox{0.75}{%
\begin{tabular}{c c l c l l l}
\toprule
\vcenthdr{Approach}
& \vcenthdr{SCM type}
& \shortstack[l]{How SCM rules\\are given}
& \shortstack[c]{Improves\\reasoning}
& \shortstack[l]{Abduction for\\counterfactuals}
& \shortstack[l]{Scalable LLM-based\\benchmark}
& \shortstack[l]{How SCM\\is created}
\\
\midrule
Ours & Logic & Description & Yes & LLM simulates $p(u\mid\cdot)$ & Yes, including complexity scaling & Step-by-step \\
\cite{xu2025re} & Mathematical & Description & No & No & Yes, including mutation & One-shot \\
\cite{huyuk2024reasoning} & Logic & Description & Yes & No & No; static benchmark & Human-created \\
\cite{maasch2025causalarc} & Spatial & Examples & No & No & No; static benchmark & Human-created \\
\cite{jin2023cladder} & Mathematical & Description & No & LLM derives $p(u\mid\cdot)$ & No; static benchmark & Human-created \\
\bottomrule
\end{tabular}%
}
\end{table}

\section{Causal Pattern Recognition versus Causal Mechanism Learning}
\label{app:pattern-vs-mechanism}

A useful way to interpret the scope of our results is to distinguish causal-query answering from learning a causal model. In the structural causal model (SCM) view, causal knowledge is not only a joint distribution or a predictor over observed variables, but a collection of autonomous structural assignments \(V_i := f_i(\mathrm{pa}_i,U_i)\), together with assumptions that license interventions and counterfactuals \citep{pearl1995causal,pearl2009causality,galles1998axiomatic,shpitser2008complete,peters2017elements}. Observational, interventional, and counterfactual queries occupy distinct levels of the causal hierarchy: two models may agree on observational associations while disagreeing under interventions, and may further disagree on counterfactual queries even when some interventional distributions coincide \citep{pearl2009causality,shpitser2008complete,bareinboim_2016}. Therefore, high supervised accuracy on a distribution of causal queries does not by itself imply that a neural model has identified the underlying causal graph, recovered the true modular mechanisms, or represented the data-generating process internally. The learner may instead exploit templates, lexical regularities, or other predictive shortcuts that are sufficient for the training and evaluation distribution but brittle under mechanism-level shifts \citep{scholkopf2012causal,geirhos2020shortcut,scholkopf2021toward}.

Our simulator-based supervision should therefore be viewed as a controlled way to train and evaluate causal-query answering, rather than as proof that the resulting model has learned invariant causal mechanisms. A key advantage of our framework is that labels are generated by explicit executable SCMs and by mechanistic evaluation of observational, interventional, and counterfactual queries, rather than inferred from surface correlations in observational data. In addition, our use of synthetic nonsensical variables, progressively constructed mechanisms, and query curricula is intended to reduce reliance on common-knowledge priors and make the relevant causal computation more explicit. Nevertheless, the fine-tuning objective remains an input-output predictive loss, and standard work on causal generalization emphasizes that robust transfer typically requires invariances across environments and representations aligned with stable causal variables or mechanisms \citep{peters2016causal,rojas2018invariant,arjovsky2019invariant,locatello2019challenging,scholkopf2021toward}. We therefore interpret our out-of-domain improvements as evidence that simulator-generated supervision can induce useful causal-reasoning behavior, but not as certification that the model's internal representation is itself an SCM.

Future work should more directly test whether the learned representations are mechanistic rather than merely predictive. One direction is to evaluate on held-out families of graphs, structural equations, noise models, and interventions that preserve superficial query form while changing the underlying causal semantics. Another is to train across multiple environments with explicit invariance, modularity, or mechanism-factorization constraints \citep{peters2016causal,rojas2018invariant,arjovsky2019invariant,scholkopf2021toward}. Complementary analyses could inspect whether internal states align with SCM variables, use interchange interventions or causal-abstraction tests to verify whether those states play the expected causal role, and apply mediation-style analyses to determine whether putative mechanism representations are actually used in producing answers \citep{vig2020investigating,geiger2021causal}. Such evaluations would help distinguish models that recognize causal-looking patterns from models that encode transportable causal mechanisms, and would make the proposed simulator framework a testbed for causal representation learning in addition to causal-query supervision.

\clearpage

\section{Limitations of Self-Improvement}
\label{app:self-improvement-limitations}

Our self-improvement experiment should be interpreted as a one-step bootstrapping result, not as evidence of unbounded recursive improvement. In our setting, the model is used to generate candidate SCMs, deterministic verification filters candidates for structural validity and executability, and the accepted SCMs are then executed to produce causal-query labels. This differs from unconstrained pseudo-labeling because the training labels are not the model's own free-form answers; they are mechanically computed from accepted simulators. Thus, for any accepted SCM, supervision is exact under the simulator semantics. This connects our method to prior self-training and reasoning-bootstrap approaches, where models improve by generating, filtering, and training on their own outputs \citep{xie2020selftraining,zelikman2022star,huang2023large}, but with the important distinction that our verifier checks the generated environment rather than only the final answer.

Nevertheless, there is an inherent ceiling: the generator determines the distribution of SCMs from which supervision is obtained. Let $G_\theta$ denote the model-induced SCM generator and let $\mathcal{V}(M)$ be the verification predicate. Training data are drawn from the accepted distribution $p_{G_\theta}(M \mid \mathcal{V}(M)=1)$, so the learner can only receive supervision for mechanisms, topologies, domains, and query structures that appear in this accepted support. Verification reduces syntactic and execution errors, but it does not by itself certify semantic richness, domain realism, mechanistic diversity, or coverage of rare causal structures. A weak generator may therefore produce valid but repetitive, shallow, or biased SCMs, creating a form of curriculum bottleneck. This is analogous to known limitations of self-training with noisy or biased pseudo-labels, where errors or biases can be reinforced over iterations \citep{arazo2020pseudo,chen2022debiased}. More broadly, work on synthetic-data feedback loops shows that repeatedly training on generated data can degrade quality or diversity when generated data replace or dominate fresh data \citep{shumailov2024aimodels,alemohammad2024selfconsuming}; conversely, accumulating generated data together with real or externally grounded data can mitigate collapse \citep{gerstgrasser2024modelcollapse}.

Within the range studied in Sec.~\ref{sec:experiments}, we do not observe this saturation: performance continues to improve as additional simulator-generated supervision is introduced. The location of any eventual plateau is therefore an empirical property of the training setting and may depend on model capacity, the diversity and quality of the generated SCMs, and the alignment between the generated and evaluation distributions.

For this reason, our results establish that self-generated SCMs can be useful within the tested distribution and under deterministic verification, but they do not characterize the long-run fixed point of recursive self-improvement. Extending bootstrapping would require measuring the generator frontier across rounds: acceptance rate, number of unique mechanisms, graph-motif diversity, support sizes, query difficulty, and held-out performance on SCM families not generated by the current model. It would also require safeguards that expand rather than recycle the training distribution, such as mixing self-generated SCMs with externally generated or domain-grounded simulators, using independent validators, adding explicit diversity objectives, and periodically evaluating against out-of-generator benchmarks. Under such a view, self-improvement saturates when the accepted SCM distribution stops expanding or when additional accepted SCMs no longer provide new causal computations for the learner.

\clearpage

\section{Incremental Construction of Executable Structural Causal Models}
\label{appendix:algorithm}

This appendix provides implementation details for the incremental construction procedure used to generate executable structural causal models (SCMs). To reduce notational clutter, we identify each endogenous node with its mechanism name \(f_i\), whose output is \(v_i\).

\subsection{SCM Specification}
\label{subsec:scm_specification}

The user specifies the target number of endogenous variables \(n\), a domain \(\mathcal{D}\) such as logic-based symbolic mechanisms, and optionally a DAG topology \(G\), such as a chain, tree, or star. These inputs are compiled into a prompt that asks the LLM to produce an initial executable SCM: exogenous samplers \(\{U_i\}_{i=1}^n\), structural mechanisms \(\{f_i\}_{i=1}^n\), and a driver function \texttt{run\_once()}.

Each \(f_i\) takes its exogenous parent \(U_i\) and any endogenous parents explicitly as arguments, and returns a deterministic symbolic value. The resulting code is parsed and checked against the topology and SCM structural constraints described in Section~\ref{subsec:scm_verification}.

\subsection{Planning}
\label{subsec:scm_planning}

To grow an SCM by one node, we first plan a \emph{symbolic} graph edit. The LLM is given a compact \emph{semantic view} of the current SCM: the exogenous samplers \(\mathcal{U}=\{U_1,\dots,U_n\}\), the structural mechanisms \(\mathcal{V}=\{f_1,\dots,f_n\}\), the directed edge set \(E \subseteq \mathcal{V}\times \mathcal{V}\), and a valid topological order \(\pi\) over \(\mathcal{V}\). To convey meaning without exposing full low-level code, we also include short natural-language summaries, implemented as docstrings, for each \(U_i\) and \(f_i\).

The planner outputs a modification decision specifying only:
(i) the name of a new mechanism \(f_{n+1}\),
(ii) the name of its exogenous sampler \(U_{n+1}\),
(iii) its direct parents \(Pa(f_{n+1}) \subseteq \mathcal{V}\), and
(iv) its direct children \(Ch(f_{n+1}) \subseteq \mathcal{V}\).
No functional code is generated at this stage. The planner is instructed to maintain acyclicity, preserve standard SCM constraints, and obey the user-specified topology \(G\). This symbolic plan is then passed to a second LLM call that performs the code update.

\subsection{Execution}
\label{subsec:scm_execution}

Given a planned edit, the executor implements it as a \emph{localized patch}. To limit context, the LLM sees only:
(i) full Python definitions for each direct child \(f_j\in Ch(f_{n+1})\), since these functions must be updated, and
(ii) docstrings for each direct parent \(f_i\in Pa(f_{n+1})\), which provide semantic compatibility without exposing unnecessary code.
No other SCM code is exposed.

The executor then:
(i) defines the new exogenous sampler \(U_{n+1}\),
(ii) implements \(f_{n+1}\) as a deterministic function of \(Pa(f_{n+1})\cup\{U_{n+1}\}\), and
(iii) updates each child \(f_j\in Ch(f_{n+1})\) to take the output of \(f_{n+1}\) as an additional input, adjusting signatures and bodies as required by the topology.

\subsection{Verification}
\label{subsec:scm_verification}

After each edit, we apply deterministic post-generation checks to enforce the constraints the LLM was instructed to satisfy. We verify that the induced causal graph \(G=(\mathcal{V},E)\) is acyclic and connected, that each mechanism has exactly one corresponding exogenous parent, and that the number of endogenous variables matches the expected count. We additionally check compliance with the user-specified topology and require that the SCM executes without runtime errors in Python.

If any check fails, the edit is rejected and retried up to a user-specified limit. This verification loop ensures that the SCM remains a valid, executable causal simulator throughout construction.

\paragraph{Summary.}
Let \(\mathcal{M}_0\) be an initial SCM with \(|\mathcal{V}_0|=n_0\) endogenous mechanisms. At iteration \(t\), we propose and verify an edit that yields an updated SCM \(\mathcal{M}_t\) with
\[
\mathcal{V}_t = \mathcal{V}_{t-1}\cup\{f_{n_0+t}\}.
\]
Each update is accepted only if it passes all verification checks, producing a sequence \(\{\mathcal{M}_t\}_{t=0}^{T}\) of progressively larger causal simulators. Algorithm~\ref{alg:incremental_scm} summarizes the procedure.

\begin{algorithm}[H]
\caption{Incremental Construction of Executable Structural Causal Models}
\label{alg:incremental_scm}
\begin{algorithmic}[1]
\Require Initial number of endogenous variables \(N_0\), target number \(N\), domain \(\mathcal{D}\), optional DAG topology \(G\), initialization retry limit \(K_0\), update retry limit \(K\)
\Ensure Executable SCM \(\mathcal{M}\) with \(N\) endogenous variables

\Statex \textbf{Stage 0: SCM specification}
\For{\(r = 1\) to \(K_0\)}
    \State Prompt LLM to generate initial SCM Python code: exogenous samplers \(\mathcal{U}\), structural mechanisms \(\mathcal{F}\), and driver function \texttt{run\_once()}
    \State Verify acyclicity, connectedness, one exogenous parent per \(f_i\), correct variable count, topology compliance, and Python execution
    \If{verification passes}
        \State Initialize \(\mathcal{M}\), \(\mathcal{U}\), \(\mathcal{V}\), \(E\), and \(n \gets N_0\)
        \State \textbf{break}
    \EndIf
\EndFor
\If{initialization failed}
    \State Abort and report failure
\EndIf

\Statex \textbf{Stages 1--3: Incremental SCM updating}
\While{\(n < N\)}
    \State Construct semantic view of current SCM: \(\mathcal{U}\), \(\mathcal{V}\), edge set \(E\), topological order \(\pi\), and docstrings
    \State Prompt planner LLM to propose a symbolic edit: new mechanism \(f_{n+1}\), new exogenous sampler \(U_{n+1}\), parents \(Pa(f_{n+1})\), and children \(Ch(f_{n+1})\)
    \State \(accepted \gets \textsc{False}\)
    \For{\(k = 1\) to \(K\)}
        \State Provide executor LLM with planner decision, full code of \(f_j \in Ch(f_{n+1})\), and docstrings for \(f_i \in Pa(f_{n+1})\)
        \State Executor defines \(U_{n+1}\), implements \(f_{n+1}\), updates children \(Ch(f_{n+1})\), and updates \texttt{run\_once()}
        \State Verify the updated SCM
        \If{verification passes}
            \State Update \(\mathcal{U} \gets \mathcal{U} \cup \{U_{n+1}\}\), \(\mathcal{V} \gets \mathcal{V} \cup \{f_{n+1}\}\), and \(n \gets n+1\)
            \State \(accepted \gets \textsc{True}\)
            \State \textbf{break}
        \EndIf
    \EndFor
    \If{\(accepted = \textsc{False}\)}
        \State Abort and report failure to extend SCM
    \EndIf
\EndWhile
\State \Return Executable SCM \(\mathcal{M} = (\mathcal{U}, \mathcal{V}, E, \texttt{run\_once})\)
\end{algorithmic}
\end{algorithm}

\clearpage

\section{Generation Prompts}
\label{appendix:prompts}

This appendix lists the prompts used in the incremental SCM construction procedure described in Appendix~\ref{appendix:algorithm}.

\begin{tcolorbox}[
    breakable,
    title=Logic-Based Medical SCM Instantiation Prompt,
    colback=gray!5,
    colframe=gray!100,
    sharp corners
]
\lstset{
    basicstyle=\ttfamily\small,
    breaklines=true,
    breakatwhitespace=true,
    showstringspaces=false
}
\begin{lstlisting}
SYSTEM_TEXT = You are a careful Python structural causal model (SCM) code generator.

USER_TEXT = You are generating a SMALL, EXECUTABLE, STOCHASTIC Structural Causal Model (SCM) in Python.

This SCM must be logic-based, symbolic, or mechanistic, based on a clinical, biological, or medical domain. DO NOT make the SCM mathematical.

Functions may return booleans, categorical strings, or symbolic states. They may use conditionals, for loops, and other logic statements.

HARD REQUIREMENTS:
- Allowed imports: random, typing. No others.
- Use docstrings ONLY (no '#' comments).
  Each docstring must include:
    - "Purpose:" Describe the logical or mechanistic rule and the variable's parents.
    - "Output:" Specify the exact symbolic/boolean values that the function may return.
- Define exogenous samplers named `U_*` (pure, no I/O). To support abduction:
  - Each `U_*` must be independent of all other exogenous variables.
  - There must be no exogenous confounding between variables.
- Define structural assignments named `f_*` (pure, no I/O):
  - Each `f_*` must take all parent variables, exogenous and structural, explicitly as arguments.
  - Each `f_*` must have exactly one exogenous parent.
  - The SCM must be acyclic.
  - Structural functions are deterministic given their inputs.
- Include a driver function:
    def run_once(seed: int | None = None) -> dict[str, object]:
  - If seed is not None, call random.seed(seed) inside the driver.
  - The driver must return a dict of output variables from f_* ONLY, not internal logic variables or outputs of U_* functions.

OUTPUT FORMAT, exact order:
1) All U_* exogenous sampler definitions.
2) All f_* structural mechanism definitions.
3) The run_once(seed) driver.

Return ONLY a single Python code block. No prose.

Task:
- Create an SCM with {num_nodes} exogenous samplers and {num_nodes} structural functions.
- The internal nodes (f_*) of the SCM must follow a {dag_type} topology. {DAG_DEFINITIONS[dag_type]}
\end{lstlisting}
\end{tcolorbox}

\begin{tcolorbox}[
    breakable,
    title=Topology Definitions Used in Prompts,
    colback=gray!5,
    colframe=gray!100,
    sharp corners
]
\lstset{
    basicstyle=\ttfamily\small,
    breaklines=true,
    breakatwhitespace=true,
    showstringspaces=false
}
\begin{lstlisting}
DAG_DEFINITIONS = {
    "chain": """
A chain topology is a single linear sequence of internal nodes.

Rules:
- Exactly one parent per node, except the first.
- Exactly one child per node, except the last.
- Shape: A -> B -> C -> ...
- If inserting a new node between X and Y:
    - Y must remove X as a parent.
    - Y must add the new node as its only parent.
    - The new node must take X as its parent.
    - The driver and signatures must reflect this reassignment.
- A new node may also be inserted at the start or end of the chain.
""",

    "star": """
A star topology has a single central parent, or hub, whose output flows to many children.

Rules:
- One hub node with many children.
- All children have exactly one parent, the hub.
- Children do not depend on one another.
- Shape: Hub -> {Child1, Child2, Child3, ...}
- New nodes can only be inserted as direct children of the hub.
""",

    "inverted_star": """
An inverted star topology is the reverse of a star: many parents converge into a single child, or collider.

Rules:
- A single sink node has multiple parents.
- All parents have exactly one child, the sink.
- The sink has no children.
- Parents do not depend on one another.
- Shape: {P1, P2, P3, ...} -> Child
- New nodes can only be inserted as direct parents of the sink node, with no direct parents themselves.
""",

    "tree": """
A tree topology has a single root and a branching structure with no merging.

Rules:
- One root node with no parents.
- Every other node has exactly one parent.
- Nodes may have multiple children.
- Shape: A -> {B, C}; B -> {D, E}
- A node can be inserted as the root, as a leaf, or between existing nodes.
    - If inserting a new node as the root R:
        - R must have no parents.
        - The previous root P must now have R as a parent.
    - If inserting a new node as a leaf L:
        - L must have no children.
    - If inserting a new node between parent P and child C:
        - C must remove P as its parent.
        - C must add the new node as its only parent.
        - The new node must take P as its parent.
""",

    "layered": """
A layered topology assigns every node to exactly one layer.

Rules:
- Multiple nodes can exist in one layer, but no edge can exist between nodes in the same layer.
- Allowed edges go only from layer k to layer k+1.
- No layer skipping, such as k -> k+2.
- No backward edges, such as k+1 -> k.

Inserting a new node:
- It may be placed in an existing layer or in a new layer.
- If inserted into an existing layer k:
    - The new node MUST NOT be a parent or child of any node in that layer.
    - Its parents must lie in layer k-1, if a previous layer exists.
    - Its children must lie in layer k+1, if a subsequent layer exists.
- If inserted into a new layer k:
    - The new node's direct children must lie in layer k+1.
    - The direct children must remove parents that are now two layers above.
    - Direct parents of the new node must lie in layer k-1.
""",

    "poly_tree": """
A polytree permits branching and merging while preserving acyclicity.

Rules:
- Multiple parents are allowed per node.
- Both branching, one parent to many children, and merging, many parents to one child, are allowed.
- When inserting a new node:
    - Children of the new node, if any, must update signatures and bodies accordingly.
    - The generator must preserve acyclicity.
"""
}
\end{lstlisting}
\end{tcolorbox}

\begin{tcolorbox}[
    breakable,
    title=Planner Prompt: Deciding the Next Incremental Growth Step,
    colback=gray!5,
    colframe=gray!100,
    sharp corners
]
\lstset{
    basicstyle=\ttfamily\small,
    breaklines=true,
    breakatwhitespace=true,
    showstringspaces=false
}
\begin{lstlisting}
You are planning the NEXT INCREMENTAL GROWTH of an existing logic-based, non-mathematical SCM, defined by the following functions:
- exogenous samplers (U_*) (external nodes)
- structural assignments (f_*) (internal nodes)

The ONLY permitted modification is adding a new internal node, which consists of:
- a new structural assignment f_*
- a new exogenous noise sampler U_* for the structural assignment

General constraints:
- The new node may be inserted anywhere in the graph as long as:
    - the graph remains acyclic.
    - all functions remain pure.
    - all modifications are compatible with topological execution.
    - no cycles are introduced.

Return STRICT JSON, with no prose, using this schema:
{
  "rationale": "one-sentence design intent",
  "new_function": "name of the new f_* function to create",
  "new_exo_noise": "name of the new U_* function to create",
  "direct_parents": ["f_*", ...],
  "direct_children": ["f_*", ...]
}

Important:
- direct_parents must include only f_* functions, never U_* functions.
- direct_children must include only f_* functions.
- The internal nodes (f_*) of the updated SCM must obey the strict {dag_type} topology.

SEMANTIC VIEW:
{semantic_view_json}

Return ONLY the strict JSON object above.
\end{lstlisting}
\end{tcolorbox}

\begin{tcolorbox}[
    breakable,
    title=Executor Prompt: Implementing the Localized Update,
    colback=gray!5,
    colframe=gray!100,
    sharp corners
]
\lstset{
    basicstyle=\ttfamily\small,
    breaklines=true,
    breakatwhitespace=true,
    showstringspaces=false
}
\begin{lstlisting}
You will IMPLEMENT a small update to a logic-based, non-mathematical structural causal model (SCM) in which ONE new internal structural function is added.

You are given:
1. A DECISION JSON with:
   - rationale
   - new_function: name of new f_*
   - new_exo_noise: name of corresponding U_*
   - direct_parents: list of internal parent nodes of new_function
   - direct_children: list of child nodes of new_function

2. CODE CONTEXT:
   - For each direct_child: full function, including signature, docstring, and body
   - For each direct_parent: docstring only, no code

3. DRIVER FUNCTION:
    def run_once(seed: int | None) -> dict[str, object]:

Your task:

(1) Define a new exogenous sampler U_new().
    - It must be pure and have no I/O.
    - It must return symbolic, categorical, or boolean states, not numeric math.
    - It must have a single docstring containing:
         Purpose: describe symbolic/mechanistic meaning
         Output: specify all symbolic values returned
    - It must be independent of all other U_* variables.

(2) Define the new structural function new_function:
    - Signature: one argument per direct_parent and the new_exo_noise.
    - Deterministic, logic-based, symbolic mechanism.
    - Single docstring containing Purpose and Output.
    - Must use ALL input arguments.

(3) Modify each direct_child:
    - Signature: add the new_function output as a new input in the signature and potentially remove other inputs depending on the specified topology.
    - Update docstring with Purpose and Output.
    - Update the body according to the newly specified inputs.
    - Preserve purity.

(4) Update the driver:
    - Insert call to U_new(), assigning it to a variable.
    - Insert call to new_function() with appropriate inputs, assigning output to a variable.
    - Update calls to direct_children to include the new_function output in the argument list, and potentially remove previously specified inputs depending on the topology.
    - Include outputs from new_function in the return statement.
    - Maintain topological order and purity.
    - Return ONLY outputs of structural functions, not exogenous variables.

Output format, Python code only:
    - Create the function for U_new.
    - Create the function for f_new.
    - Modify the direct_child functions to incorporate the output of f_new.
    - Modify the driver to make the new calls and amend the existing direct_child calls.

Constraints:
    - Allowed imports: random, typing.
    - Only docstrings; no '#' comments.
    - Non-mathematical: use symbolic, conditional, boolean, or categorical logic ONLY.
    - Do NOT emit unrelated functions.
\end{lstlisting}
\end{tcolorbox}

\clearpage

\section{Evaluation Prompt Examples}
\label{sec:prompt}

This section gives representative prompts used to evaluate causal-query answering. The examples show how the executable SCM, query type, factual evidence, intervention, and required output format are presented to the model.

\begin{tcolorbox}[
    breakable,
    title=Evaluation Example: Real Causal Graph (Asia SCM),
    colback=gray!5,
    colframe=gray!100,
    sharp corners,
    boxsep=1mm,
    left=1mm,
    right=1mm,
    top=1mm,
    bottom=1mm,
    fontupper=\ttfamily\tiny
]
\lstset{
    basicstyle=\ttfamily\tiny,
    breaklines=true,
    breakatwhitespace=true,
    showstringspaces=false
}
\begin{lstlisting}
def U_Cancer() -> float:
    choices = [i / 19 for i in range(20)]
    return random.choice(choices)

def U_Dyspnoea() -> float:
    choices = [i / 19 for i in range(20)]
    return random.choice(choices)

def U_Pollution() -> float:
    choices = [i / 19 for i in range(20)]
    return random.choice(choices)

def U_Smoker() -> float:
    choices = [i / 19 for i in range(20)]
    return random.choice(choices)

def U_Xray() -> float:
    choices = [i / 19 for i in range(20)]
    return random.choice(choices)

def f_Pollution(u_pollution: float) -> str:
    probs = [0.9, 0.1]; states = ['low', 'high']; cum=0.0
    for p, s in zip(probs, states):
        cum += p
        if u_pollution < cum: return s
    return states[-1]

def f_Smoker(u_smoker: float) -> str:
    probs = [0.3, 0.7]; states = ['True', 'False']; cum=0.0
    for p, s in zip(probs, states):
        cum += p
        if u_smoker < cum: return s
    return states[-1]

def f_Cancer(smoker: str, pollution: str, u_cancer: float) -> str:
    states = ['True', 'False']
    poll_index = 0 if pollution=='low' else 1
    smoke_index = 0 if smoker=='True' else 1
    p_true = [[0.03,0.001],[0.05,0.02]][poll_index][smoke_index]
    p_false=[[0.97,0.999],[0.95,0.98]][poll_index][smoke_index]
    probs = [p_true,p_false]; cum=0.0
    for p,s in zip(probs, states):
        cum+=p
        if u_cancer<cum: return s
    return states[-1]

def f_Dyspnoea(cancer: str, u_dyspnoea: float) -> str:
    states=['True','False']; probs=[0.65,0.35] if cancer=='True' else [0.3,0.7]
    cum=0.0
    for p,s in zip(probs,states):
        cum+=p
        if u_dyspnoea<cum: return s
    return states[-1]

def f_Xray(cancer: str, u_xray: float) -> str:
    states=['positive','negative']; probs=[0.9,0.1] if cancer=='True' else [0.2,0.8]
    cum=0.0
    for p,s in zip(probs,states):
        cum+=p
        if u_xray<cum: return s
    return states[-1]

def run_once(seed: int|None) -> dict[str,any]:
    if seed is not None: random.seed(seed)
    u_pollution=U_Pollution(); pollution=f_Pollution(u_pollution)
    u_smoker=U_Smoker(); smoker=f_Smoker(u_smoker)
    u_cancer=U_Cancer(); cancer=f_Cancer(smoker,pollution,u_cancer)
    u_dyspnoea=U_Dyspnoea(); dyspnoea=f_Dyspnoea(cancer,u_dyspnoea)
    u_xray=U_Xray(); xray=f_Xray(cancer,u_xray)
    return {'Pollution':pollution,'Smoker':smoker,'Cancer':cancer,'Dyspnoea':dyspnoea,'Xray':xray}
\end{lstlisting}
\end{tcolorbox}

\begin{tcolorbox}[
    breakable,
    title=Evaluation Example: Informal Medical SCM,
    colback=gray!5,
    colframe=gray!100,
    sharp corners,
    fontupper=\ttfamily\small
]
uv is set as the returned output of U\_vaccination\_history. The generation rule for U\_vaccination\_history is: generate a random number uniformly between 0 and 1; if it is less than 0.3, return "none"; else if it is less than 0.7, return "partial"; otherwise return "full".

\vspace{0.5em}

ui is set as the returned output of U\_infection\_pressure. The generation rule for U\_infection\_pressure is: generate a random number uniformly between 0 and 1; if it is less than 0.5, return "low"; else if it is less than 0.85, return "moderate"; otherwise return "high".

\vspace{0.5em}

ug is set as the returned output of U\_genetic\_immunity. The generation rule for U\_genetic\_immunity is: generate a random number uniformly between 0 and 1; if it is less than 0.25, return "weak"; else if it is less than 0.75, return "typical"; otherwise return "robust".

\vspace{0.5em}

ubias is set as the returned output of U\_infection\_outcome\_bias. The generation rule for U\_infection\_outcome\_bias is: generate a random number uniformly between 0 and 1; if it is less than 0.2, return "attenuated"; else if it is less than 0.85, return "wildtype"; otherwise return "hypervirulent".

\vspace{0.5em}

adaptive\_status is set as the returned output of f\_adaptive\_status. The input to f\_adaptive\_status is U\_vaccination\_history. The generation rule is: if U\_vaccination\_history equals "none", return "naive"; if it equals "partial", return "primed"; otherwise return "boosted".

\vspace{0.5em}

exposure is set as the returned output of f\_exposure. The input is U\_infection\_pressure. The generation rule is: if U\_infection\_pressure equals "low", return "minimal"; if it equals "moderate", return "community"; otherwise return "close\_contact".

\vspace{0.5em}

innate\_tone is set as the returned output of f\_innate\_tone. The input is U\_genetic\_immunity. The generation rule is: if U\_genetic\_immunity equals "weak", return "hypo"; if it equals "typical", return "normo"; otherwise return "hyper".

\vspace{0.5em}

infection\_outcome is set as the returned output of f\_infection\_outcome. The inputs are exposure, adaptive\_status, innate\_tone, and U\_infection\_outcome\_bias. The generation rule is: first compute a risk score as the sum of three components derived from the inputs: base risk by exposure where "minimal" maps to 0, "community" maps to 1, and "close\_contact" maps to 2; modification by adaptive\_status where "naive" adds 2, "primed" adds 1, and "boosted" adds 0; modification by innate\_tone where "hypo" adds 2, "normo" adds 1, and "hyper" adds 0. Then apply U\_infection\_outcome\_bias: if it is "attenuated", subtract 1 but do not let the risk go below 0; if it is "wildtype", make no change; if it is "hypervirulent", add 1. Finally map the resulting risk score to the outcome: 0 yields "no\_infection"; 1 or 2 yields "mild"; 3 or 4 yields "moderate"; any value 5 or greater yields "severe".
\end{tcolorbox}

\begin{tcolorbox}[
    breakable,
    title=Prompt Example: Counterfactual Query,
    colback=gray!5,
    colframe=gray!100,
    sharp corners
]
\lstset{
    basicstyle=\ttfamily\small,
    breaklines=true,
    breakatwhitespace=true,
    showstringspaces=false
}
\begin{lstlisting}
### System prompt
```
You're a helpful assistant solving problems based on user requests. The user provides a problem, possibly with instructions and a desired output format. You must first clearly outline your reasoning process enclosed within <think> reasoning process here </think>, followed immediately by the user's requested output format.
```

### User prompt
```
Given this code:

from __future__ import annotations
import random
from typing import Optional
from typing import Dict
from typing import Any

def U_ZY() -> Any:
    pool = ['xo', 'mi', 'za', 'qe', 'tu']
    pick = random.choice(pool)
    if random.random() < 0.5:
        return pick
    else:
        return pick + random.choice(['', 'x'])

def f_ZY(u_zy: str) -> Any:
    s = u_zy
    if len(s) == 0:
        t = 'a'
    elif s[0] in ['x', 'q']:
        t = s + 'v'
    else:
        t = s + 'k'
    flag = 'x' in t and (t.endswith('v') or ('k' in t and (not t.endswith('v'))))
    if flag:
        return True
    else:
        for ch in ['m', 'z']:
            if ch in t and (not t.startswith('t')):
                return True
        return False

def run_once(seed: Optional[int]) -> Dict[str, bool]:
    if seed is not None:
        random.seed(seed)
    u_zy = U_ZY()
    zy = f_ZY(u_zy)
    return {'zy': zy}

Use the Python code above as the SCM definition.
Rules:
- Exogenous variables are named U_* and are sampled once per world.
- Endogenous (inner) variables are computed deterministically by their f_* equations given U_*.
- do(X=v) forces X=v and ignores f_X for that variable only; all other equations stay the same.
- Never resample randomness. For unknown U_*, treat them as free variables that can take any value allowed by the sampler code.

Query type: counterfactual
Scenario label: deterministic
Given:
- fixed_exogenous (known values): {}
- unknown_exogenous (missing names): ["U_ZY"]
- do (intervention): {"ZY":false}
- observed_endogenous (factual evidence, pre-do): {"ZY":true}

Task:
- Counterfactual = abduction + intervention + prediction.
  1) Abduction: infer unknown_exogenous assignment(s) consistent with observed_endogenous and fixed_exogenous.
  2) Intervention: apply do(...).
  3) Prediction: using the SAME abduced exogenous values (no resampling), compute required outputs under do(...).
  Important: observed_endogenous is only evidence for step (1); do NOT enforce it after the intervention.

  Return that unique output.

Required output keys (EXACT match; no extras): ["ZY"]

Answer format (STRICT):
- Your message may optionally start with a <think>...</think> block (reasoning).
- The LAST thing in your message must be the JSON object described below (so it can be parsed reliably).
- Output exactly ONE JSON object with exactly the required keys.
```
\end{lstlisting}
\end{tcolorbox}

\begin{tcolorbox}[
    breakable,
    title=Prompt Example: Intervention Query,
    colback=gray!5,
    colframe=gray!100,
    sharp corners
]
\lstset{
    basicstyle=\ttfamily\small,
    breaklines=true,
    breakatwhitespace=true,
    showstringspaces=false
}
\begin{lstlisting}
### System prompt
```
You're a helpful assistant solving problems based on user requests. The user provides a problem, possibly with instructions and a desired output format. You must first clearly outline your reasoning process enclosed within <think> reasoning process here </think>, followed immediately by the user's requested output format.
```

### User prompt
```
Given this code:

from __future__ import annotations
import math
import random
import statistics
from typing import Dict
from typing import List
from typing import Any
from typing import Literal
from typing import Optional

def U_ZY() -> Any:
    return random.choice(['ax', 'by', 'cz'])

def U_XQ() -> str:
    return random.choice(['aa', 'bb', 'cc'])

def U_VI() -> str:
    choices = ['ga', 'ho', 'ti']
    return random.choice(choices)

def U_TU() -> str:
    return random.choice(['ra', 'so', 'ti'])

def U_QE() -> str:
    if random.choice([True, False]):
        return 'qe'
    else:
        return 'qx'

def f_ZY(u_zy: str, XQ: str, VI: str, TU: str, QE: str) -> str:
    if XQ == 'XQ':
        if VI == 'VI':
            if TU == 'TU':
                if QE == 'QE':
                    if u_zy == 'ax':
                        return 'vok'
                    elif u_zy == 'by':
                        return 'pem'
                    else:
                        return 'nul'
                else:
                    return 'nul'
            else:
                return 'nul'
        else:
            return 'nul'
    else:
        return 'nul'

def f_XQ(u_xq: str) -> str:
    if u_xq in ('aa', 'bb'):
        return 'XQ'
    return 'xq'

def f_VI(u_vi: str) -> str:
    if u_vi in ('ga', 'ho'):
        return 'VI'
    return 'VX'

def f_TU(u_tu: str) -> str:
    if u_tu in ('ra', 'so'):
        return 'TU'
    return 'TU0'

def f_QE(u_qe: str) -> str:
    if u_qe == 'qe':
        return 'QE'
    else:
        return 'QX'

def run_once(seed: int | None) -> dict[str, str]:
    if seed is not None:
        random.seed(seed)
    u_xq = U_XQ()
    XQ = f_XQ(u_xq)
    u_vi = U_VI()
    VI = f_VI(u_vi)
    u_tu = U_TU()
    TU = f_TU(u_tu)
    u_qe = U_QE()
    QE = f_QE(u_qe)
    u_zy = U_ZY()
    zy = f_ZY(u_zy, XQ, VI, TU, QE)
    return {'xq': XQ, 'vi': VI, 'tu': TU, 'qe': QE, 'zy': zy}

Use the Python code above as the SCM definition.
Rules:
- Exogenous variables are named U_* and are sampled once per world.
- Endogenous (inner) variables are computed deterministically by their f_* equations given U_*.
- do(X=v) forces X=v and ignores f_X for that variable only; all other equations stay the same.
- Never resample randomness. For unknown U_*, treat them as free variables that can take any value allowed by the sampler code.

Query type: intervention
Scenario label: deterministic
Given:
- fixed_exogenous (known values): {"U_QE":"qe","U_TU":"so","U_VI":"ga","U_XQ":"cc","U_ZY":"by"}
- unknown_exogenous (missing names): []
- do (intervention): {"TU":"TU0"}

Task:
- Intervened world (apply do(...)).
  All exogenous values are fixed => the required outputs are uniquely determined under the intervention.
  Return that unique output.

Required output keys (EXACT match; no extras): ["QE","TU","VI","XQ","ZY"]

Answer format (STRICT):
- Your message may optionally start with a <think>...</think> block (reasoning).
- The LAST thing in your message must be the JSON object described below (so it can be parsed reliably).
- Output exactly ONE JSON object with exactly the required keys.
```
\end{lstlisting}
\end{tcolorbox}

\clearpage

\section{Training Details}
\label{appendix:exp_details}

\subsection{Training and Evaluation Pipeline}
\label{app:training_pipeline}

We use a staged pipeline that converts a library of structural causal models (SCMs) into prompts, collects baseline and training signals via inference, constructs a supervised fine-tuning (SFT) dataset by filtering model outputs, fine-tunes the model either in a single pass or as a curriculum \cite{rauba2026deep}, and finally re-evaluates on a fixed held-out test set. The pipeline is executed with a unified runner that selects an experimental variant, such as structure, curriculum, or data amount, and a run mode, such as full training, test-only re-run, or re-plotting. All generated artifacts are written into a per-project workspace, while the original SCM JSON store is treated as read-only.

\paragraph{Step 1: Data preparation: SCMs to prompts.}
For each domain and structure, the pipeline:
(i) materializes SCMs into an executable form,
(ii) deterministically selects train and test SCM IDs using fixed seeds, and
(iii) builds prompt files for each task type, or prompt mode. For causal prompt modes, including deduction, abduction, intervention, and counterfactual, we first sample valid causal-query instantiations under deterministic constraints, such as intervention policy and counterfactual mode, and then format them into prompts. Train and test prompts are cached under dataset tags to ensure reproducibility and enable reuse across runs.

\paragraph{Step 2: Baseline inference before SFT.}
Before fine-tuning, the base model is evaluated on a fixed test set shared across experiments through fixed seeds and counts. This produces baseline metrics and plots. For efficiency, test prompts may be merged into a single inference job and split back into per-domain, per-structure, and per-prompt-mode result files.

\paragraph{Step 3: Training-signal collection before SFT.}
To obtain supervision, the base model is run on training prompts with multiple samples per instance. Outputs are stored as JSONL result files. If curriculum blocks are used, training results are organized by block to preserve the intended progression.

\paragraph{Step 4: SFT dataset construction by rejection sampling.}
The pipeline converts pre-SFT results into an SFT dataset by selecting one completion per prompt according to a selection strategy. The default strategy is \texttt{first\_correct}. When curriculum blocks are enabled, the final SFT dataset is assembled in block order to preserve block boundaries.

\paragraph{Step 5: Fine-tuning.}
We support two regimes:
(i) \emph{single-shot} SFT, which trains once on the full block-ordered dataset; and
(ii) \emph{sequential curriculum} SFT, which iterates block-by-block, fine-tuning after each block and carrying the checkpoint forward. Optional LoRA fine-tuning is supported. When enabled, the pipeline automatically chooses micro-batch size, gradient accumulation, and learning rate based on dataset size and model family.

\paragraph{Step 6: Post-SFT inference and evaluation.}
The final checkpoint is evaluated on the same fixed test prompts as the baseline. The pipeline computes metrics and generates paper-ready plots comparing pre-SFT and post-SFT performance, with optional grouped comparisons across experimental conditions.

\subsection{Fine-Tuning Hyperparameters}
\label{app:fine_tuning_hyperparameters}

Table~\ref{tab:sft_defaults} reports the default SFT configuration used by the pipeline.

\begin{table}[H]
\centering
\caption{\textbf{Default SFT configuration.}}
\label{tab:sft_defaults}
\small
\scalebox{0.75}{%
\begin{tabular}{ll}
\hline
Parameter & Default value \\
\hline
Epochs (\texttt{NUM\_EPOCHS}) & 2 \\
Max sequence length (\texttt{MAX\_SEQ\_LENGTH}) & 2048 \\
Optimizer (\texttt{OPTIM}) & \texttt{adamw\_torch\_fused} \\
Warmup ratio (\texttt{WARMUP\_RATIO}) & 0.03 \\
Weight decay (\texttt{WEIGHT\_DECAY}) & 0.0 \\
Max gradient norm (\texttt{MAX\_GRAD\_NORM}) & 1.0 \\
bf16 (\texttt{BF16}) & 1 \\
tf32 (\texttt{TF32}) & 1 \\
Gradient checkpointing (\texttt{GRAD\_CKPT}) & 1 \\
Dataloader workers (\texttt{DATALOADER\_NUM\_WORKERS}) & 2 \\
LoRA (\texttt{USE\_LORA}) & 0, off by default \\
Merge LoRA and save (\texttt{MERGE\_LORA\_AND\_SAVE}) & 1 \\
4-bit quantization (\texttt{SFT\_USE\_4BIT}) & 0, off by default \\
Dataloader shuffle (\texttt{NO\_SHUFFLE}) & 1 in single-shot training to preserve block order \\
\hline
\end{tabular}%
}
\end{table}

\paragraph{Automatic hyperparameter selection.}
When \texttt{AUTO\_SFT\_HPARAMS=1}, the pipeline selects micro-batch size, gradient accumulation, and learning rate based on dataset size, bucketed by total examples, and model family, split into 3B and \(\geq\)7B settings. Tables~\ref{tab:sft_auto_nolora} and~\ref{tab:sft_auto_lora} report the target effective global batch size used by the heuristic. For all experiments, we use 4A100.

\begin{table}[H]
\centering
\caption{\textbf{Automatic SFT hyperparameter heuristic without LoRA.}
\texttt{micro} is the per-device batch size, \texttt{ga} is gradient accumulation, and \texttt{global} is the target effective global batch size.}
\label{tab:sft_auto_nolora}
\small
\scalebox{0.95}{%
\begin{tabular}{lll}
\hline
Dataset bucket & 3B, no LoRA & 7B, no LoRA \\
\hline
\(\leq\)7.5k, ``5k'' &
micro=4, ga=2, global=16, lr=\(1{\times}10^{-5}\) &
micro=2, ga=4, global=16, lr=\(8{\times}10^{-6}\) \\
\(\leq\)17.5k, ``10k'' &
micro=4, ga=4, global=32, lr=\(2{\times}10^{-5}\) &
micro=2, ga=8, global=32, lr=\(1.5{\times}10^{-5}\) \\
\(>\)17.5k, ``25k'' &
micro=4, ga=8, global=64, lr=\(3{\times}10^{-5}\) &
micro=2, ga=16, global=64, lr=\(2{\times}10^{-5}\) \\
\hline
\end{tabular}%
}
\end{table}

\begin{table}[H]
\centering
\caption{\textbf{Automatic SFT hyperparameter heuristic with LoRA.}
\texttt{micro} is the per-device batch size, \texttt{ga} is gradient accumulation, and \texttt{global} is the target effective global batch size.}
\label{tab:sft_auto_lora}
\small
\scalebox{0.95}{%
\begin{tabular}{lll}
\hline
Dataset bucket & 3B, LoRA & 7B, LoRA \\
\hline
\(\leq\)7.5k, ``5k'' &
micro=16, ga=1, global=32, lr=\(1{\times}10^{-4}\) &
micro=8, ga=2, global=32, lr=\(6{\times}10^{-5}\) \\
\(\leq\)17.5k, ``10k'' &
micro=16, ga=2, global=64, lr=\(1.5{\times}10^{-4}\) &
micro=8, ga=4, global=64, lr=\(8{\times}10^{-5}\) \\
\(>\)17.5k, ``25k'' &
micro=16, ga=4, global=128, lr=\(2{\times}10^{-4}\) &
micro=8, ga=8, global=128, lr=\(1{\times}10^{-4}\) \\
\hline
\end{tabular}%
}
\end{table}

\clearpage

\section{Data Generation Procedure}
\label{app:data_generation}

This appendix describes the pipeline used to construct our datasets from stochastic structural causal models (SCMs). The pipeline consists of:
(i) rehydrating SCM code from a serialized JSON state representation,
(ii) sampling forward instantiations, or traces, from each SCM,
(iii) generating causal-query instances for deduction, intervention, abduction, and counterfactual tasks with fully enumerated supports, and
(iv) optionally assembling LLM-ready prompts.

\subsection{Support-Valued Semantics for Stochastic Queries}
\label{app:stochastic_support}

Our main experiments focus on deterministic query instances, where the verifier-induced answer support is a singleton. However, the same executable-SCM framework naturally extends to stochastic and partially observed settings. The key distinction in our approach is that we do not require the model to estimate calibrated probabilities over answers, since our goal is to study human-like causal reasoning: people typically do not answer causal queries by assigning explicit probabilities to all possible outcomes. Instead, each query is associated with a \emph{support set} of answers that have nonzero probability under the SCM and the revealed information, and a model output is considered correct if it belongs to this support. We note that other formulations are possible; the support-set criterion is the one we use for abduction in this work, and it provides one natural way to extend causal-reasoning evaluations to stochastic settings.

Let
\[
M=\langle \mathbf{U},p(\mathbf{u}),\mathbf{V},\mathbf{F}\rangle
\]
be a discrete SCM. For a fixed exogenous assignment $\mathbf{u}$, let
\[
\mathbf{v}(\mathbf{u})\in\mathcal{V}
\]
denote the full endogenous assignment obtained by executing the structural equations. If a query reveals only a subset of exogenous variables indexed by $\mathcal{I}$, with values $\bar{\mathbf{u}}_{\mathcal I}$, define the compatible exogenous contexts as
\[
\Omega(\bar{\mathbf{u}}_{\mathcal I})
\coloneqq
\left\{
\mathbf{u}\in\mathrm{supp}(p)
:
\mathbf{u}_{\mathcal I}=\bar{\mathbf{u}}_{\mathcal I}
\right\}.
\]
Thus, stochasticity and missingness are both represented as multiple feasible exogenous completions. Smaller $\mathcal I$ corresponds to more missing exogenous information and therefore typically a larger answer support.

\paragraph{Deduction.}
A deduction query asks what endogenous assignment can occur given the revealed exogenous information. Its answer support is
\[
\mathcal{A}^{\mathrm{ded}}(\bar{\mathbf{u}}_{\mathcal I})
\coloneqq
\left\{
\mathbf{v}(\mathbf{u})
:
\mathbf{u}\in\Omega(\bar{\mathbf{u}}_{\mathcal I})
\right\}.
\]
In deterministic query instances, this support has size one. In stochastic or partially observed instances, several endogenous assignments may be valid.

\paragraph{Intervention.}
For an intervention $do(\mathbf{X}=\mathbf{x}^{\star})$, let
$\mathbf{v}_{do(\mathbf{X}=\mathbf{x}^{\star})}(\mathbf{u})$
denote the endogenous assignment produced by replacing the structural equations for $\mathbf{X}$ with the externally assigned values $\mathbf{x}^{\star}$ while leaving all other mechanisms unchanged. The interventional answer support is
\[
\mathcal{A}^{\mathrm{int}}_{do(\mathbf{X}=\mathbf{x}^{\star})}
(\bar{\mathbf{u}}_{\mathcal I})
\coloneqq
\left\{
\mathbf{v}_{do(\mathbf{X}=\mathbf{x}^{\star})}(\mathbf{u})
:
\mathbf{u}\in\Omega(\bar{\mathbf{u}}_{\mathcal I})
\right\}.
\]

\paragraph{Abduction.}
Abduction is the clearest stochastic case. Given endogenous evidence
$\bar{\mathbf{v}}_{\mathcal O}$ over observed variables indexed by $\mathcal O$, the SCM induces a posterior distribution over exogenous contexts:
\[
p(\mathbf{u}\mid
\bar{\mathbf{u}}_{\mathcal I},
\bar{\mathbf{v}}_{\mathcal O})
\propto
p(\mathbf{u})\,
\mathbf{1}\{
\mathbf{u}_{\mathcal I}=\bar{\mathbf{u}}_{\mathcal I}
\}\,
\mathbf{1}\{
\mathbf{v}_{\mathcal O}(\mathbf{u})
=
\bar{\mathbf{v}}_{\mathcal O}
\}.
\]
Our verifier does not require estimating this posterior distribution. It only uses its support:
\[
\mathcal{A}^{\mathrm{abd}}
(\bar{\mathbf{u}}_{\mathcal I},\bar{\mathbf{v}}_{\mathcal O})
\coloneqq
\left\{
\mathbf{u}\in\Omega(\bar{\mathbf{u}}_{\mathcal I})
:
\mathbf{v}_{\mathcal O}(\mathbf{u})
=
\bar{\mathbf{v}}_{\mathcal O}
\right\}.
\]
An abductive answer $\hat{\mathbf{u}}$ is therefore correct whenever
\[
\hat{\mathbf{u}}
\in
\mathcal{A}^{\mathrm{abd}}
(\bar{\mathbf{u}}_{\mathcal I},\bar{\mathbf{v}}_{\mathcal O}).
\]
Equivalently, the model is rewarded for proposing any exogenous completion with nonzero posterior mass, not for recovering the posterior weights.

\paragraph{Counterfactuals.}
A counterfactual query combines abduction, action, and prediction. Given factual evidence $\bar{\mathbf{v}}_{\mathcal O}$ and an intervention $do(\mathbf{X}=\mathbf{x}')$, we first restrict to exogenous contexts consistent with the factual evidence and then evaluate the intervened SCM without resampling exogenous noise. The counterfactual support is
\[
\mathcal{A}^{\mathrm{cf}}_{do(\mathbf{X}=\mathbf{x}')}
(\bar{\mathbf{u}}_{\mathcal I},\bar{\mathbf{v}}_{\mathcal O})
\coloneqq
\left\{
\mathbf{v}_{do(\mathbf{X}=\mathbf{x}')}(\mathbf{u})
:
\mathbf{u}\in
\mathcal{A}^{\mathrm{abd}}
(\bar{\mathbf{u}}_{\mathcal I},\bar{\mathbf{v}}_{\mathcal O})
\right\}.
\]
This preserves the usual counterfactual semantics: factual evidence is used only to abduce possible exogenous contexts, and the intervention is then evaluated under those same contexts.

\paragraph{Verification.}
For any query $q$, let $\mathcal{A}(q)$ denote the corresponding support set:
\[
\mathcal{A}(q)
\in
\left\{
\mathcal{A}^{\mathrm{ded}},
\mathcal{A}^{\mathrm{int}},
\mathcal{A}^{\mathrm{abd}},
\mathcal{A}^{\mathrm{cf}}
\right\}.
\]
We score a model output $\hat a$ by support membership:
\[
r(\hat a,q)
=
\mathbf{1}\{\hat a\in\mathcal{A}(q)\}.
\]
Thus, exact-match verification in deterministic settings is a special case of support-membership verification where $|\mathcal{A}(q)|=1$. In stochastic or partially observed settings, multiple answers may be valid, and the verifier rewards any answer that is feasible under the executable SCM.

\paragraph{Relation to probability estimation.}
The executable SCM also induces probability weights over the support. For example, a deduction or intervention query could define
\[
P_q(a)
=
\sum_{\mathbf{u}\in\Omega_q}
p(\mathbf{u}\mid q)\,
\mathbf{1}\{g_q(\mathbf{u})=a\},
\]
where $g_q$ denotes the relevant SCM execution map for the query and $\Omega_q$ is the set of exogenous contexts compatible with the query information. Our current framework uses only
\[
\mathrm{supp}(P_q)=\{a:P_q(a)>0\}
\]
for verification. Learning calibrated probabilities over valid support elements is a natural extension, but is not required for the support-valued causal reasoning studied here.

\subsection{SCM Representation}
\label{app:scm_representation}

Each SCM is stored in a JSON state file containing a list of SCM entries under a top-level \texttt{scms} field. An SCM entry contains an \texttt{scm\_id} and a \texttt{history} list. Each element of \texttt{history} is a \emph{bundle} corresponding to a model version \texttt{v}, including:
(i) a set of exogenous sampler functions \texttt{U\_*},
(ii) structural assignment functions \texttt{f\_*} implementing deterministic structural equations conditional on exogenous noise, and
(iii) a driver function \texttt{run\_once(seed)} that samples a single world and returns outputs, either as a Python dictionary or as a dict-like object.

We use the convention that exogenous variables are sampled once per world, endogenous variables are computed via structural assignments, and interventions \texttt{do(X=x)} override only the equation for \texttt{X} while leaving all other structural equations unchanged.

\subsection{Stage 1: Rehydrating SCMs from JSON}
\label{app:rehydrating_scms}

We reconstruct standalone Python modules from each stored bundle using \texttt{rehydrate\_scms\_from\_json.py}. For each SCM \texttt{scm\_id} and each version \texttt{v} in an optional version range, the pipeline:
\begin{enumerate}
  \item Writes an executable Python module \texttt{scm\_\{id\}\_v\{v\}.py} by emitting imports, function signatures, and function bodies from the bundle. A \texttt{from \_\_future\_\_ import annotations} header is included to avoid import-time failures due to unevaluated type annotations.
  \item Optionally validates executability by importing the module and running \texttt{run\_once(seed=0)}. Versions failing validation are skipped.
  \item Writes a comment-stripped variant, \texttt{scm\_\{id\}\_v\{v\}\_nocomments.py}, obtained by removing docstrings and \texttt{\#}-style comments. This reduces prompt length and removes non-semantic text for downstream prompt construction.
  \item Writes the bundle metadata itself to \texttt{bundle\_v\{v\}.json}.
\end{enumerate}

\paragraph{Parallelism and CPU hygiene.}
Rehydration and sampling are parallelized across SCMs using a process pool. The worker count is selected using cpuset-aware or cgroup-aware CPU detection, or scheduler environment variables such as SLURM and PBS, and capped by the number of SCMs. To prevent oversubscription when using multiprocessing, numerical backends are constrained by setting \texttt{OMP\_NUM\_THREADS}, \texttt{MKL\_NUM\_THREADS}, and related variables to 1 unless already set.

\subsection{Stage 1b: Sampling Forward Instantiations}
\label{app:sampling_forward_instantiations}

For each validated SCM version, we generate a set of forward samples using \texttt{generate\_instantiations} in \texttt{rehydrate\_scms\_from\_json.py}. A single instantiation corresponds to one call to \texttt{run\_once(seed)} and produces:
\begin{itemize}
  \item the seed,
  \item the returned outputs, normalized to a dictionary when possible,
  \item a log of all values returned by exogenous samplers \texttt{U\_*} during execution, and
  \item a log of all values returned by structural functions \texttt{f\_*} during execution.
\end{itemize}

To obtain these traces, the pipeline monkey-patches \texttt{U\_*} and \texttt{f\_*} in the imported module with wrappers that record return values each time a function is called. Values are stored per function; if a function is called once, we store a scalar, and if it is called multiple times, we store a list.

\paragraph{Robust sampling under runtime errors.}
Some generated SCMs may fail for particular seeds, for example due to invalid casts or division by zero. Rather than aborting the entire SCM version, we iterate through increasing seeds and skip seeds that raise exceptions, collecting successful samples until a target count is reached. The number of attempts is bounded, by default proportional to the requested sample count with a hard cap, to avoid infinite loops on pathological models. If no seeds succeed, the version is marked as failed for instantiation generation. Instantiation files are written to \texttt{instantiations\_v\{v\}.json}, and latest-version convenience copies are written to \texttt{instantiations.json} and \texttt{bundle.json}.

\subsection{Stage 2: Generating Causal-Query Datasets}
\label{app:generating_causal_queries}

From each SCM version, we construct supervised causal-query instances using \texttt{build\_causal\_query\_instantiations.py}. Each instance is defined by:
(i) a partial specification of the exogenous variables, with some fixed and others declared missing,
(ii) optionally an intervention,
(iii) optionally observed endogenous evidence, and
(iv) a \emph{support set} of all valid answers under the SCM semantics, subject to computational caps.

\paragraph{Domain inference.}
Before generating queries, the pipeline estimates discrete domains for exogenous and endogenous variables by forward sampling, using parameter \texttt{domain\_samples}; the default shown in the code is 2000. These inferred domains are used to choose feasible subsets of missing variables and interventions, and to bound enumeration of supports.

\paragraph{Seeds and attempt schedule.}
For each SCM version, seeds are scanned in a contiguous range:
\[
\texttt{seed} \in [\texttt{seed\_start},\ \texttt{seed\_start}+\texttt{seed\_budget}),
\]
with configurable \texttt{seed\_start}, also settable by environment variable, and \texttt{seed\_budget}. For each seed, we generate a factual world by sampling all exogenous variables and forward-computing the corresponding endogenous values. We then try a small number of randomized attempts per seed, controlled by \texttt{attempts\_per\_seed}, to construct a valid query meeting the desired constraints.

\paragraph{Missing-exogenous selection and feasibility caps.}
In the stochastic setting, we hide a subset of exogenous variables, controlled by \texttt{missing\_exo\_pct}, but enforce feasibility: the Cartesian product of missing-variable domain sizes is constrained by \texttt{max\_combinations}. We also cap the final enumerated support size by \texttt{max\_support\_size}. Candidate missing variables with trivial domains, meaning size less than 2, are avoided.

\paragraph{Query types.}
We generate four query types, each stored as a JSON file with \texttt{meta} and \texttt{instances} fields.

\begin{enumerate}
  \item \textbf{Deduction.} Given fixed exogenous values and a list of missing exogenous names, the model must predict the possible endogenous outputs in the factual, no-intervention world. The support is computed by enumerating all completions of missing exogenous assignments, within \texttt{max\_combinations}, and forward-propagating through the SCM. We keep only instances where enumeration is exhaustive, \texttt{exhausted=True}, and non-empty. For the \texttt{deterministic} scenario label, we further require a unique output, with support size 1; for the \texttt{stochastic} scenario label, we require multiple outputs, with support size at least 2.

  \item \textbf{Intervention.} This is analogous to deduction, but with a \texttt{do}-operator applied to a randomly selected subset of endogenous variables, controlled by \texttt{intervention\_node\_pct}. Intervention values are chosen to differ from the factual value when possible. We compute the enumerated support of outputs under the intervention. An optional constraint, \texttt{require\_intervention\_effect}, filters out degenerate deterministic interventions that do not change the output relative to the factual world.

  \item \textbf{Abduction.} Given observed endogenous values from the factual world, the model must infer consistent assignments to missing exogenous variables. In the default formulation, all exogenous variables are treated as missing and the full endogenous assignment is provided as evidence. The support is the set of all exogenous completions consistent with the evidence, enumerated up to the same caps.

  \item \textbf{Counterfactual.} Counterfactual queries implement the standard three-step semantics: (1) abduce exogenous values consistent with observed endogenous evidence, (2) apply an intervention \texttt{do}, and (3) predict outputs under the intervention using the same abduced exogenous values, with no resampling. We support two modes: a minimal-evidence mode where evidence is restricted to a small observed endogenous subset linked to a missing exogenous parent, selected using SCM graph metadata when available; and a full-evidence mode using all endogenous variables as observations. As with other query types, we require exhaustive enumeration and scenario-dependent support-size constraints. An optional \texttt{require\_counterfactual\_effect} filter removes deterministic counterfactuals with no effect.
\end{enumerate}

\paragraph{Outputs.}
For each SCM \texttt{scm\_id} and version \texttt{v}, causal-query instances are written to:
\[
\texttt{scm\_\{id\}/causal\_queries/<query\_type>/<scenario>/causal\_queries\_v\{v\}.json},
\]
and a convenience copy \texttt{causal\_queries.json} is maintained for the latest version.

\subsection{Stage 3: Building Prompt Instances}
\label{app:building_prompt_instances}

Finally, we assemble LLM-ready prompt instances from the causal-query datasets using \texttt{build\_causal\_query\_prompts.py}. Each prompt instance contains:
(i) SCM code, preferably the comment-stripped module, or an unstructured natural-language SCM description if provided in metadata; and
(ii) a task instruction block specifying the query type, given exogenous values, missing exogenous names, optional \texttt{do}, optional observed endogenous evidence, and strict output-formatting requirements, namely JSON with a fixed set of keys.

Each prompt also stores an \texttt{expected\_output} object containing the enumerated support set and the required output keys. This enables automatic verification of model outputs.

\paragraph{Selection of complete SCMs and robustness.}
Prompt construction scans SCMs in the state file and selects the first \(N\) \emph{complete} SCMs, where all required versions and causal-query files exist and meet minimum example counts. To mitigate attrition from invalid SCMs, the generation scripts optionally reserve a margin of extra SCMs, by default 20, during earlier processing. The prompt builder writes a sidecar file listing selected SCM IDs for reproducibility and, optionally, for aligning intervention instances with deduction instances.

\subsection{Summary}
\label{app:data_generation_summary}

Overall, the dataset is generated by executable SCM reconstruction, robust forward sampling with seed-skipping for failures, exhaustive capped support enumeration for each query instance, and optional prompt assembly for LLM evaluation. The pipeline records all hyperparameters, including seed ranges, support caps, missing-variable fractions, and intervention fractions, in per-file metadata to support exact reproduction.

\section{Additional Results}
\label{app:additional_results}

This section reports supplementary results for formal causal-query training, NIHSS data augmentation, pass@k evaluation, and additional ablations.

\subsection{Training on Formal Causal Queries}
\label{app:formal_causal_queries}

\begin{figure}[H]
  \centering
  \includegraphics[width=1.\textwidth]{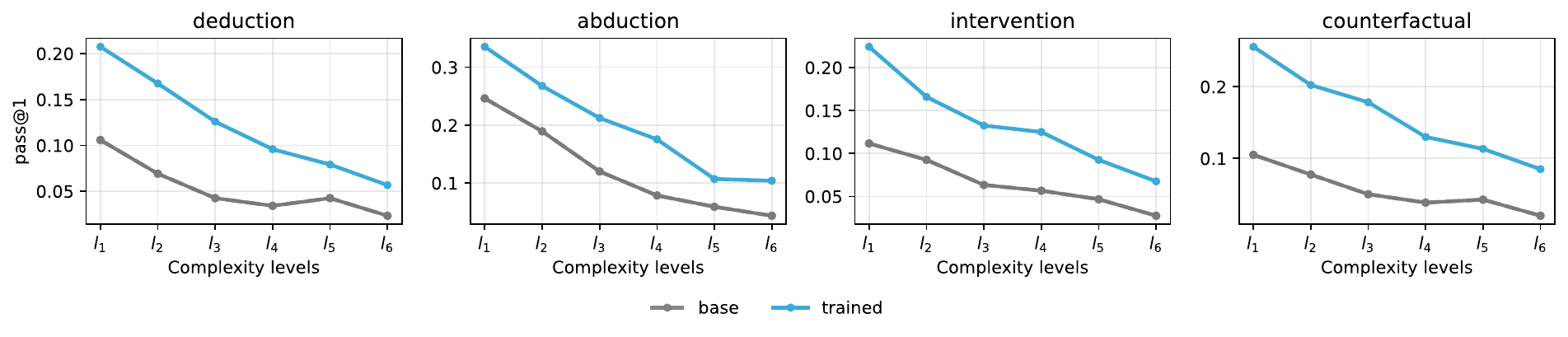}
  \vspace{-1.5em}
  \captionsetup{font=footnotesize}
  \caption{\textbf{Training on formal causal queries.}
  Pass@1 accuracy by causal query type for models trained with and without simulator-generated causal queries. Fine-tuning with causal-simulator data consistently improves performance over the base model.}
  \label{fig:formal_causal}
  \vspace{-.75em}
\end{figure}

\subsection{Evaluation of Causal-Simulator Data Augmentation}
\label{subsec:exps_p3}

We evaluate the value of transforming an existing informal SCM into executable form, automatically generating additional training data, and augmenting data in the original domain.

\paragraph{Experimental setup.}
We construct a controlled ``original'' dataset derived from the NIHSS dataset, consisting of 50 interventional and 50 counterfactual samples. We then convert this dataset into a formal SCM representation, following the procedure in Experiment~3, and generate an additional 1{,}000 interventional and counterfactual instances. We train two models: (i) a model trained only on the original dataset and (ii) a model trained on the original dataset augmented with simulator-generated data.

\paragraph{Results.}
Table~\ref{tab:causal_simulator_augmentation} shows that adding causal-simulator data significantly improves performance in the unstructured domain, where ground-truth answers to causal queries would otherwise be unavailable.

\begin{table}[H]
\centering
\caption{\textbf{Causal reasoning performance on the NIHSS dataset with and without causal-simulator data augmentation.}
The augmented model uses 1{,}000 simulator-generated interventional and counterfactual instances, while the original model uses 50 of each. Columns indicate the number of nodes in the SCM.}
\label{tab:causal_simulator_augmentation}
\small
\setlength{\tabcolsep}{4pt}
\begin{tabular}{c|cccccc}
\hline
\textbf{Model} & \textbf{4} & \textbf{5} & \textbf{6} & \textbf{7} & \textbf{8} & \textbf{9} \\
\hline
\textbf{Original} & 0.239 & 0.230 & 0.261 & 0.255 & 0.213 & 0.191 \\
\textbf{Augmented} & \textbf{0.423} & \textbf{0.344} & \textbf{0.398} & \textbf{0.351} & \textbf{0.349} & \textbf{0.378} \\
\hline
\end{tabular}
\end{table}

\begin{table}[H]
\centering
\caption{\textbf{Pass@k performance on the NIHSS benchmark with and without causal-simulator data augmentation.}
The augmented model, INT\_CF\_1000, uses 1{,}000 simulator-generated interventional and counterfactual instances. The base model uses 50 of each. Columns indicate the number of nodes in the SCM.}
\label{tab:passk_nihss}
\resizebox{\textwidth}{!}{%
\begin{tabular}{lccccccccc}
\hline
SCM size & 1 & 2 & 3 & 4 & 5 & 6 & 7 & 8 & 9 \\
\hline
Base & 0.98\basedelta & 0.76\basedelta & 0.72\basedelta & 0.65\basedelta & 0.66\basedelta & 0.70\basedelta & 0.56\basedelta & 0.49\basedelta & 0.43\basedelta \\
INT\_CF\_1000
& \textbf{0.99} \textcolor{green!60!black}{(+0.01)}
& \textbf{0.82} \textcolor{green!60!black}{(+0.06)}
& \textbf{0.82} \textcolor{green!60!black}{(+0.10)}
& \textbf{0.80} \textcolor{green!60!black}{(+0.16)}
& \textbf{0.76} \textcolor{green!60!black}{(+0.10)}
& \textbf{0.76} \textcolor{green!60!black}{(+0.06)}
& \textbf{0.75} \textcolor{green!60!black}{(+0.18)}
& \textbf{0.66} \textcolor{green!60!black}{(+0.17)}
& \textbf{0.66} \textcolor{green!60!black}{(+0.23)} \\
\hline
\end{tabular}%
}
\end{table}

\subsection{All Causal Query Types}
\label{app:all_causal_query_types}

\begin{figure}[H]
  \centering
  \includegraphics[width=1.\textwidth]{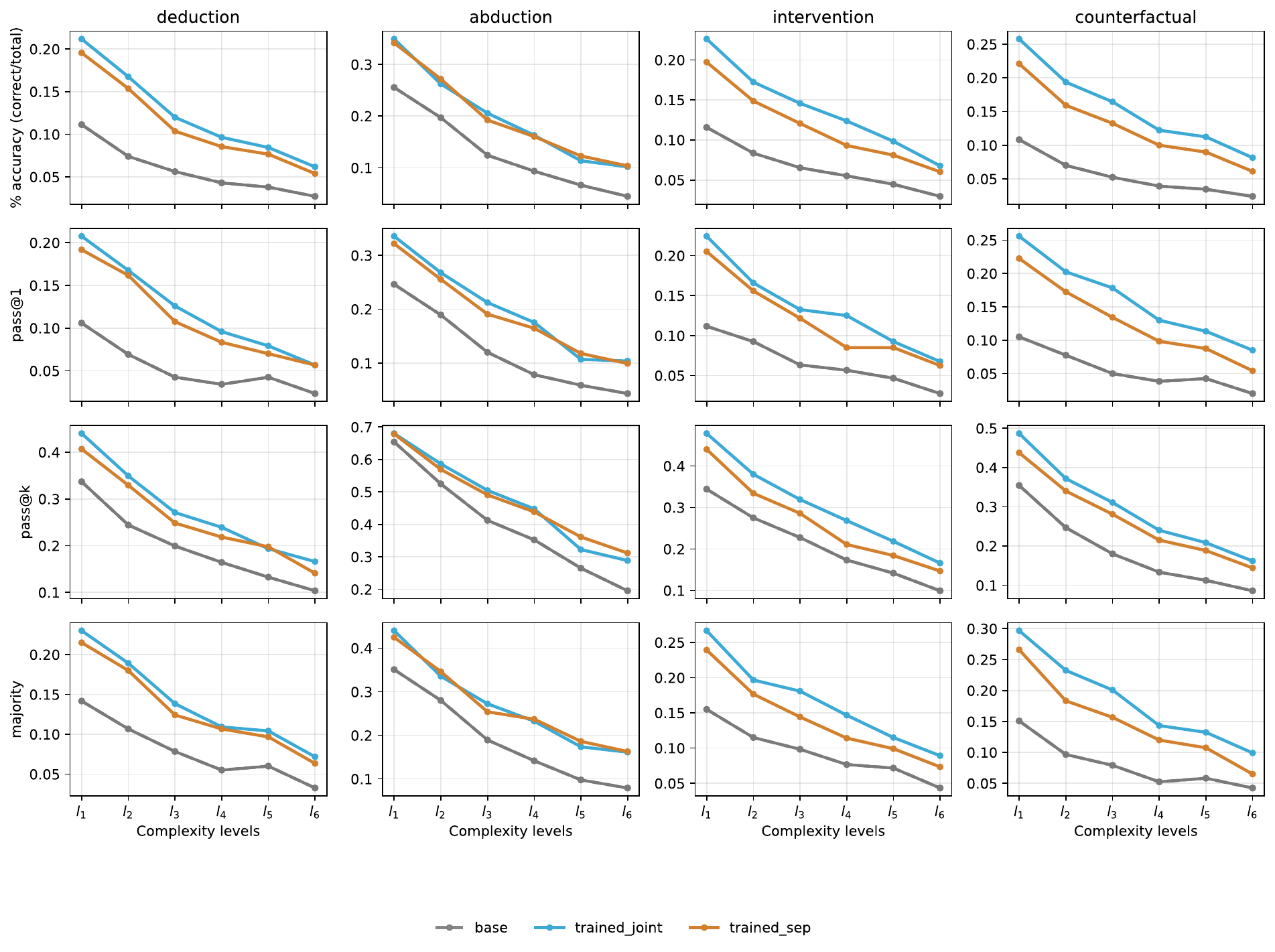}
  \vspace{-1.5em}
  \captionsetup{font=footnotesize}
  \caption{\textbf{Joint results across all causal query types.}
  Results for deduction, abduction, intervention, and counterfactual queries.}
  \label{fig:joint_query_results}
  \vspace{-.75em}
\end{figure}

\clearpage

\subsection{Data Amount Ablation}
\label{app:data_amount_ablation}

\begin{figure}[H]
  \centering
  \includegraphics[width=.8\textwidth]{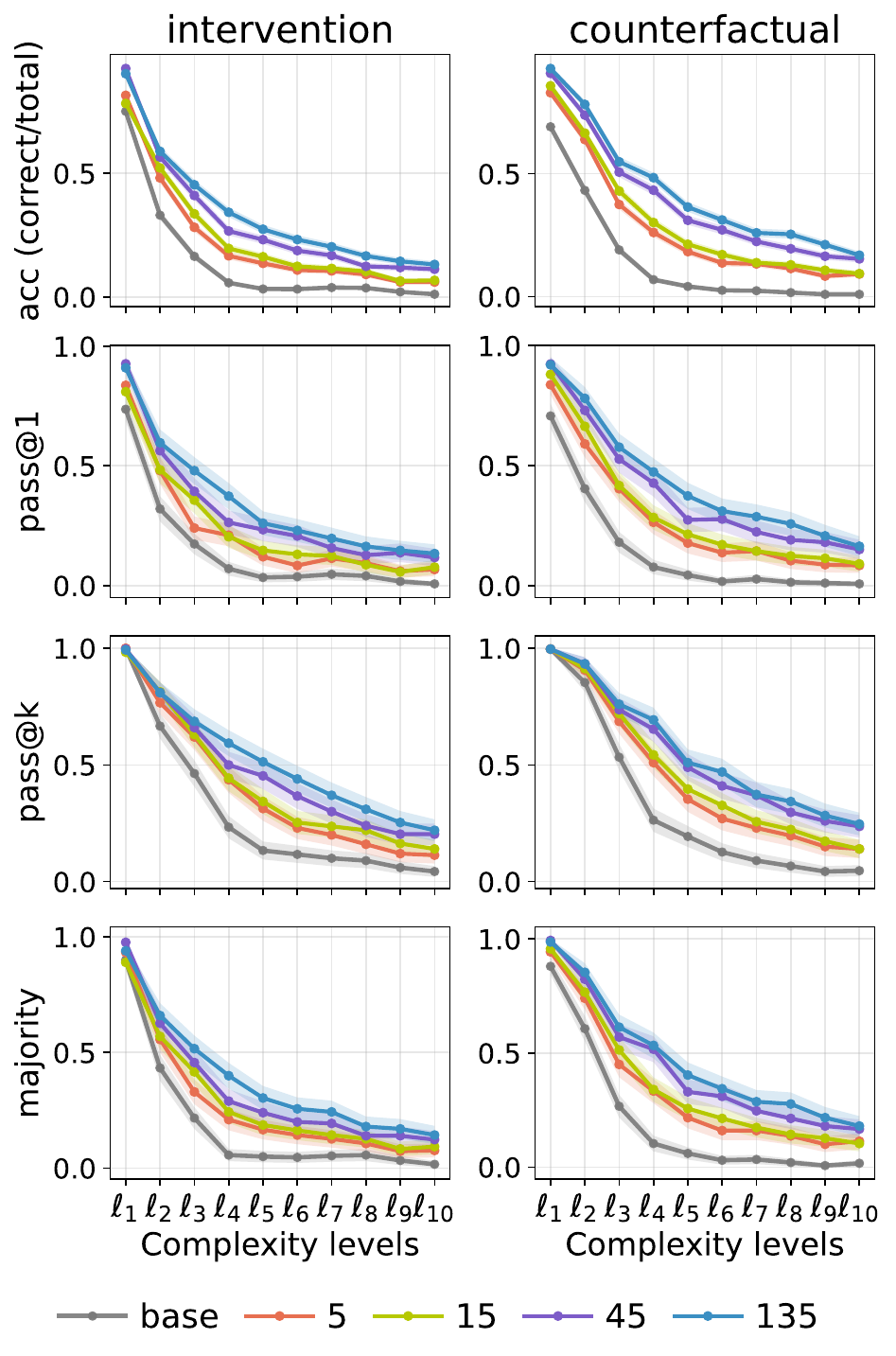}
  \vspace{-1.5em}
  \captionsetup{font=footnotesize}
  \caption{\textbf{Ablation varying the amount of training data.}
  Test results on \textsc{nonsense}-formal SCMs for intervention and counterfactual queries, averaged across structures. Models are trained jointly on counterfactual and intervention causal queries. We show levels 1-10.}
  \label{fig:data_amount_ablation}
  \vspace{-.75em}
\end{figure}

\clearpage
\begin{figure}[H]
  \centering
  \includegraphics[width=.8\textwidth]{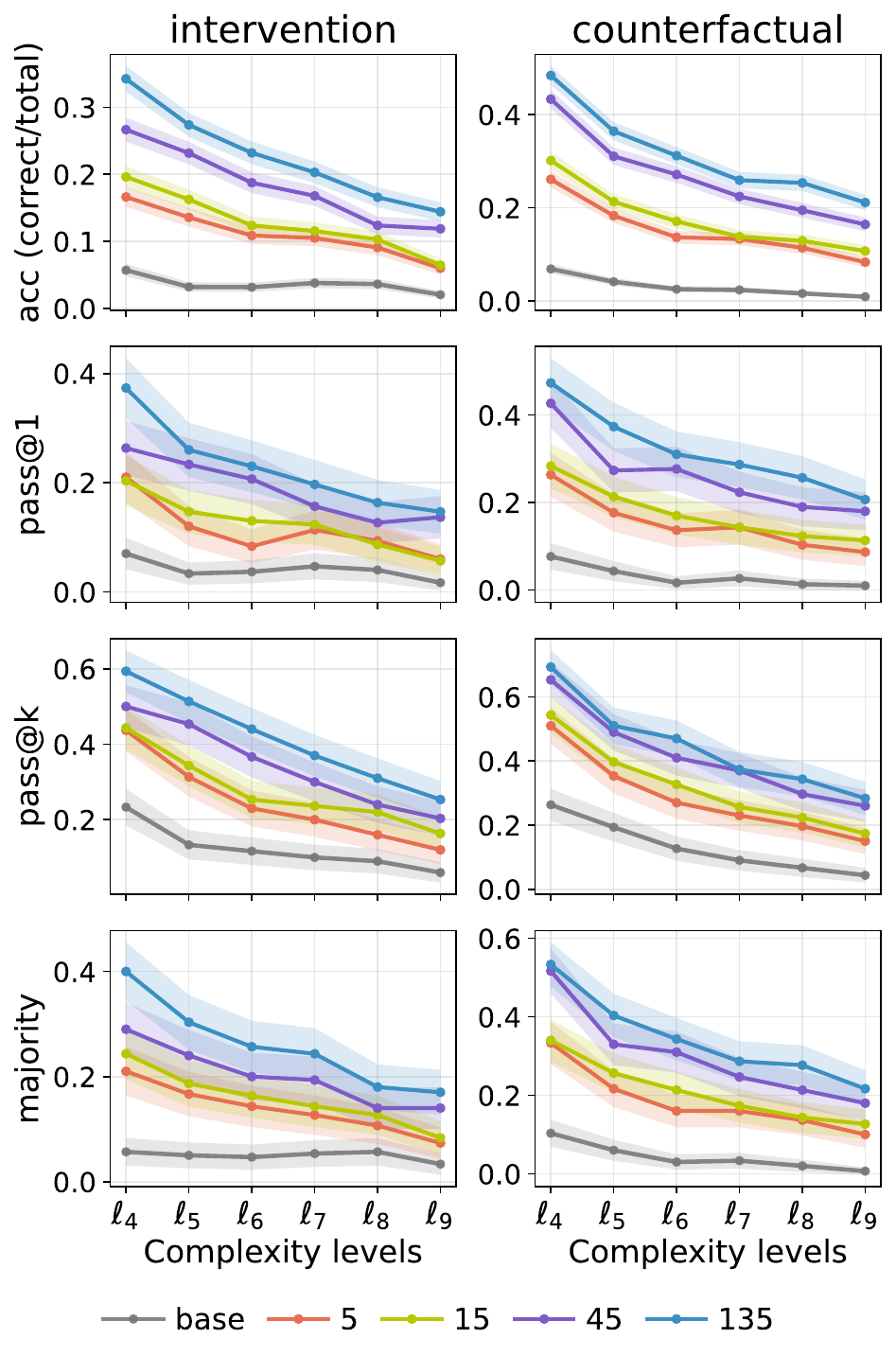}
  \vspace{-1.5em}
  \captionsetup{font=footnotesize}
  \caption{\textbf{Ablation varying the amount of training data.}
  Test results on \textsc{nonsense}-formal SCMs for intervention and counterfactual queries, averaged across structures. Models are trained jointly on counterfactual and intervention causal queries. We show levels 4-9.}
  \label{fig:data_amount_ablation2}
  \vspace{-.75em}
\end{figure}

\clearpage

\subsection{Curriculum Ablation}
\label{app:curriculum_ablation}

\begin{figure}[H]
  \centering
  \includegraphics[width=.8\textwidth]{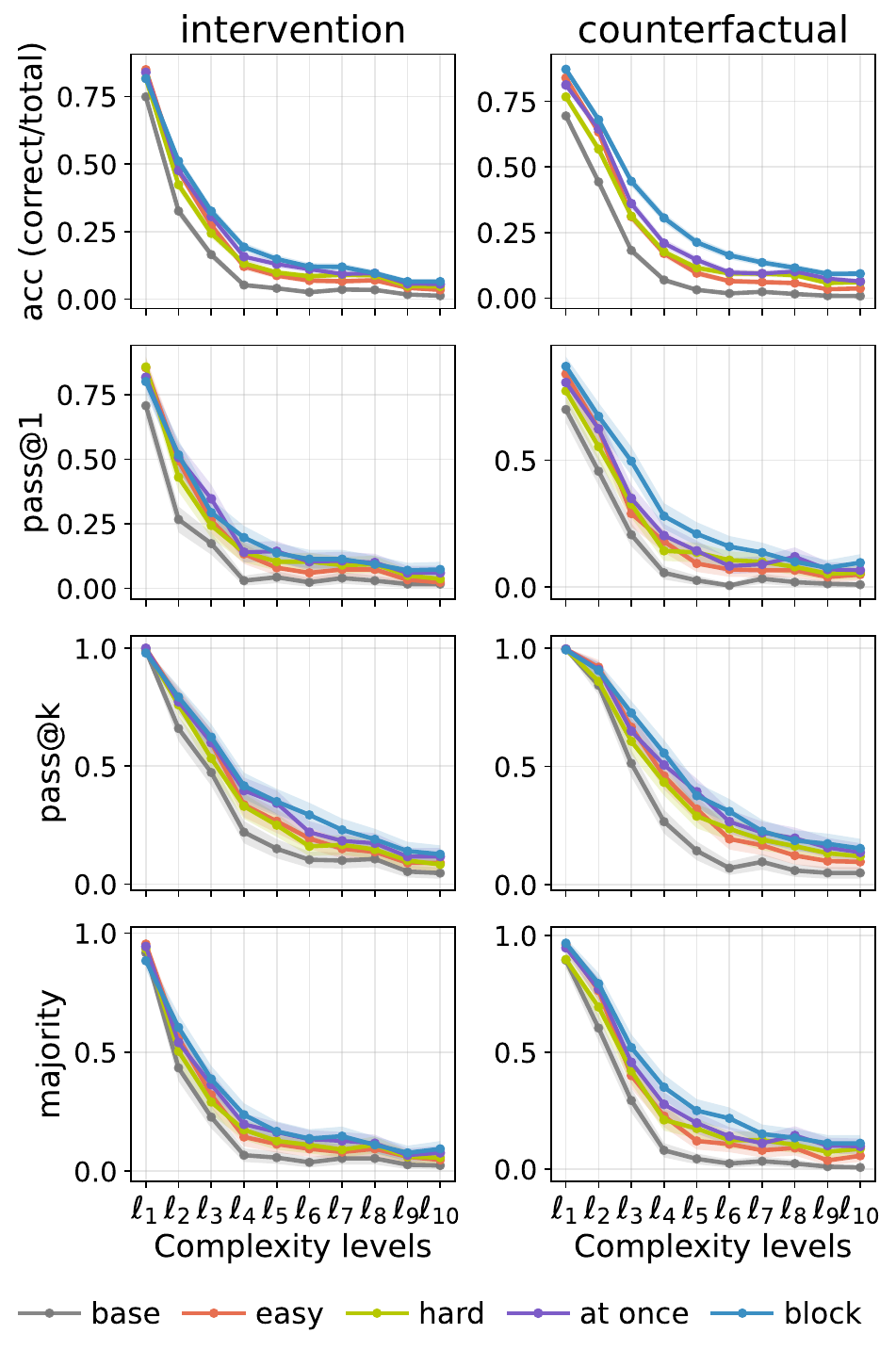}
  \vspace{-1.5em}
  \captionsetup{font=footnotesize}
  \caption{\textbf{Ablation varying the curriculum.}
  Test results on \textsc{nonsense}-formal SCMs for intervention and counterfactual queries, averaged across structures. Models are trained jointly on counterfactual and intervention causal queries. We show levels 1-10.}
  \label{fig:curriculum_ablation}
  \vspace{-.75em}
\end{figure}
\clearpage
\begin{figure}[H]
  \centering
  \includegraphics[width=.8\textwidth]{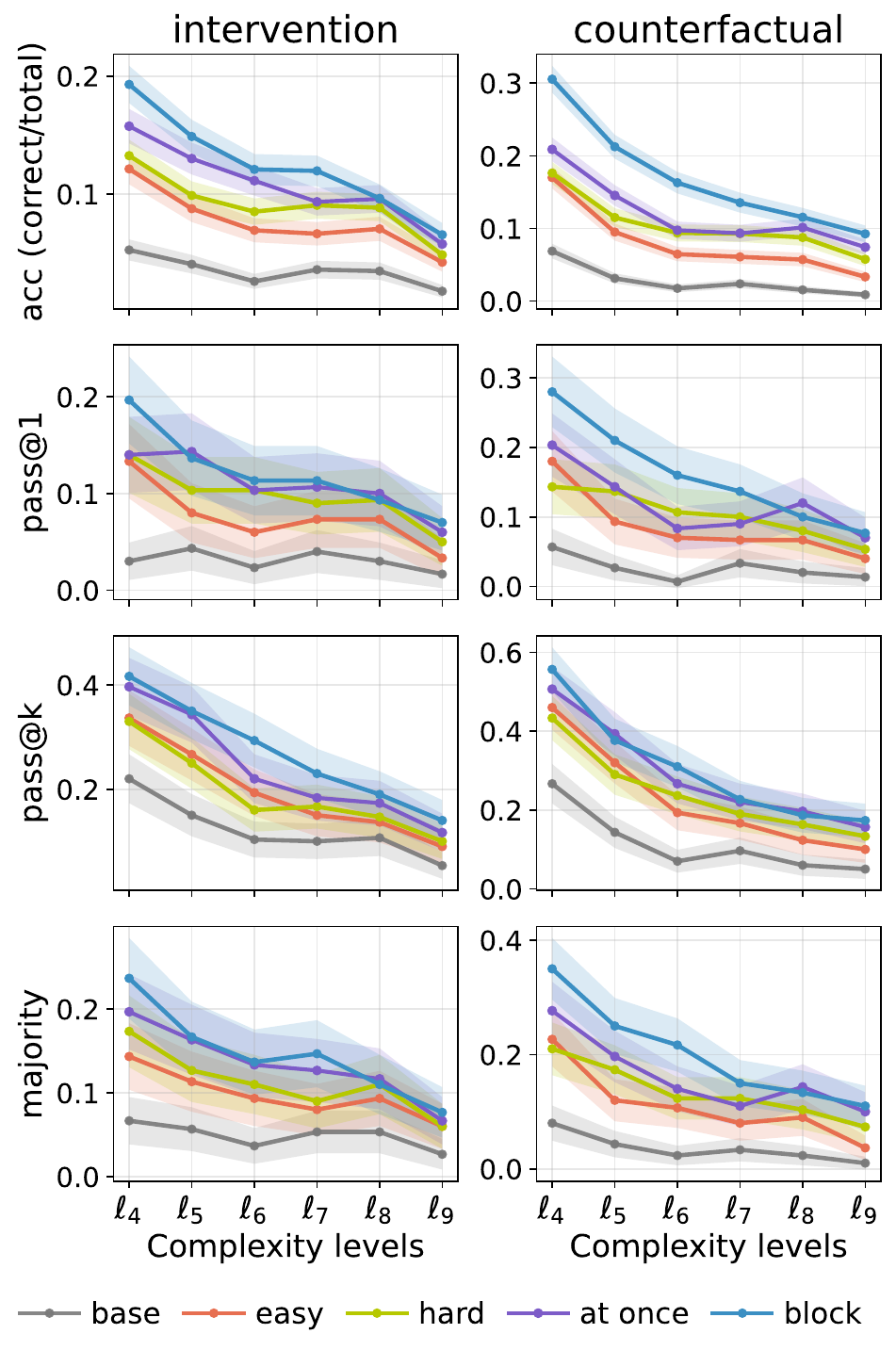}
  \vspace{-1.5em}
  \captionsetup{font=footnotesize}
  \caption{\textbf{Ablation varying the curriculum.}
  Test results on \textsc{nonsense}-formal SCMs for intervention and counterfactual queries, averaged across structures. Models are trained jointly on counterfactual and intervention causal queries. We show levels 4-9.}
  \label{fig:curriculum_ablation2}
  \vspace{-.75em}
\end{figure}

\subsection{Number of SCMs vs.\ Queries per SCM}
\label{app:scm_query_ablation}

Our data-scaling experiments vary the total amount of generated supervision. We additionally ask whether, for a fixed query budget, it is preferable to increase the number of distinct SCMs or instead generate more queries from each SCM.

We train in the \textsc{nonsense}-\textit{code} domain while varying the number of SCM IDs and the number of queries per $(\text{SCM ID}, \text{complexity level})$. Complexity ranges from 1 to 10 endogenous nodes. We compare three allocations with the same number of queries per complexity level: 180 SCM IDs with 5 queries each, 60 SCM IDs with 15 queries each, and 20 SCM IDs with 45 queries each. We evaluate on both held-out \textsc{nonsense}-\textit{code} and \textsc{medical}-\textit{code}.

\begin{table}[t]
\centering
\small
\begin{tabular}{lccccc}
\toprule
\textbf{Evaluation} & \textbf{SCM IDs} & \textbf{Queries/SCM} &
\textbf{Avg. Pass@1} & \textbf{Counterfactual} & \textbf{Intervention} \\
\midrule
\textsc{nonsense}-\textit{code}
& 180 & 5  & 25.63 & \textbf{28.23} & 23.03 \\
& 60  & 15 & 25.72 & 27.63 & 23.80 \\
& 20  & 45 & \textbf{26.03} & 27.20 & \textbf{24.87} \\
\midrule
\textsc{medical}-\textit{code}
& 180 & 5  & 24.98 & 24.52 & \textbf{25.45} \\
& 60  & 15 & 24.57 & 25.62 & 23.52 \\
& 20  & 45 & \textbf{25.09} & \textbf{26.21} & 23.97 \\
\bottomrule
\end{tabular}
\captionsetup{font=small}
\caption{Ablation of the number of distinct SCMs versus the number of queries generated per SCM at a fixed query budget.}
\label{tab:scm_query_ablation}
\end{table}
Performance is highly consistent across the three allocations. Average Pass@1 varies by at most 0.40 percentage points on \textsc{nonsense}-\textit{code} and 0.52 points on \textsc{medical}-\textit{code}. Within the range considered, we therefore find little sensitivity to whether a fixed supervision budget is allocated toward more distinct SCMs or more queries per SCM. A broader study of this breadth--depth trade-off remains an interesting direction for future work.

\end{document}